\documentclass[acmtog,nonacm]{acmart}

\usepackage{booktabs} 
\usepackage{adjustbox}
\usepackage{calc}
\usepackage{wrapfig}
\usepackage{multirow}

\usepackage[ruled]{algorithm2e} 

\SetAlFnt{\small}
\SetAlCapFnt{\small}
\SetAlCapNameFnt{\small}
\SetAlCapHSkip{0pt}
\setcopyright{none}

\begin{document}
\title{Content-aware Tile Generation using Exterior Boundary Inpainting}

\author{Sam Sartor}
\affiliation{%
  \institution{College of William \& Mary}
  \city{Williamsburg}
  \country{USA}
}
\orcid{0009-0001-1915-6887}

\author{Pieter Peers}
\affiliation{%
  \institution{College of William \& Mary}
  \city{Williamsburg}
  \country{USA}
}
\orcid{0000-0001-7621-9808}

\renewcommand\shortauthors{Sartor and Peers}

\def\equationautorefname~#1\null{Equation~(#1)\null}
\def\sectionautorefname{Section}
\def\subsectionautorefname{Section}

\def\etal{{et al.}}
\def\ie{{i.e.}}
\def\eg{{e.g.}}

\def\BF{\textbf}
\def\UL{\underline}

\newcommand{\scheme}[1]{\textsc{#1}}
\newcommand{\prompt}[1]{\textsc{#1}}

\newcommand{\rowfont}[1]{
\gdef\rowfonttype{#1}#1\ignorespaces%
}

\def\C{C}
\def\T{\mathcal{T}}
\def\D{\mathcal{D}}
\def\x{x}
\def\y{y}
\def\c{c}
\def\N{{N}}
\def\S{{S}}
\def\E{{E}}
\def\W{{W}}
\def\mod{~mod~}

\begin{abstract}
  We present a novel and flexible learning-based method for generating
  tileable image sets.  Our method goes beyond simple self-tiling,
  supporting sets of mutually tileable images that exhibit a high
  degree of diversity.  To promote diversity we decouple structure
  from content by foregoing explicit copying of patches from an
  exemplar image.  Instead we leverage the prior knowledge of natural
  images and textures embedded in large-scale pretrained diffusion
  models to guide tile generation constrained by exterior boundary
  conditions and a text prompt to specify the content. By carefully
  designing and selecting the exterior boundary conditions, we can
  reformulate the tile generation process as an inpainting problem,
  allowing us to directly employ existing diffusion-based inpainting
  models without the need to retrain a model on a custom training set.
  We demonstrate the flexibility and efficacy of our content-aware
  tile generation method on different tiling schemes, such as Wang
  tiles, from only a text prompt.  Furthermore, we introduce a novel
  Dual Wang tiling scheme that provides greater texture continuity and
  diversity than existing Wang tile variants.
\end{abstract}

%
%



\begin{teaserfigure}
\centering
\includegraphics[width=.99\textwidth]{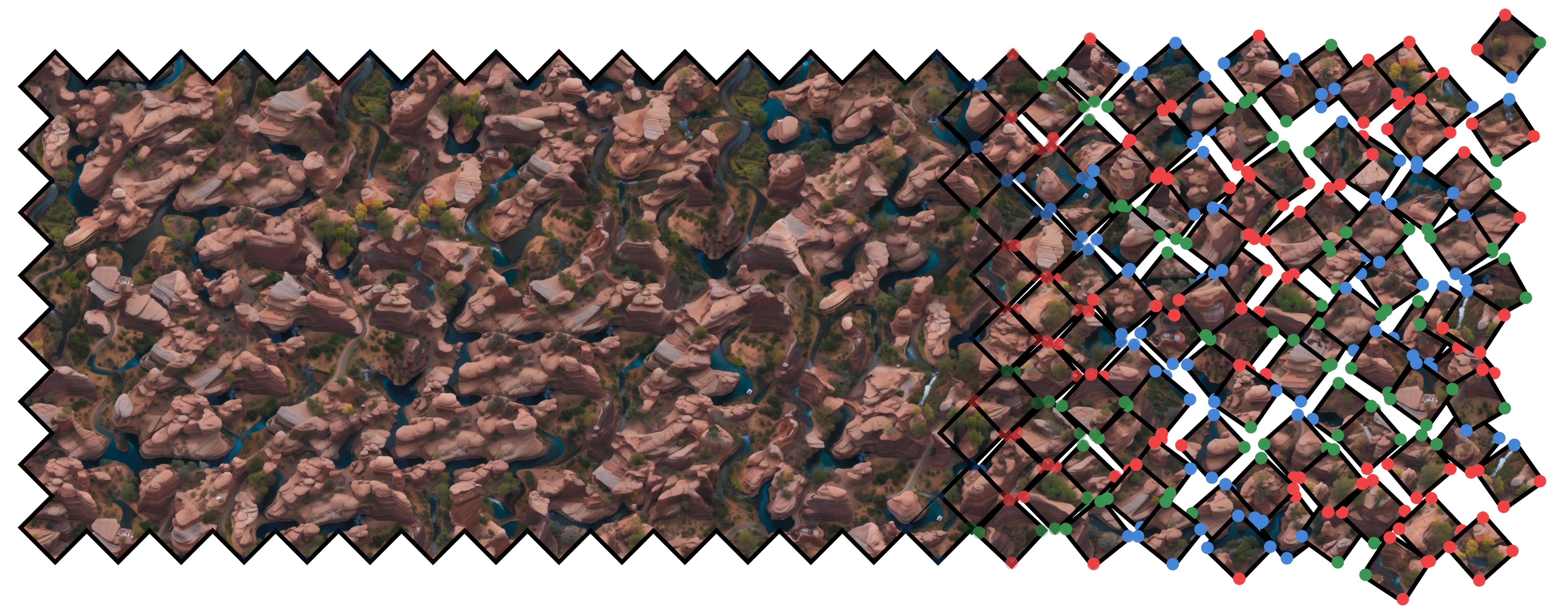}
\vspace{-0.5cm}
\caption{\emph{Left:} An example of a textured $3$-color Dual Wang
  tiling with diamond shaped tiles generated from the text-prompt
  \textit{``drone view of canyonlands in utah, rivers, tiny
    trees''}. \emph{Right:} selected textured tiles used to create the
  tiling shown on the left.}
\label{fig:teaser}
\end{teaserfigure}

\maketitle

\section{Introduction}
\label{sec:intro}

Textures are ubiquitous in computer graphics. A rich variety of
methods have been introduced to aid in the creation of textures,
ranging from specialized texture editors, to exemplar-based texture
synthesis methods~\cite{Efros:1999:TSN}, and to procedural
methods~\cite{Perlin:1985:AIS}.  Classic texture synthesis methods,
however, generally couple structure and content by verbatim copying
parts from an exemplar without understanding the semantics of the
content, thereby adversely affecting the diversity of the generated
textures.

Advances in machine learning have renewed the interest in texture
synthesis. Generative networks, especially prompt-conditioned
diffusion models, have good semantic understanding of the image
content.  The availability of large pretrained diffusion models makes
these models an attractive candidate for generating new textures
without the need to gather large sets of training textures.  However,
diffusion models generate whole textures/images at once, and thus
these models are limited in the size of the texture they can
synthesize.  Tileable textures circumvent this problem by synthesizing
one or more small textures that can be seamlessly tiled into a larger
texture.  However, existing learning-based
methods~\cite{Vecchio:2023:CAC,Rodriguez-Pardo:2024:TAD} are limited
to generating a single self-tiling texture, yielding a tiling that
exhibits noticeable repetition.

In this paper we address both tileability as well as diversity in
content-aware tile generation. To promote diversity, we forego
explicit copying of parts of an exemplar texture, and instead rely on
the prior knowledge embedded in prompt-conditioned diffusion models to
synthesize a unique content per tile.  Our key idea is to condition
the synthesis of the image on exterior boundary conditions derived
from the exemplar.  Conceptually, this is akin to solving for the
interior of a domain using partial differential equations (PDE) with
boundary conditions at the edge of the domain, except that we leverage
a diffusion model for computing the solution corresponding to the
content specified by the prompt.  For flexibility and simplicity, we
avoid designing and training a specialized diffusion model to generate
tiles conditioned on prompt and exterior boundary conditions, but
instead we leverage existing pre-trained diffusion-based inpainting
models by carefully designing and selecting the \emph{exterior}
boundary conditions to synthesize each tile's \emph{interior}
separately while ensuring tileability.

We demonstrate the flexibility of our content-aware tile generation on
a wide variety of tile types: self-tiling tiles, stochastic
self-tiling tiles, textured Escher tiles, Wang tiles, and a novel Dual
Wang tiling that addresses the difference in diversity between the
corners and the interior of Corner Wang tiles~\cite{Lagae:2005:POD}.
To the best of our knowledge, this simple yet flexible method for
generating tile sets using exterior boundary inpainting has not been
explored in literature, nor have we found any similar implementation
on the various public repositories in the diffusion community. To
facilitate reproduction of our results and to stimulate further
research, a reference implementation of our context-aware
tile generation can be found at
\url{https://github.com/samsartor/content_aware_tiles}.

In summary, our contributions are:
\begin{enumerate}
\item a flexible method for content-aware generation of diverse tile
  sets by specifying exterior boundary conditions;
\item that enables tileability for diffusion-based image synthesis
  beyond self-tiling; and
\item new tiling schemes such as stochastic self-tiling and textured
  Escher tiling, as well as a Dual Wang tile formulation that solves
  the corner versus interior diversity problem with Corner Wang
  tiles~\cite{Lagae:2005:POD}.
\end{enumerate}
This paper is an extended version of our ACM TOG paper with the same
title which can be found at: \url{https://doi.org/10.1145/3687981}.

\section{Related Work}
\label{sec:related}

\paragraph{Texture Synthesis}
Classic texture synthesis algorithms can be categorized as either
parametric methods~\cite{Heeger:1995:PTA} that optimize a texture to
match learned statistics from an exemplar, or as non-parametric
techniques that reassemble patches from an exemplar either by
copying~\cite{Efros:1999:TSN}, quilting~\cite{Efros:2001:IQT},
optimization~\cite{Kwatra:2005:TOE},
graph-cuts~\cite{Kwatra:2003:GTI}, or randomized
correspondences~\cite{Barnes:2009:PMR}.

Recent successes of neural networks for various image processing tasks
also generated a renewed interest in texture synthesis. The vast
majority of machine learning based texture synthesis approaches fall
in the first category of parametric texture synthesis.
Optimization-based approaches replace the manually crafted filters by
statistics over activations from a pretrained convolutional neural
network~\cite{Gatys:2015:TSC,Heitz:2021:SWL}.  However, optimization
methods are computationally costly, and various feed-forward networks
have been introduced to directly output the resulting texture.  These
feed-forward network solutions can be categorized on whether they are
trained per texture
exemplar~\cite{Rodriguez:2022:SSS,Li:2016:PRT,Ulyanov:2016:TNF,Zhou:2018:NST,Nardani:2020:NFU,Zhou:2023:NTS}
or per texture
category~\cite{Ulyanov:2017:ITN,Li:2017:DTS,Bergmann:2017:LTM,Guo:2022:UAT,Yu:2019:TMN}.
The vast majority of these methods can only synthesize a larger
texture as a whole, either by repeated
doubling~\cite{Guo:2022:UAT,Ulyanov:2017:ITN,Ulyanov:2016:TNF,Li:2017:DTS,Zhou:2018:NST,Zhou:2023:NTS},
leveraging a fully convolutional
architecture~\cite{Li:2016:PRT,Bergmann:2017:LTM}, or by applying a
neural Fast Fourier Transform~\cite{Nardani:2020:NFU}.  An alternative
approach to feed-forward networks are pointwise evaluation models
parameterized as a Multi-Layer Perceptron
(MLP)~\cite{Henzler:2020:LN3,Henzler:2021:GMB,Portenier:2020:GD3} that
take a position and a crop from a large noise field as input.  While
these methods can synthesize an infinitely large texture, they are
limited to textures with a high degree of stochasticity.

A final class, to which our method also belongs, are tiling-based
synthesis methods which sit in the middle between point-evaluation and
feed-forward methods. Tile-based methods precompute small tiles that
are merged together at run-time.  Classic tileable synthesis methods
either explicitely maximize stationarity~\cite{Moritz:2017:TST} or
extract the largest tileable patch from an
exemplar~\cite{Rodriguez:2019:AES}.  However, tiling a single texture
often results in a clear visible repetition.
Vanhoey~\etal~\shortcite{Vanhoey:2013:OMS} and
Kolvar~\etal~\shortcite{Kolavr:2016:RTS} exchange patches in a
self-tiling texture with a small precomputed set of compatible (\ie,
seamless) patches from elsewhere in the tile.  While more diverse,
these methods still rely on verbatim copying from the same tile.
Rodriguez-Pardo and Garces~\shortcite{Rodriguez:2022:SSS} specialize a
Generative Adversarial Network (GAN) architecture to synthesize a
single tileable texture.
Fr\"uhst\"uck~\etal~\shortcite{Fruhstuck:2019:TSL} tile, based on a
guidance map, the activations of an intermediate layer of a GAN
trained to synthesize tiles for a particular texture class (\eg, a
terrain map). The resulting tiled latent field is then processed by
the remainder of the GAN to produce a seam-free final texture.
Zhou~\etal~\shortcite{Zhou:2022:TCM} condition a GAN on a template to
control the tileable structure.  All three prior learning-based tiling
methods employ a GAN architecture which often needs to be retrained
for new texture categories.  We circumvent the problem of retraining
(and thus gathering a sufficiently large and diverse set of texture
exemplars) by leveraging existing pretrained text-to-image diffusion
models.  Furthermore, using a text-to-image diffusion model also
allows the user the specify the texture with a text-prompt instead of
providing an appropriate exemplar and/or structure-template.  Finally,
we revisit Wang tiles in the context of learning-based texture
synthesis, which to the best of our knowledge has not yet been
explored.

\paragraph{Wang Tiles}
Wang tiles~\shortcite{Wang:1961:PTP} are squares with colored edges
that can tile the 2D plane by adjoining tiles with matching colored
edges.  Cohen~\etal~\shortcite{Cohen:2003:WTI} introduced a
patch-based method for synthesizing textured tiles that meet the Wang
tile matching rules to enable fast synthesis of large
textures. Wei~\shortcite{Wei:2004:TBT} showed that, once the texture
tiles are precomputed, the tiling process can be directly implemented
on graphics hardware.  Furthermore, Wei showed that the Wang tiling
can be evaluated on the fly without the need to synthesize the full
tiling. Other graphics applications of Wang tiles include blue noise
generation~\cite{Lagae:2005:POD,Kopf:2006:RWT},
fabrication~\cite{Liu:2022:FMS}, and texturing an arbitrary 3D
shape~\cite{Fu:2005:TTA}.  Texture synthesis with edge-colored Wang
tiles suffers from the aptly named \emph{corner-problem} where the
corners of each texture tile are the same for all tiles thereby
creating a noticeable repetition in the tiled textures.  The
corner-problem is inherent to edge-colored Wang tiles because
diagonally neighboring tiles are not directly constrained, and
therefore each corner must match all other possible corners.  Corner
Wang tiles~\cite{Lagae:2005:POD,Ng:2005:GWT} overcome this issue by
matching colored corners instead of colored edges.  We also leverage
Wang tiles to support fast synthesis without the need to synthesize
the whole 2D plane.  However, unlike the above methods, we do not
require an exemplar sample, but instead allow the user to specify the
texture via a text-prompt and directly synthesize the different Wang
tiles.  In addition, we introduce a novel Dual Wang tile variant that
increases diversity and that does not suffer from the
corner-problem. Finally, due to the flexibility of our content-aware
tile generation, our tiles show greater diversity than prior graph-cut
generated tile textures.

\paragraph{Generative Diffusion Models}
Diffusion models formulate the generative process of a signal as an
iterative neural denoising
process~\cite{Song:2021:SBG,Karras:2022:EDS} outperforming the
state-of-the art in image synthesis
tasks~\cite{Dhariwal:2021:DMB}. When conditioned on
text-prompts~\cite{Nichol:2022:GTP,Ramesh:2022:HTC,Rombach:2022:HRI,Saharia:2022:PTI},
diffusion models enable non-artists to concretize their mental images.
Diffusion models have been successfully applied to a wide variety of
downstream tasks, including text-based image
editing~\cite{Kawar:2023:MTR,Kim:2022:DTG,Liu:2020:OEO,Mokady:2023:NTI,Tumanyan:2023:PPD},
sketch and depth-based
synthesis~\cite{Zhang:2023:ACC,Voynov:2023:SGT,Ye:2023:IAT,Ham:2023:MPD,Subrtova:2023:DIA},
and appearance capture~\cite{Sartor:2023:MGD,Vecchio:2023:CAC}.

A related class of prior work are methods that leverage diffusion
models to directly synthesize textures on 3D
shapes~\cite{Richardson:2023:TTG,Xu:2023:MMT,Chen:2023:FDG,Xiang:2023:3IG,Zeng:2023:PPA,Liu:2023:Z1T}. However,
these methods generate a fixed texture of finite size.  In contrast,
we leverage diffusion models to generate textured tiles suitable for
real-time synthesis of, potentially, infinite textures.  Similar to
prior work, we rely on powerful pretrained text-to-image diffusion
models, without fine-tuning or retraining, to sample the space of
textures.

Most related to our method are tileable diffusion variants.
Vecchio~\etal~\shortcite{Vecchio:2023:CAC} introduced a method to
generate tileable SVBRDFs (\ie, a $10$-channel image) using
noise-rolling. Noise-rolling \emph{``rolls''} the noise tensor by a
random translation each diffusion step, thereby placing the seam at a
random location. The diffusion models subsequently attempts to remove
this seam as it is not a natural feature.  Furthermore, to condition
noise-rolling on a non-tileable exemplar, Vecchio~\etal~\ introduce a
conditional noise rolling variant that masks a small region around the
\emph{input} border ($1/16$ of the image size) allowing the diffusion
model to seamlessly 'inpaint' the missing
texels. Rodriguez-Pardo~\etal~\shortcite{Rodriguez-Pardo:2024:TAD}
introduce \emph{``TexTile''}, a differentiable tileability metric.
TexTile can be leveraged to force a diffusion model (\ie,
SinFusion~\cite{Nikankin:2023:STD}) to produce self-tiling images by
interleaving each diffusion step with a TexTile optimization step.  It
is unclear how either noise-rolling or TexTile can be extended beyond
the generation of a self-tiling image. In contrast, we demonstrate
that our content-aware tile generation method is also applicable to
more general tiling schemes such as Wang tiles.

\section{Exterior Boundary Inpainting}
\label{sec:method}

Our goal is to generate a small set of one or more mutually tileable
images from a text-prompt and optionally an exemplar image. When
provided, the content of the optional exemplar and text-prompt should
match.  Alternatively, the exemplar image can also be generated from
the text-prompt. In our implementation, we use
\emph{Stable-Diffusion-XL}~\cite{StableDiffusionXL} to generate the
exemplars from a text-prompt and apply (unconditional)
noise-rolling~\cite{Vecchio:2023:CAC} to obtain a more texture-like
exemplar; a similar result can be obtained with appropriate prompt
engineering.

We introduce our method starting with the simplest tiling
configuration (\ie, a self-tiling texture), and then
demonstrate our method's flexibility by applying the same methodology
to more complex tiling schemes, culminating in a novel Dual Wang
tiling.

\paragraph{Self-tiling Texture}
We start with the most straightforward tileable texture (a self-tiling
texture that does not introduce visible seams when tiled over a 2D
plane) as a didactic example to explain the core of our method;
similar results can also be obtained with prior self-tiling
methods~\cite{Vecchio:2023:CAC,Rodriguez-Pardo:2024:TAD}.

\begin{figure}[t!]
    \includegraphics[height=0.3\columnwidth]{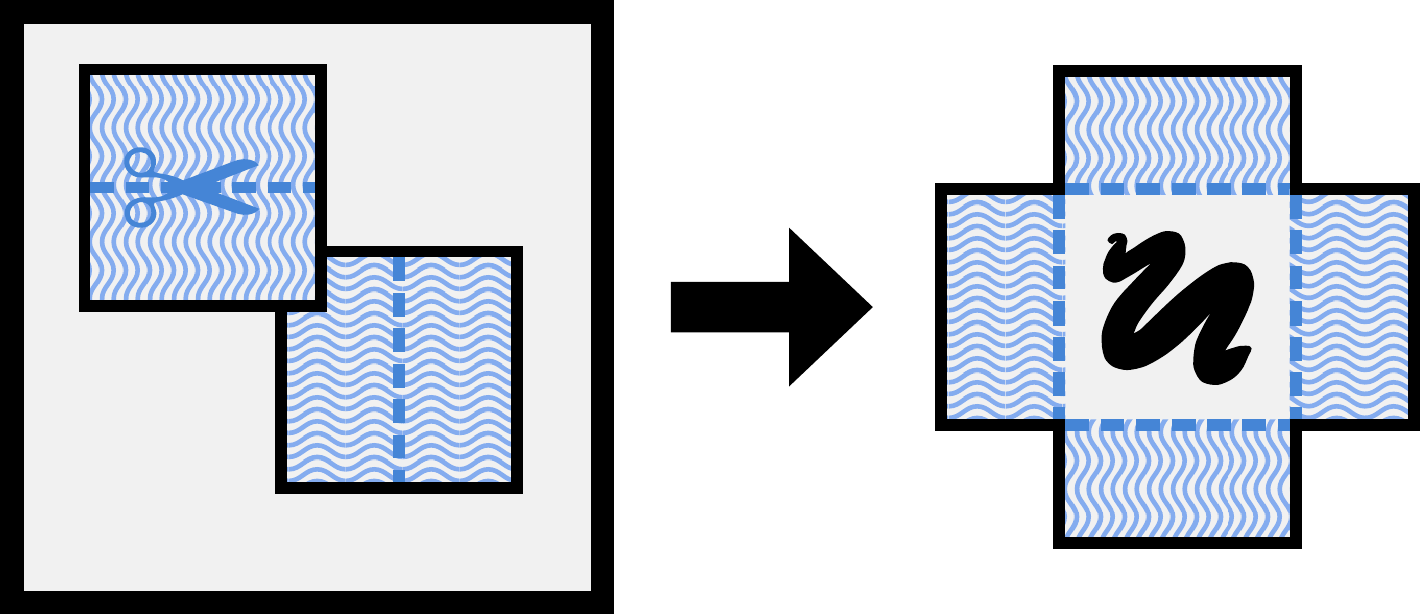}
    \caption{\textbf{Self-tiling Texture Generation:} We establish
      contiguous horizontal and vertical boundary conditions by
      selecting two template patches from the exemplar, and cutting
      them in half horizontally and vertically respectively.  Each cut
      half is placed on the outside of the tile with the cut edge
      (dashed line) aligned with a tile edge. The interior of the tile
      (scribbled area) is then inpainted. The final tile is then
      cropped to only retain the synthesized part.}
    \label{fig:singletile}
\end{figure}

\begin{figure*}
    \centering
    \begingroup
    \setlength{\fboxrule}{2pt}
    \setlength{\fboxsep}{0pt}
    \setlength{\lineskip}{2pt}
    \fbox{\includegraphics[width=0.5\textwidth-5pt]{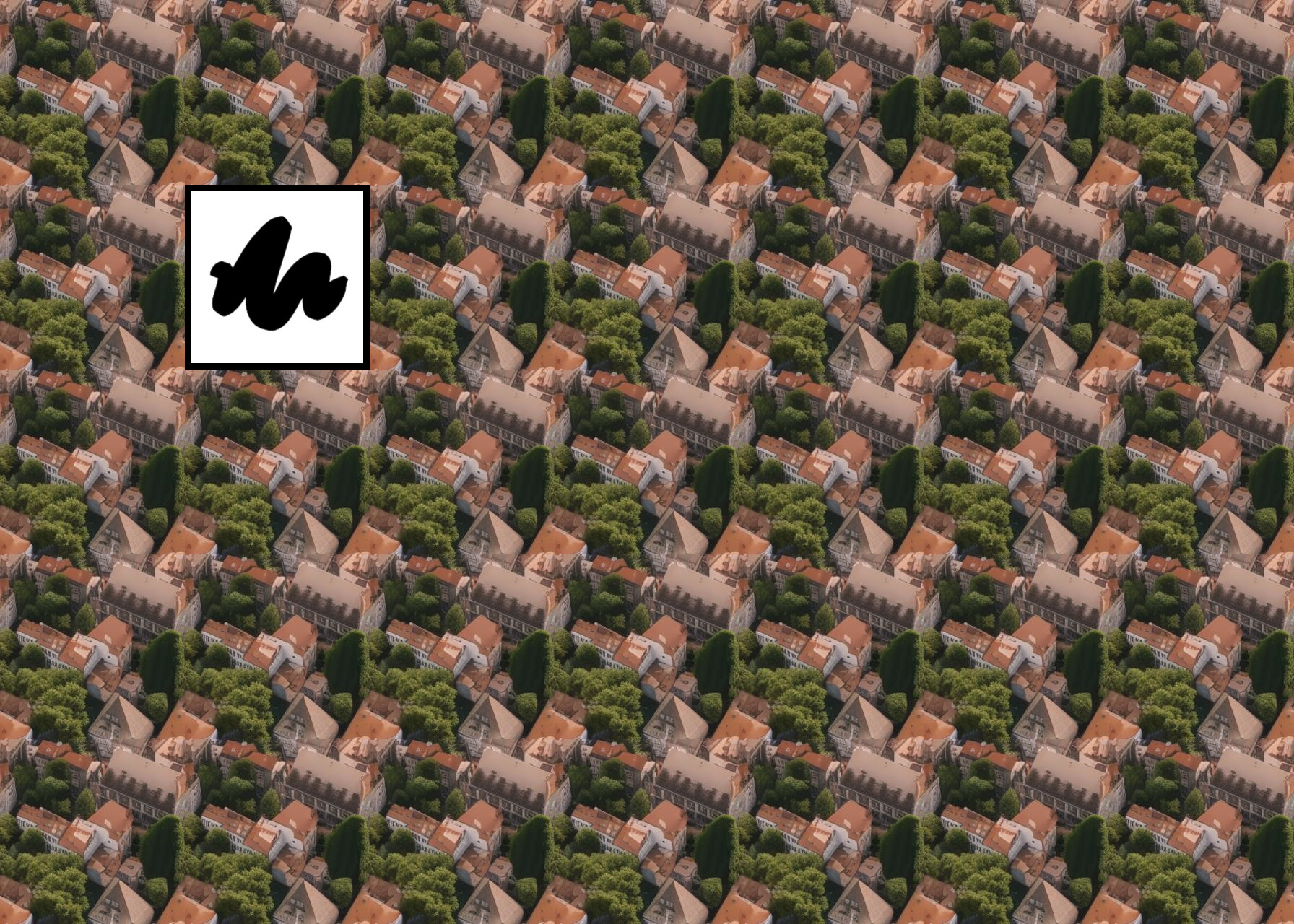}}
    \hfill
    \fbox{\includegraphics[width=0.5\textwidth-5pt]{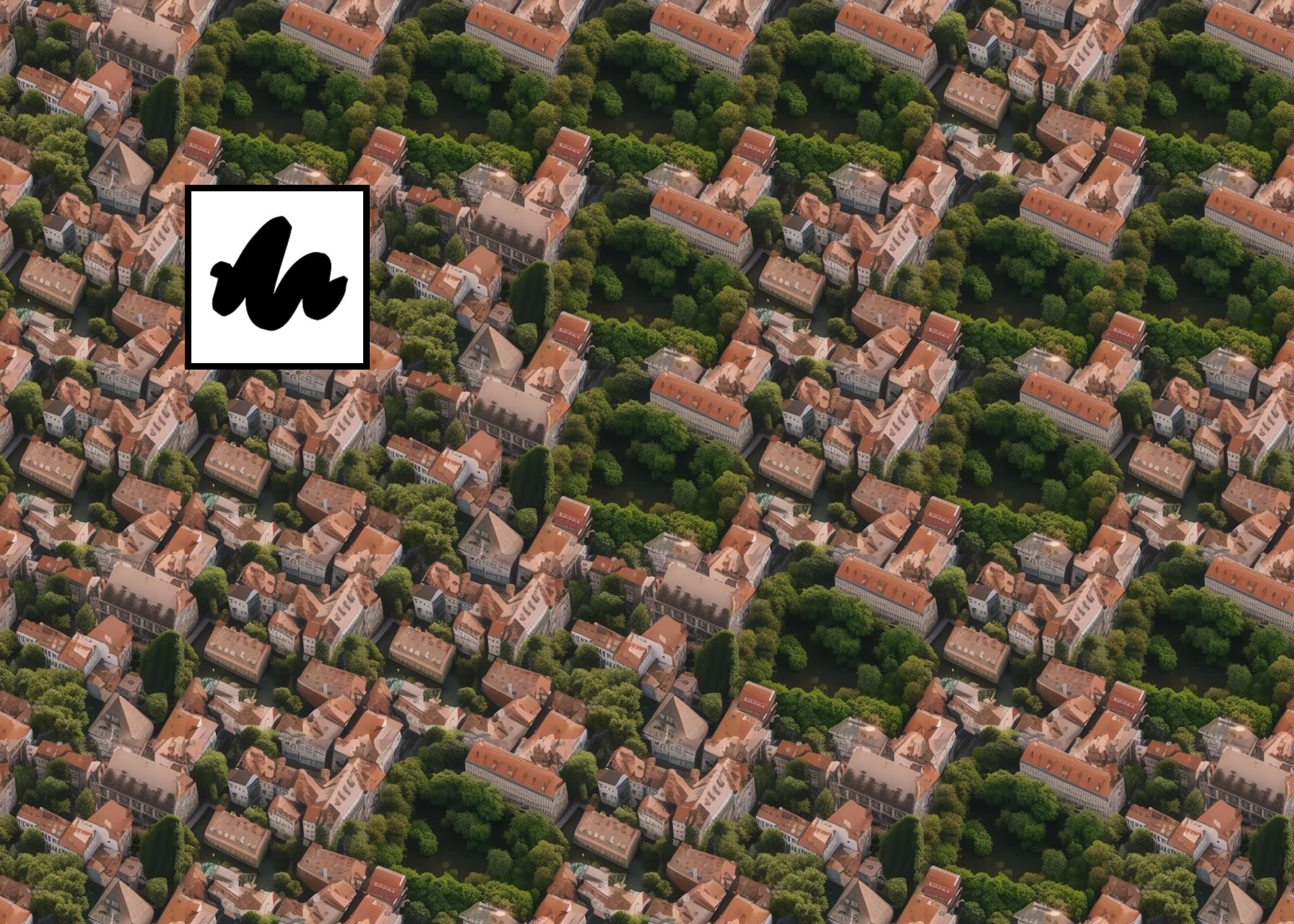}} \\
    \fbox{\includegraphics[width=0.5\textwidth-5pt]{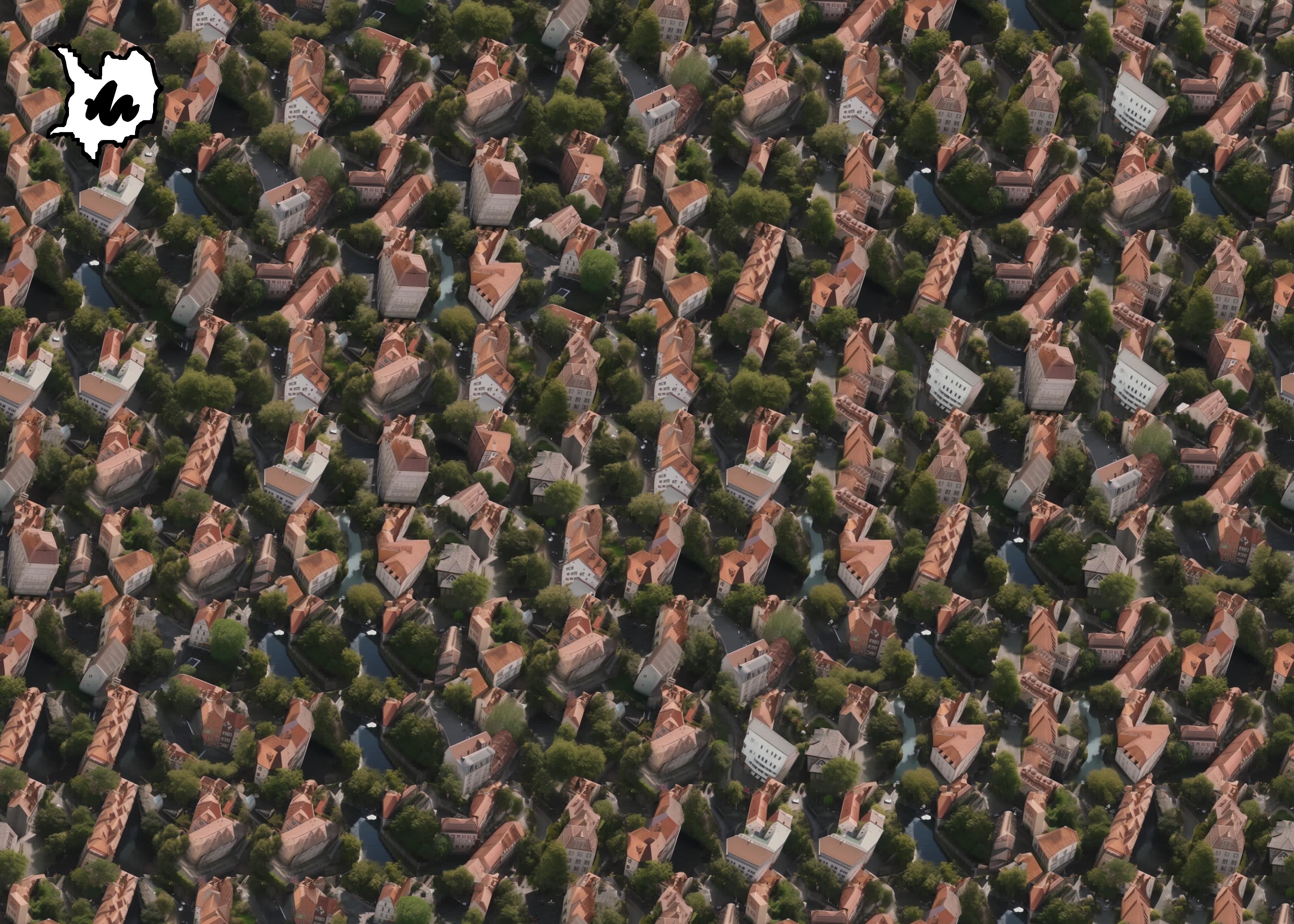}}
    \hfill
    \fbox{\includegraphics[width=0.5\textwidth-5pt]{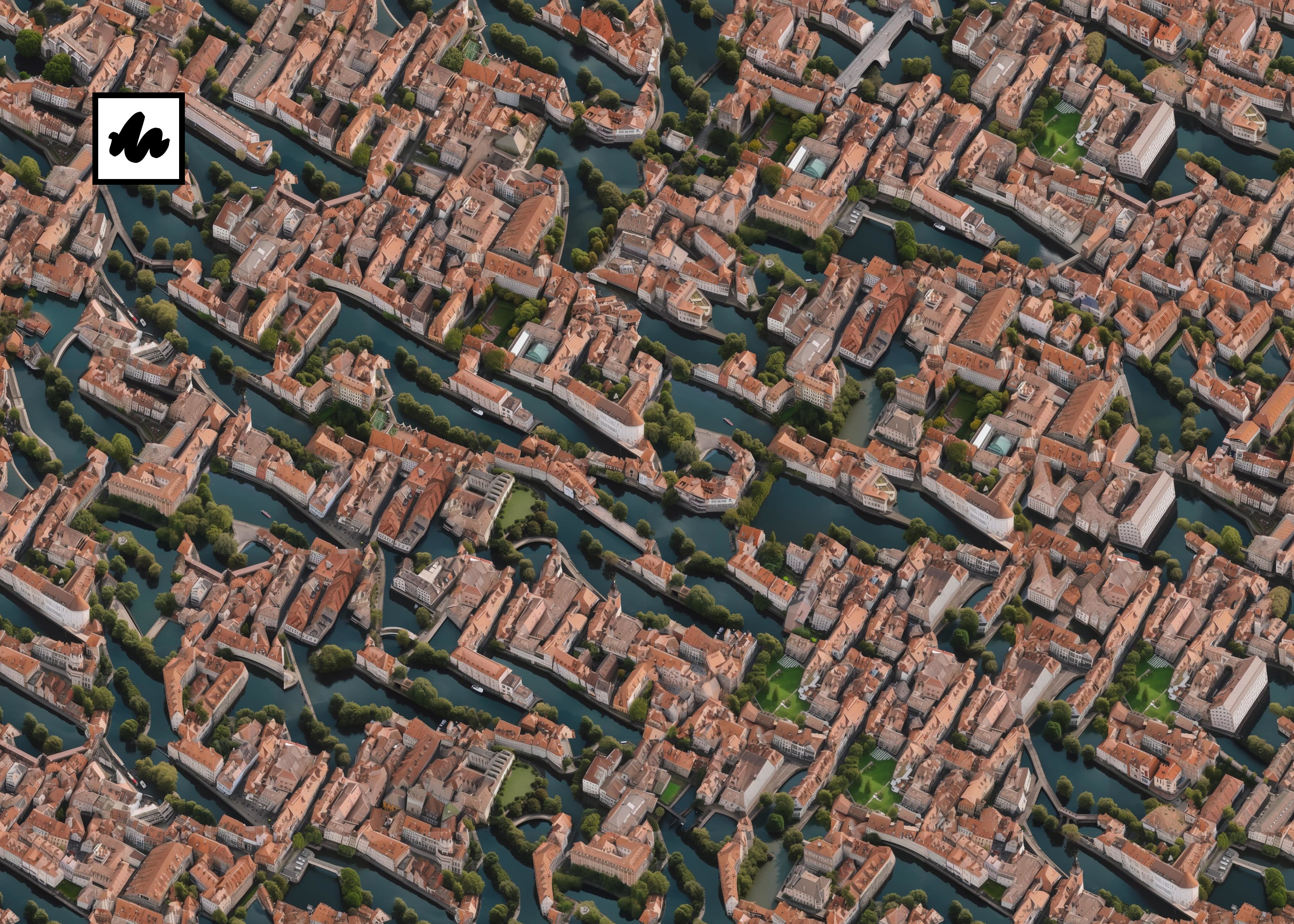}} \\
    \fbox{\includegraphics[width=\textwidth-4pt]{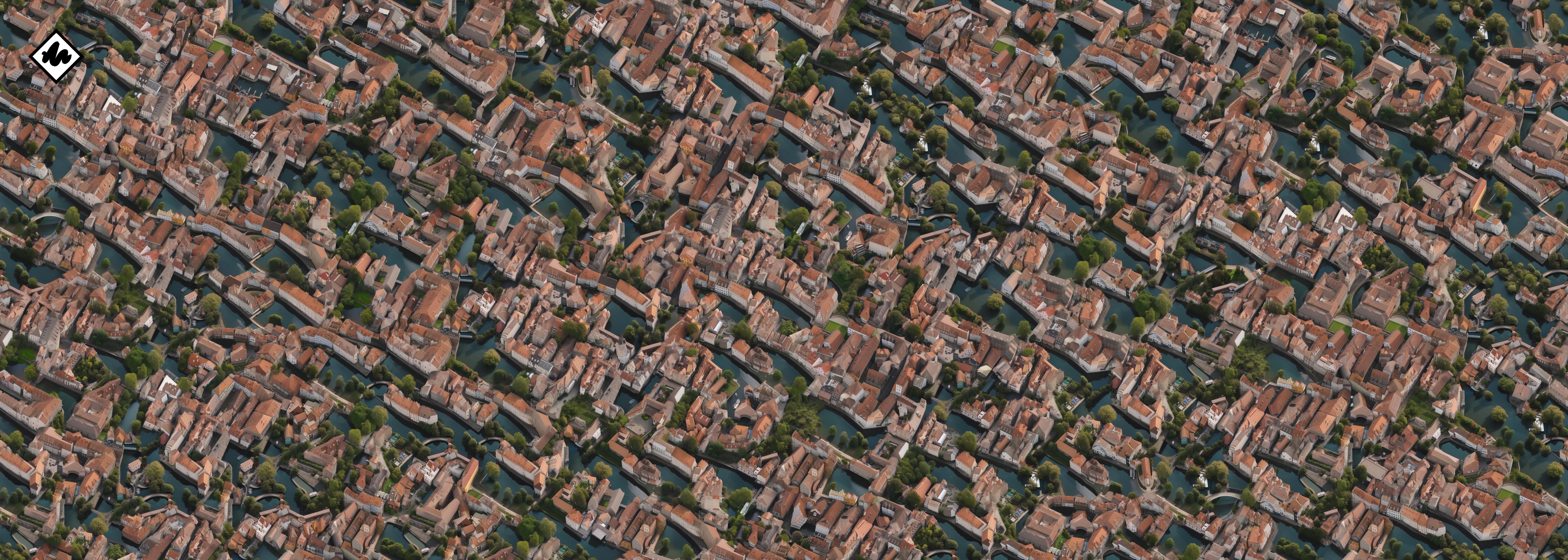}} \\
    \endgroup
    \caption{\textbf{Comparison of Tiling Schemes:} A comparison of
      the different tiling schemes (first row: self-tiling and
      stochastic self-tiling (with $4$ tiles); second row: stochastic
      Escher self-tiling (with $4$ tiles) and regular $3$-color Wang
      tiling ($81$ tiles); last row: $3$-color Dual Wang tiling)
      demonstrated on texture tiles generated with the prompt
      ``\textit{European city blocks, tile roofs, streams, drone
        footage}''. For each example we also show the tile shape and
      size with the scribbled overlay.}.
    \label{fig:comparison}
\end{figure*}

To promote diversity, we aim to fully synthesize the interior tile
without verbatim copying of parts from the exemplar image. Instead, we
will leverage the exemplar to establish exterior boundary conditions
that constrain the synthesis of the image to be tileable.  To
synthesize the self-tiling texture, we employ an existing
prompt-conditioned diffusion-based inpainting model
(\emph{Stable-Diffusion-2-Inpainting}~\cite{StableDiffusionInpainting}).
The role of the prompt is to constrain the content, whereas the
boundary conditions define the structure near the tile edges.
Inpainting methods require a wide enough strip of example pixels
surrounding the target region in order to guarantee continuity and a
consistent structure.  At the same time, to ensure continuity at
matching edges, the combined boundary strips at matching edges need to
be contiguous and not exhibit any seams.  We can fulfill both goals,
by selecting a template patch from the exemplar (with a size similar
to the target tile size) for each pair of matching edges. In the case
of a self-tiling image, we have two pairs of matching edges: the
horizontal and vertical edge pairs. Next, we cut the template patches
in half horizontally and vertically respectively, and copy each half
template patch to the \emph{outside} of the tile with the cut-edge
abutting the tile edges acting as the exterior boundary conditions.
By construction, these boundary conditions will be contiguous across
matching edges.  Finally, we generate the interior of the tile by
inpainting. We only retain the synthesized interior as the final tile
such that no pixels from the exemplar image end up in the final tile
texture. Note, that we also inpaint the corners of the image not
covered by the template halves; this helps in creating reasonable
content for the corners of the tile.  \autoref{fig:singletile}
illustrates the process, and~\autoref{fig:comparison} (top row, 1st
column) shows an example of a tiled texture.

\begin{figure}[t!]
    \includegraphics[width=.95\columnwidth]{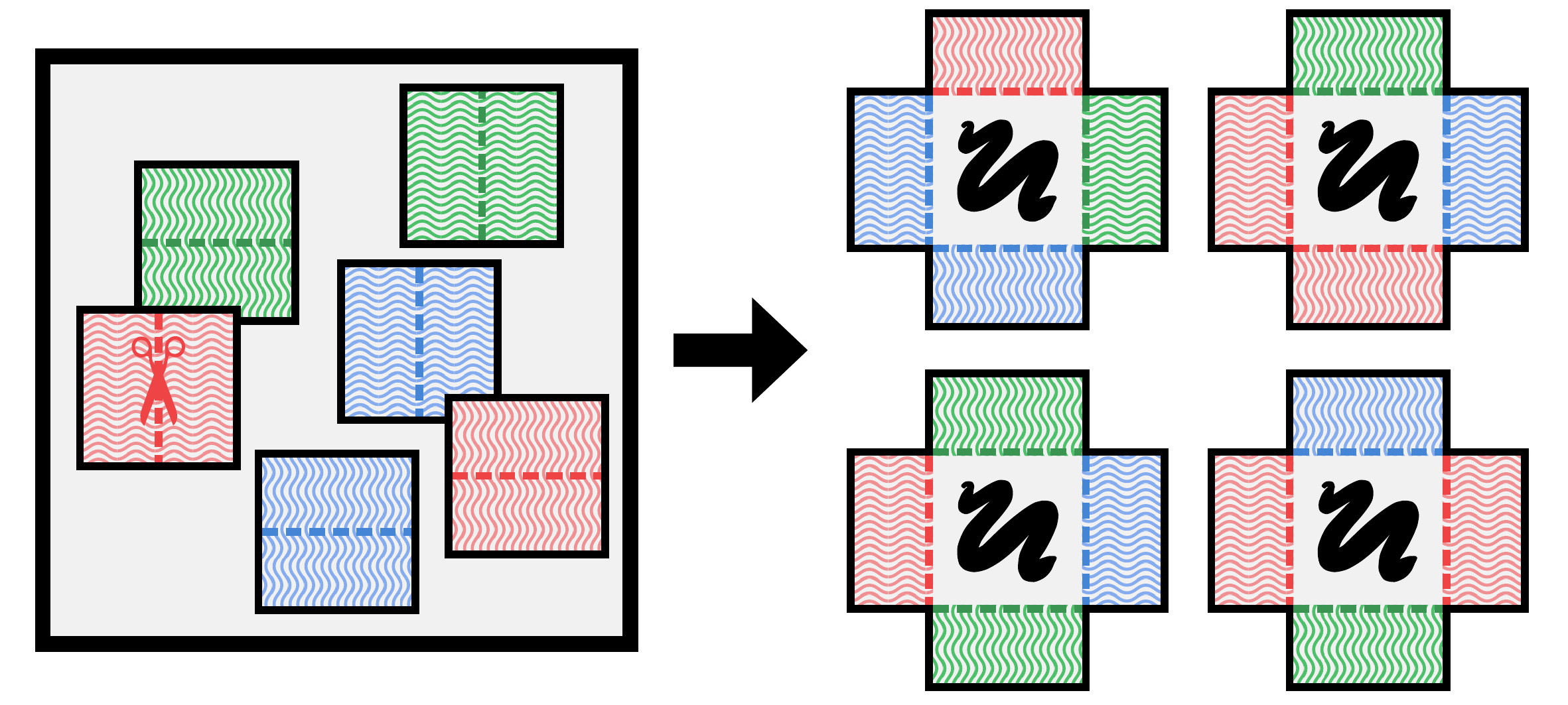}
    \caption{\textbf{Wang Tile Generation:} Given $C$ colors, we
      select $2 \times C$ template patches from an exemplar image, and
      cut each template patch in half (a horizontal and a vertical cut
      per color, marked by the dashed line). For each Wang tile we
      set the boundary conditions by coping each half template patch
      to the exterior of the Wang tile with the (dashed) edge matching
      the corresponding Wang tile edge.  Finally, we generate the Wang
      tile texture by inpainting the interior region (\ie, the
      scribbled area).}
    \label{fig:wang}
\end{figure}

\paragraph{Stochastic Self-tiling Texture}
Self-tiling textures often produce visibly repeating patterns, and the
resulting texture is visually not very diverse.  To enrich the tiling,
we leverage that diffusion-based inpainting takes, besides an image
and a mask indicating the area to be inpainted, also a seed as input.
Changing the seed will result in a slightly different generated
texture.  Keeping the boundary conditions (\ie, template patches) the
same, allows us to generate multiple textured tiles that are mutually
tileable.  This yields a simple tiling algorithm where we randomly
select, during tiling, which texture tile to use, yielding a more
diverse tiled texture.  \autoref{fig:comparison} (top row, 2nd column)
shows an example of such a stochastic self-tiling texture.

\paragraph{Escher Tiles}
Salient features that cross a tile edge impose stricter constraints on
the tile generation process. While we could select different boundary
conditions, it is not always possible to find straight boundary
conditions without salient features.  However, we observe that
diffusion-based inpainting imposes no constraints on the shape of the
boundaries.  Hence, we can trivially generate non-square
\emph{``Escher''} tiles by cutting each template patch along an
arbitrary path.  \autoref{fig:comparison} (middle row, 1st column)
shows an example of such a self-tiling Escher texture.

\paragraph{Wang Tiles}
The above stochastic tiles are good for textures with isolated objects
and simple boundaries (\eg, shells in sand). However, if the features
at the tile boundaries are distinct, the resulting texture will
exhibit a clearly visible
repetition. Cohen~\etal~\shortcite{Cohen:2003:WTI} proposed to use
textured Wang tiles to produce tiled textures with greater diversity.
We can apply a similar method for generating Wang tiles as the
previous tiling variants.  However, unlike the self-tiling case, we
select two template patches (for horizontal and vertical splitting
respectively) per Wang tile edge color ($= 2C$ template patches for
$C$ colors).  For each of the $C^4$ Wang tiles, we generate the
texture by copying the appropriate half template patches (based on the
edge color) as an exterior boundary condition, and inpaint the
interior region as before
(\autoref{fig:wang}). \autoref{fig:comparison} (middle row, 2nd
column) shows an example of a titled texture generated with a
$3$-color textured Wang tile set ($= 81$ tiles).  An advantage of our
exterior boundary inpainting strategy is that our textured Wang tiles
will exhibit high diversity since each tile only shares the boundary
conditions and the interior of each textured tile is inpainted
separately.  In contrast, the graph-cut based method of
Cohen~\etal~reuses the same diamond templates, and thus the same
texture content appears in multiple textured tiles.  Similar as in
stochastic self-tiling, we can also further increase diversity by
generating multiple textures per Wang tile and randomly selecting one
at tiling-time (a similar strategy was also proposed by
Cohen~\etal~\shortcite{Cohen:2003:WTI}).

\begin{figure}
    \includegraphics[width=\columnwidth]{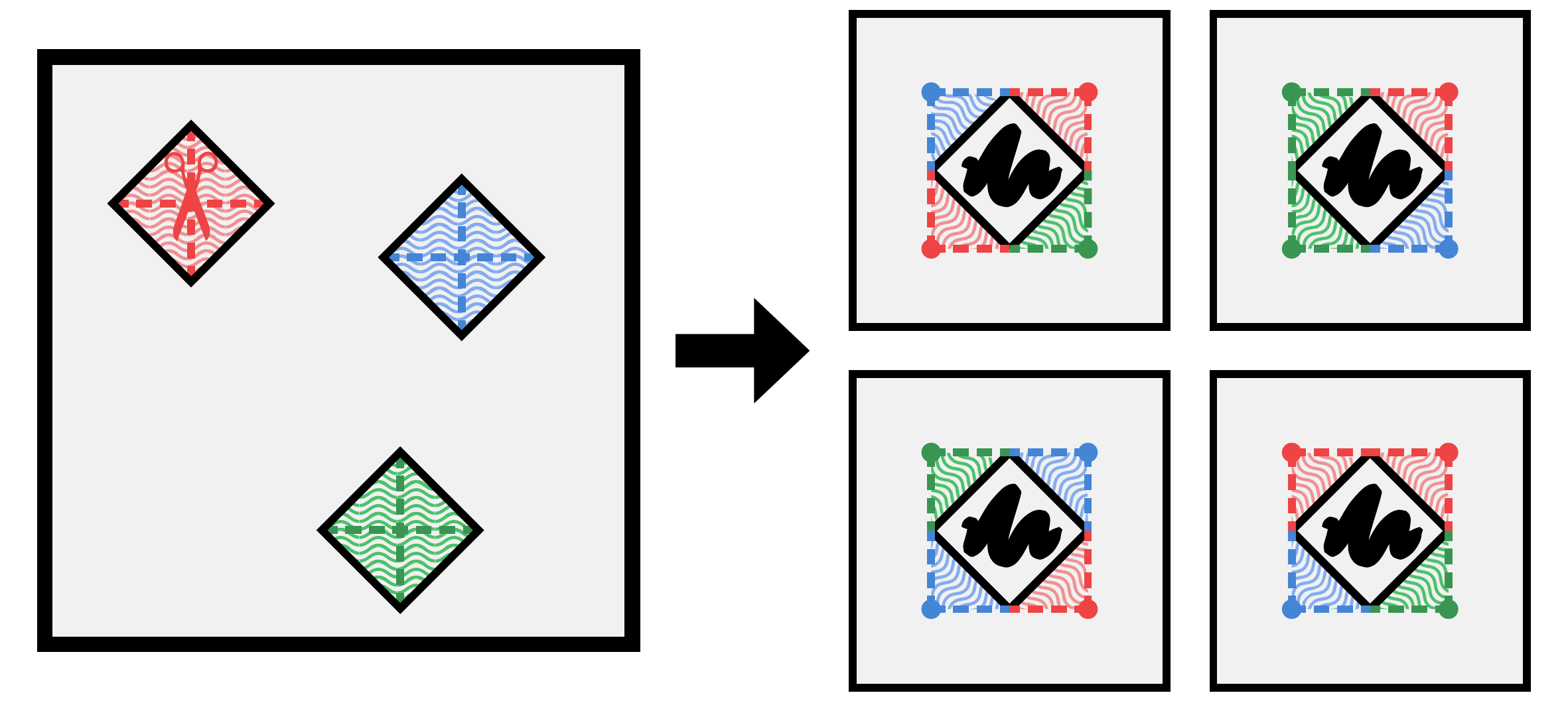}
    \caption{\textbf{Corner Wang Generation:} We follow a very similar
      generation process as Lagae and
      Dutr\'e~\shortcite{Lagae:2005:POD}.  Instead of using square
      template patches, we select $C$ diamond shaped template patches,
      and cut them both horizontally and vertically. Next, we copy the
      triangular patches into the corner tiles based on corner colors,
      and inpaint the interior. Unlike the other tiling schemes,
      Corner Wang tiles include (parts of) the template patches in the
      final tiled texture.}
    \label{fig:corner}
\end{figure}

\begin{figure}[t!]
  \setlength{\fboxrule}{2pt}
  \setlength{\fboxsep}{0pt}
  \fbox{\includegraphics[width=\columnwidth-4pt]{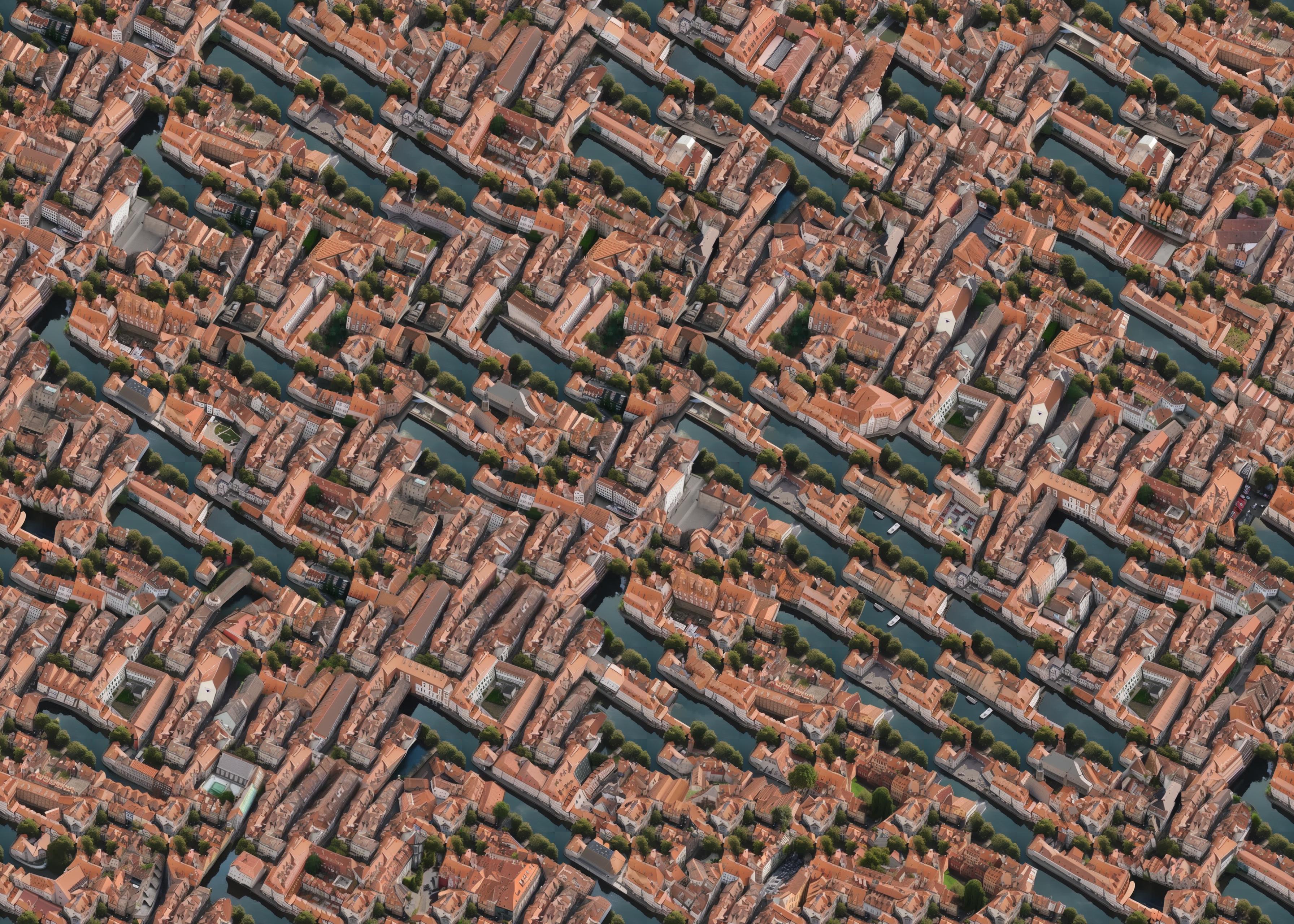}}
  \caption{\textbf{Content-aware Corner Wang Tiling:} An example of a
    Corner Wang tiling~\cite{Lagae:2005:POD} generated with an
    adaptation of our content-aware tile synthesis method.  Unlike
    regular Wang tiles and Dual Corner Wang tiles, the resulting tiles
    contain verbatim copied texels from the exemplar image, resulting
    in a less diverse tiling.}
  \label{fig:corner_example}
\end{figure}

\begin{figure*}
    \includegraphics[width=.99\textwidth]{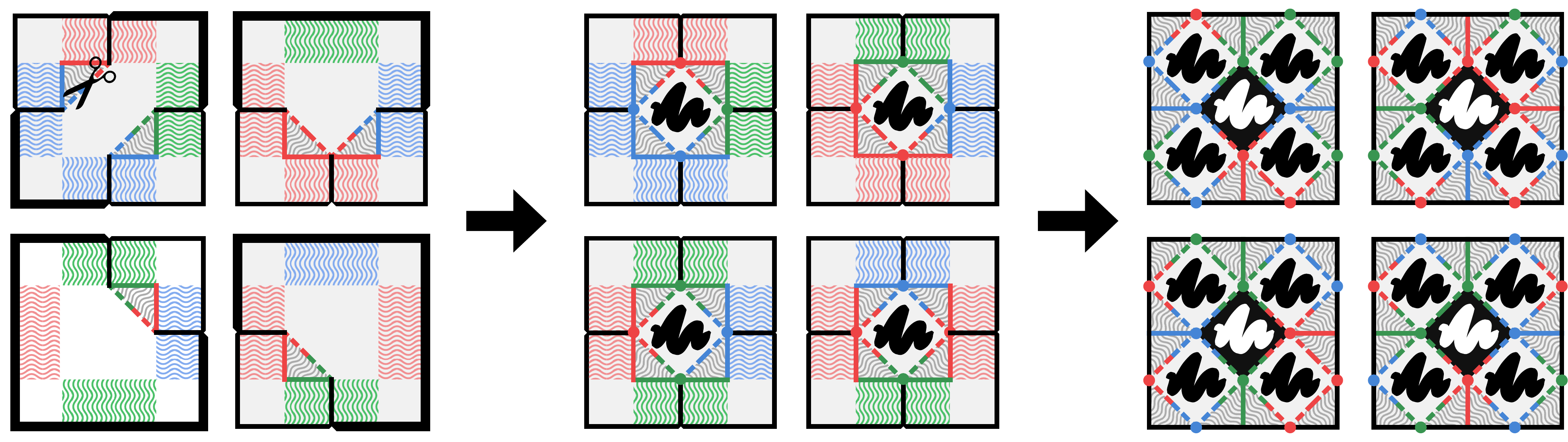}
    \caption{\textbf{Dual Wang Tiles} are created in a three-stage
      process. Left: we first generate a subset of $4 \times C^2$
      regular Wang tiles, and extract the boundary conditions for each
      interior tile edge (\ie, $4$ edges, each identified by $2$
      colors or $\C^2$ combinations).  Middle: we assemble the
      generated boundary conditions corresponding to each interior
      tile's color combinations, and inpaint and extract the central
      interior tile texture (black scribble).  Right: the cross
      texture tiles (white scribble) are generated by inpainting a
      $2 \times 2$ tiling of the plane with the interior tiles as
      boundary conditions.}
    \label{fig:dual}
\end{figure*}

\paragraph{Corner Wang Tiles}
Wang tiles suffer from the so-called \emph{``corner
  problem''}~\cite{Cohen:2003:WTI,Lagae:2005:POD}, \ie, each corner
texel is shared between all tiles because the tiles placed diagonally
across a corner do not share any edges and thus each corner needs to
match to the opposing corner of \emph{any} other tile.  Lagae and
Dutr\'e~\shortcite{Lagae:2005:POD} introduced Corner Wang tiles as a
solution Corner Wang tiles associate a color with each corner instead
of an edge of the tile.

We can also adapt our content-aware tile generation procedure to
Corner Wang tiles (\autoref{fig:corner}) following a similar process
as Lagae and Dutr\'e~\shortcite{Lagae:2005:POD} but with the graph-cut
based generation of the interior diamond replaced with inpainting. In
short, instead of selecting $2C$ square template patches as for Wang
tiles, we select $C$ diamond shaped template patches.  Each template
is cut in four parts by splitting the diamond horizontally and
vertically.  Each triangular template patch is then copied to the
matching corners of the Corner tiles and the interior diamond is
inpainted.  This process, however, breaks the cardinal rule that we do
not verbatim copy patches from the exemplar texture.

\autoref{fig:corner_example} shows an example of a content-aware
generated Corner Wang tiling.

\paragraph{Dual Wang Tiles}
While Corner Wang tiles avoid the corner problem, it also
introduces two new practical problems.  First, there is a difference
in diversity between the corners (copied from the exemplar) and the
center (synthesized).  Second, as observed by Lagae and
Dutr\'e~\shortcite{Lagae:2005:POD}, discontinuities often occur close
to the center of each tile edge as these regions are weakly
constrained by the surrounding copied patches.

We introduce a new Dual Wang tile variant to solve both problems.  The
key idea is to make the content of the corner regions of a Wang tile
dependent on its neighboring tiles.  We start from a regular Wang
tiling.  However, instead of attaching a square texture to each tile,
we attach a diamond-shaped texture (each diamond's corner touches the
center of each Wang tile edge). We call these diamond shaped tex-
\setlength{\intextsep}{0mm}%
\setlength{\columnsep}{2mm}%
\begin{wrapfigure}{r}{0.15\textwidth}
  \centering
  \includegraphics[width=0.14\textwidth]{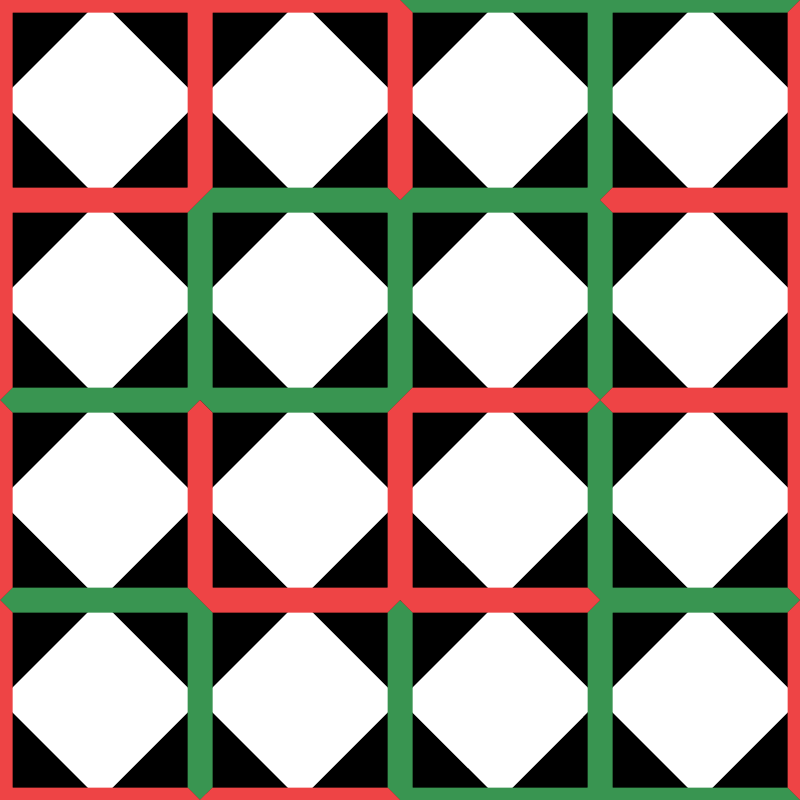}
\end{wrapfigure}

\!tures the \emph{``interior tiles''} (white diamonds in the inset).
Each interior tile is identified by the edge-colors of the surrounding
Wang tile, and thus there are $\C^4$ different interior tiles.  Tiling
a plane with such interior tiles results in a texture with diamond
shape holes between each four interior tiles that share a corner. We
employ a second set of diamond shaped textures, named \emph{``cross
  tiles''} (black diamonds in the inset), to seamless fill the
resulting hole. Each cross tile is identified by the colors of the
edges incident on the Wang tile-corner at the center of the cross
tile.  Hence, there also exist $\C^4$ possible cross tiles.  The
combined set of interior and cross tiles form the Dual Wang tile set.
Since every interior tile can be combined with different cross tiles
(depending on its neighboring tiles), both the interior and corner
regions are now equally diverse.

Our goal now is to create two sets of textured tiles, one for the
interior tiles and one for the cross tiles.  Each tile does not need
to seamlessly tile (along an edge) with another tile from the same
set, but it does need to tile with corresponding tiles from from the
other set.  Consequently, we can synthesize both sets separately; once
we have one set (\eg, interior tile set), we can use it as boundary
conditions for generating the second set.  Additionally, we want to
impose a corner-continuity-constraint at the (diamond) corners in both
sets in order to avoid discontinuities as in Corner Wang tiles.

We start with the observation that the boundary conditions used for
generating regular Wang tiles already meet the (diamond)
corner-continuity-constraint.  However, these are not sufficient
conditions for seamless Dual Wang tiles, as we further need to
constrain each interior/cross tile edge. Since each interior/cross
tile edge is identified by two colors, there are $\C^2$ possible
conditions per edge (or $4 \times \C^2$ in total).  Unlike the
corner-continuity-constraint, the edge constraints do not need to
enforce continuity between multiple sides of tiles from the same set;
it only needs to impose the same constraint for corresponding tile
edges in the other set.  Both constraints cannot be met by simply
cutting the boundary constraints from the exemplar.  Instead, we first
generate a regular Wang tile set, which by construction is contiguous
across the edges (with matching colors), and thus also across the
diamond-corners. It therefore forms a potential source for extracting
the Dual Wang tile boundary conditions. However, each interior/cross
edge-color combination occurs multiple times over the Wang tile set,
each with potentially different synthesized pixel values along the
diamond-tile edges.  We therefore select a subset of the Wang tile set
such that each combination only occurs once (\ie, a subset of
$4 \times \C^2$ Wang tiles). From this subset, we can then extract the
boundary conditions for generating the interior tiles
(\autoref{fig:dual}, left).  Note, because we want to retain
continuity across the interior/cross tile corners, we also include the
original boundary conditions for each Wang tile, yielding a
square-shaped exterior boundary condition with a triangular corner
(\ie, overlapping with the interior tile) cut out.

To generate the interior tiles, we assemble the Wang-tile-extracted
boundary conditions for each interior edge, and inpaint the interior
tiles (\autoref{fig:dual}, middle).  Once we have generated all
interior tiles, we can then synthesize the cross tiles by creating a
$2 \times 2$ tiling of interior tiles with corresponding colors which
we use as boundary conditions for inpainting the cross tiles
(\autoref{fig:dual}, right).

The generated Dual Wang tiles solve both of the identified issues with
Corner Wang tiles.  First, all tiles in both sets are fully generated,
and no part of the boundary conditions are included in the final
textured tile sets.  Therefore, no part of the textured tile set
occurs at a different frequency.  Second, the Wang-tile-extracted
boundary conditions provide boundary conditions across the diamond
boundaries, as well as across the diamond-tile corners.
\autoref{fig:comparison} (bottom row) shows an example of a Dual Wang
tiled texture.

\begin{figure*}
    \def\restrim{0.28}
    \centering
    \begingroup
    \setlength{\fboxrule}{2pt}
    \setlength{\fboxsep}{0pt}
    \setlength{\lineskip}{2pt}
    \textit{``woven basket closeup, sharp focus, wooden strips''}
    \fbox{\adjincludegraphics[width=\textwidth-2\fboxrule,trim={0 {\restrim\height} 0 0},clip]{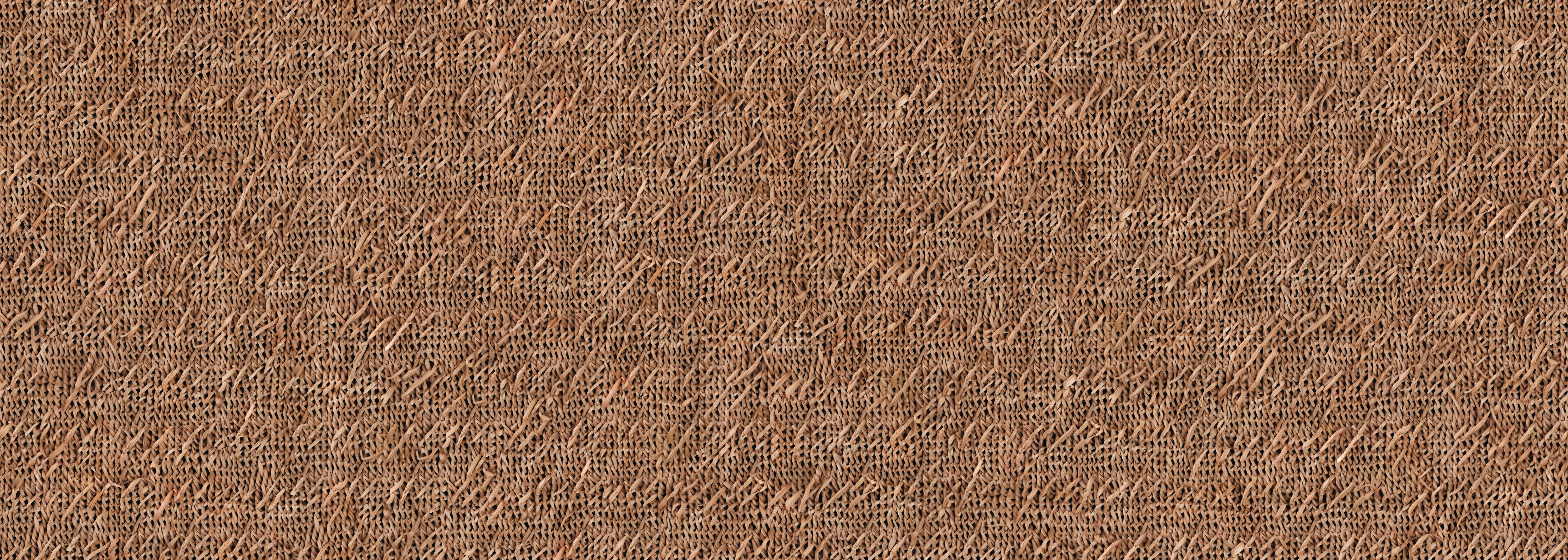}}
    \textit{``pebbles in a stream''}
    \fbox{\adjincludegraphics[width=\textwidth-2\fboxrule,trim={0 {\restrim\height} 0 0},clip]{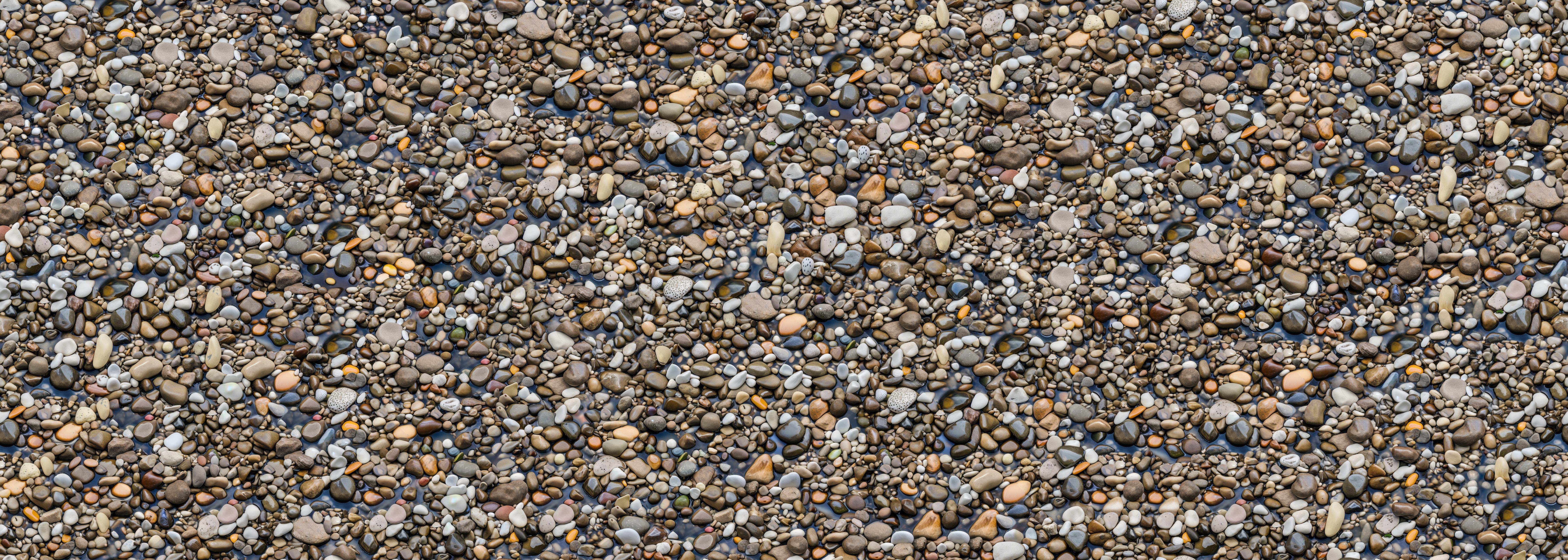}}
    \textit{``misty mountains illustration, pine trees''}
    \fbox{\adjincludegraphics[width=\textwidth-2\fboxrule,trim={0 {\restrim\height} 0 0},clip]{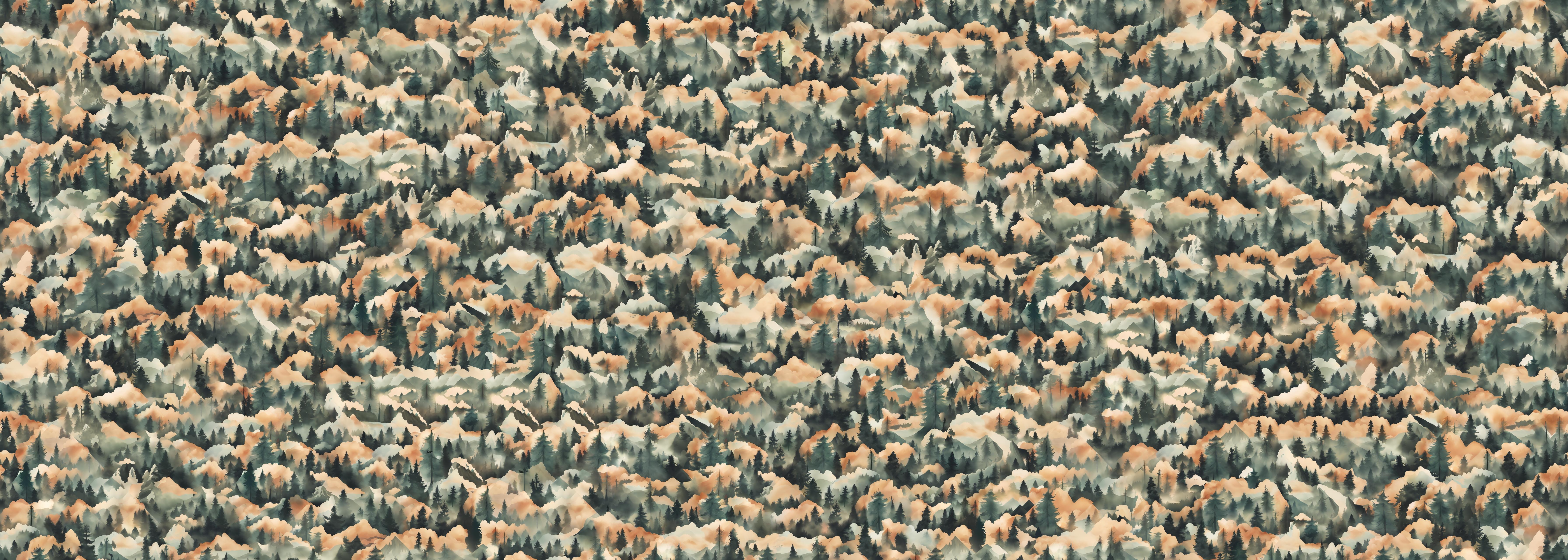}}
    \textit{``volcanic crevice, lava flow, bright magma, jagged rocks, small plants''}
    \fbox{\adjincludegraphics[width=\textwidth-2\fboxrule,trim={0 {\restrim\height} 0 0},clip]{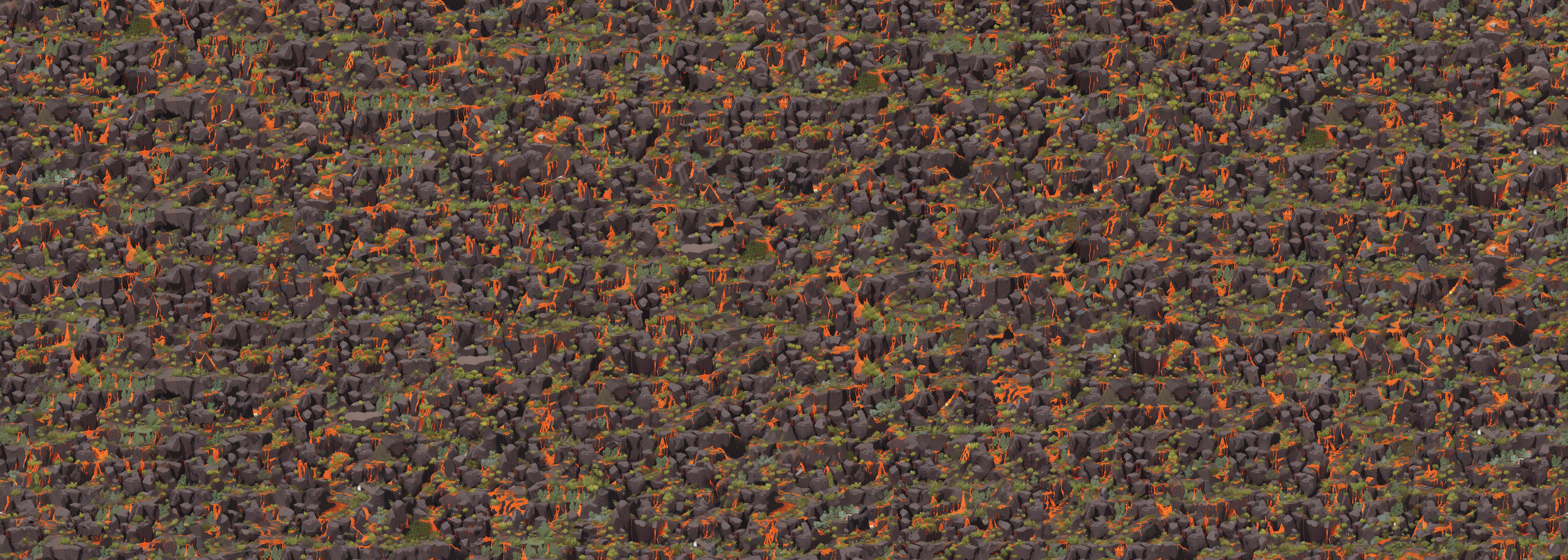}}
    \endgroup
    \caption{\textbf{Dual Wang Tile Results:} High resolution
      ($7168 \times 2560$) Dual Wang tiled textures generated from a
      text-prompt (listed above each example).  Please zoom-in on the
      tiled textures to fully appreciate the generated texture
      detail.}
    \label{fig:results}
\end{figure*}

\begin{figure*}[t!]
    \setlength{\fboxrule}{2pt}
    \setlength{\fboxsep}{0pt}
    \fbox{\includegraphics[width=0.24\linewidth-5pt]{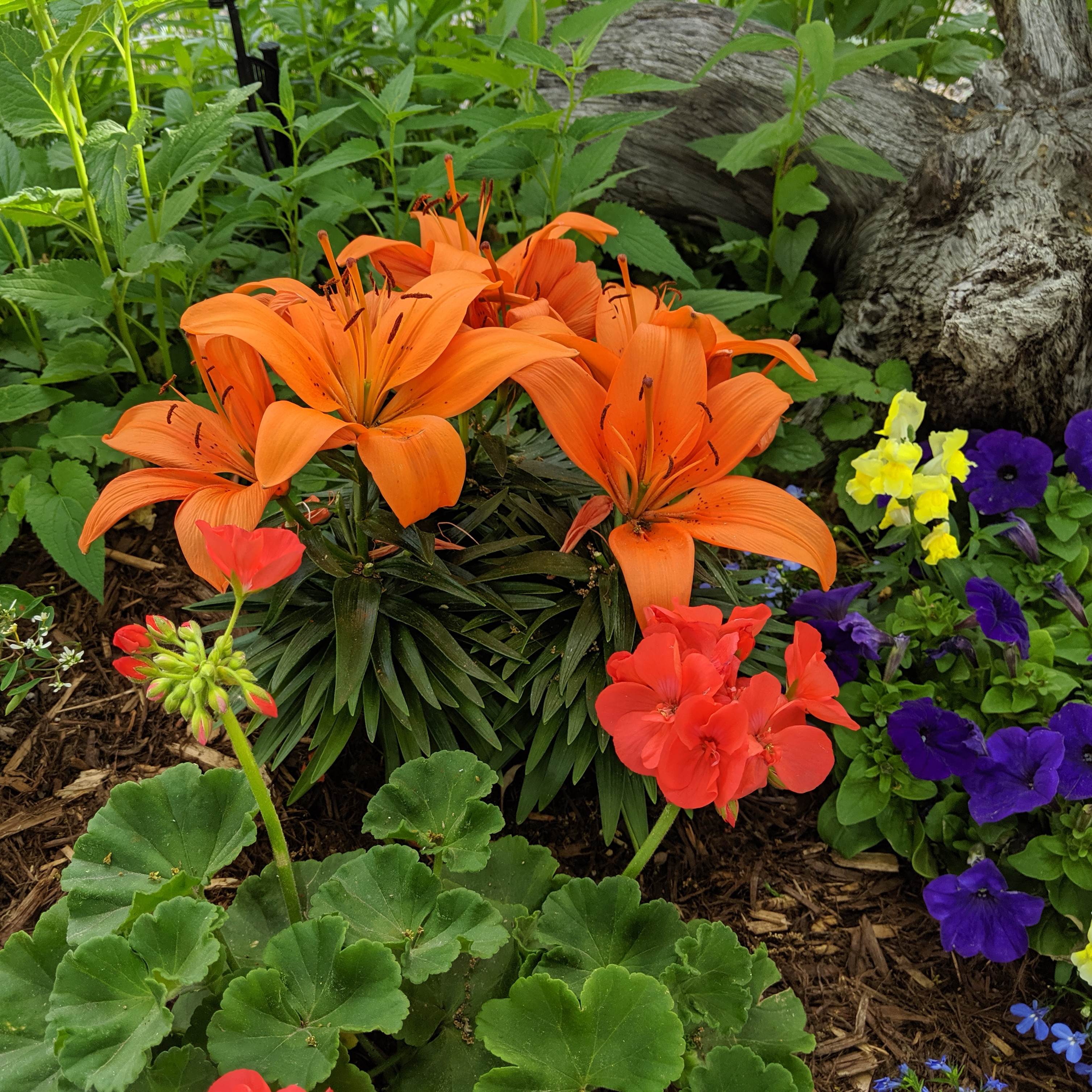}}
    \fbox{\includegraphics[width=0.24\linewidth-5pt]{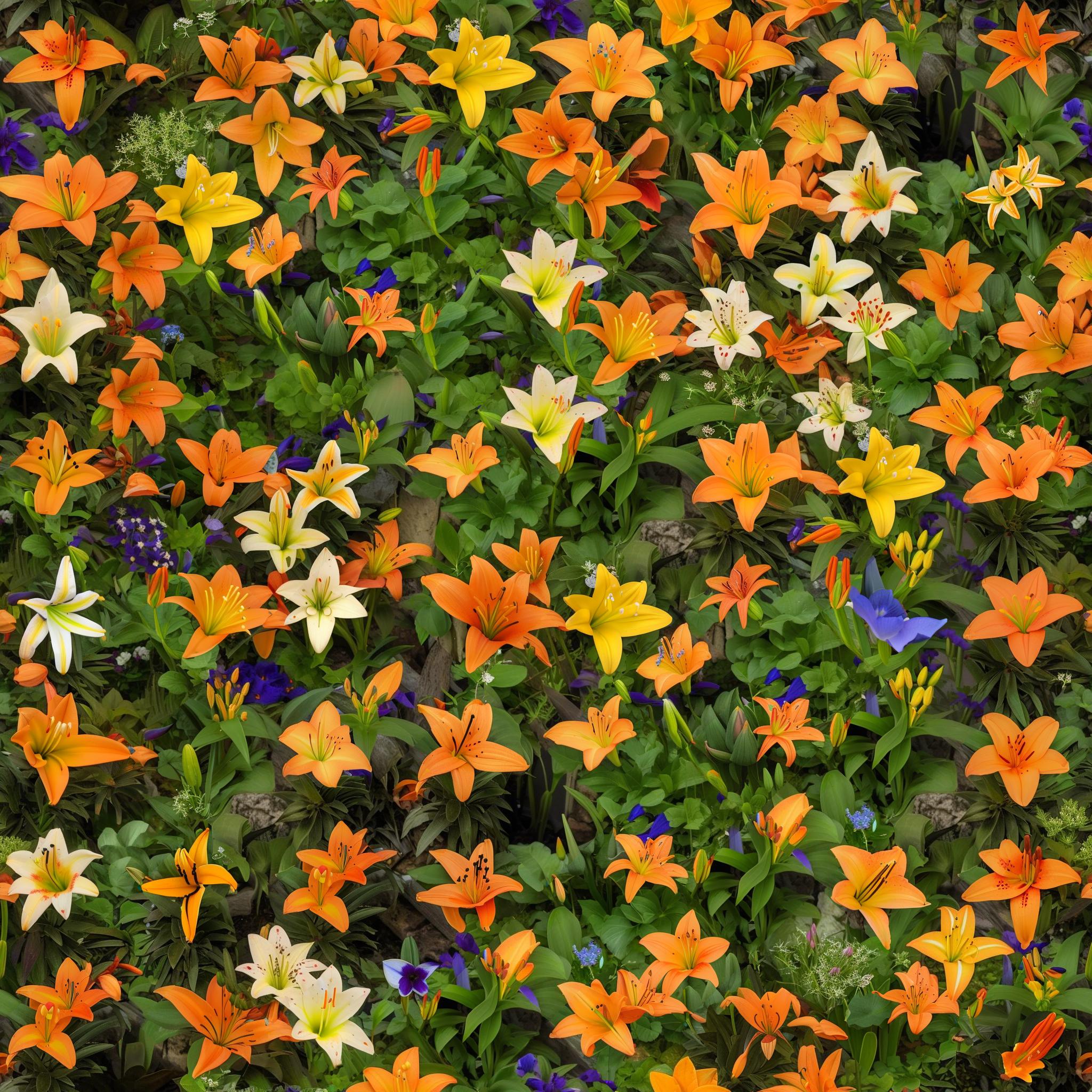}}
    \fbox{\includegraphics[width=0.24\linewidth-5pt]{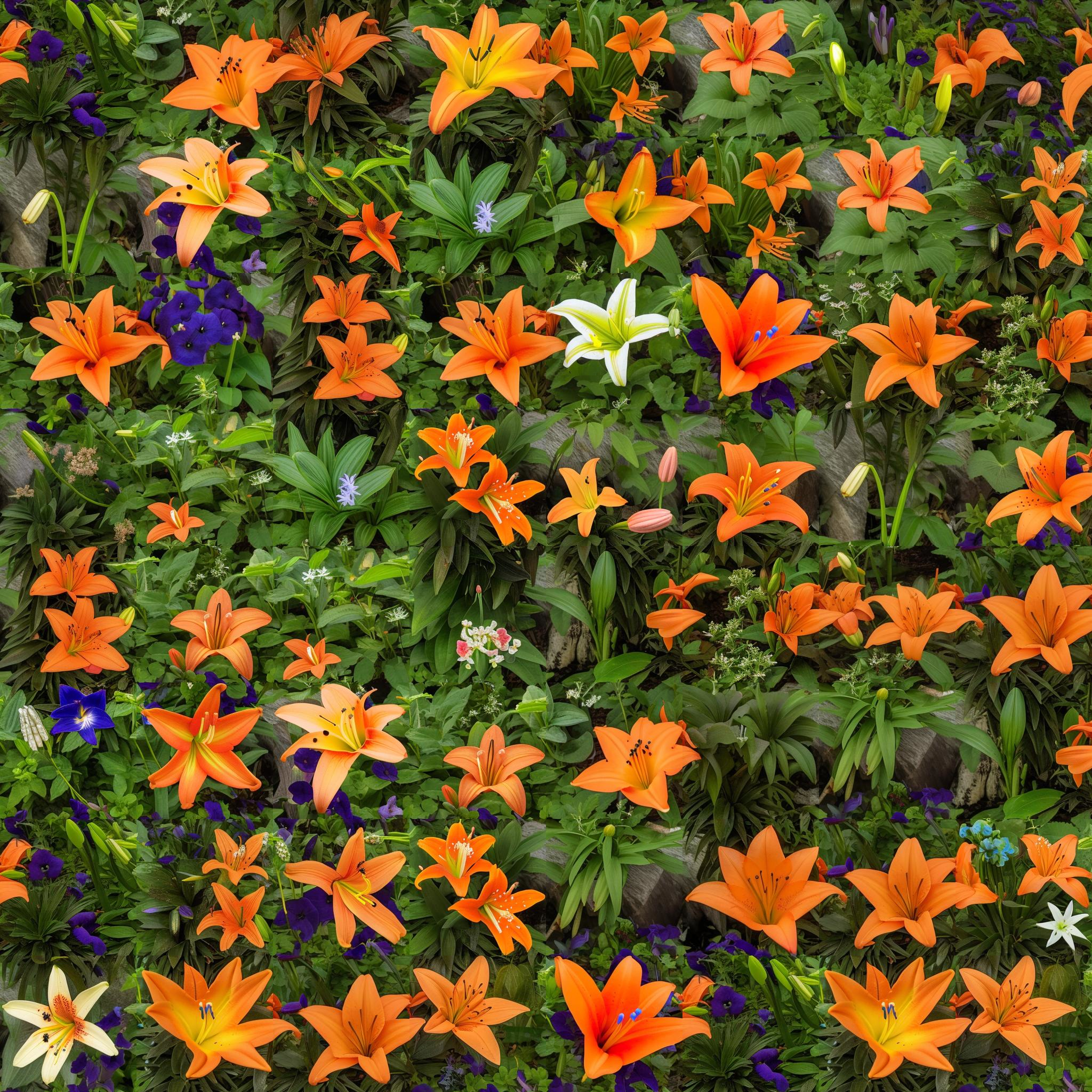}}
    \fbox{\adjincludegraphics[width=0.24\linewidth-5pt]{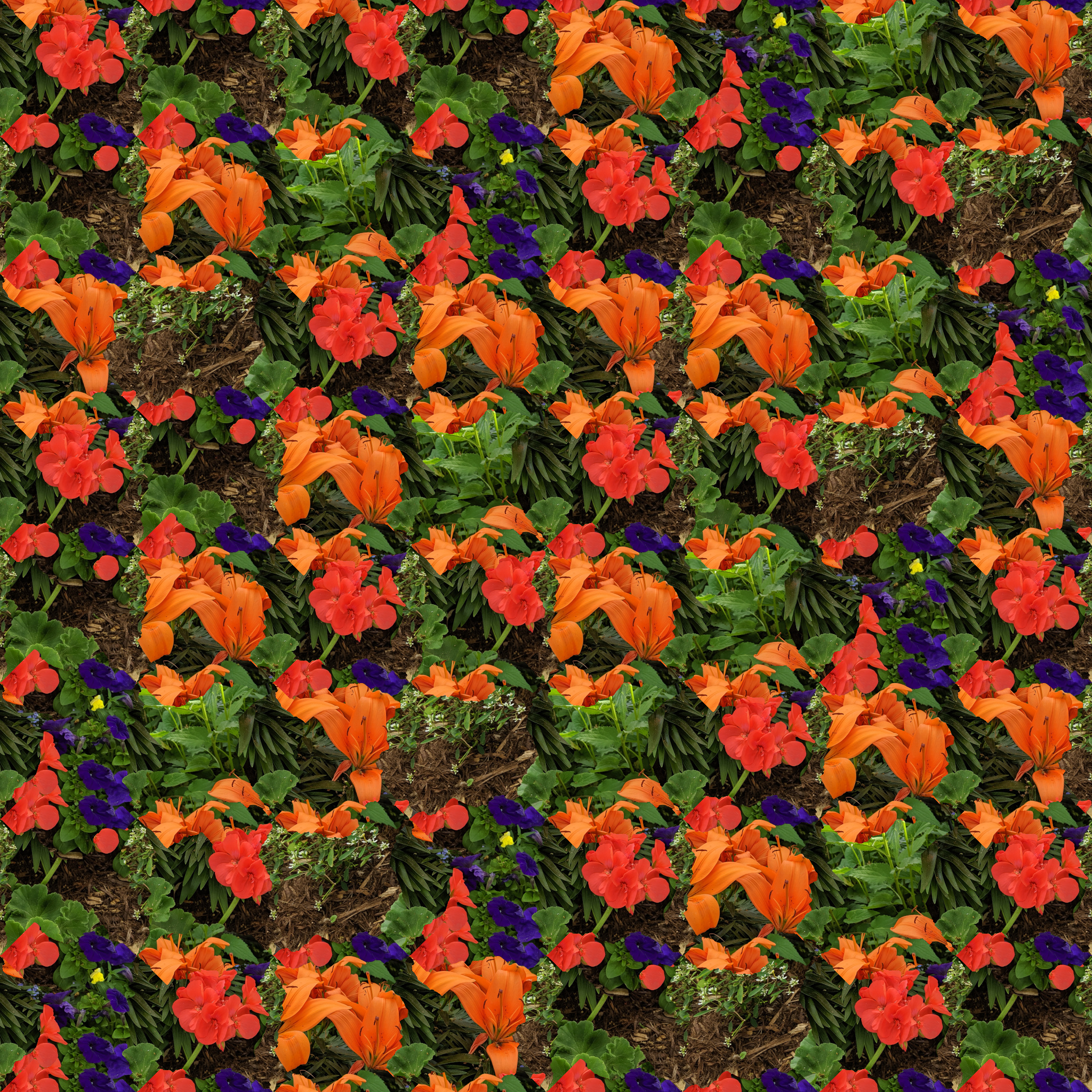}}
    \caption{\textbf{Img2Tile:} Our method can also start from
      template patches selected from a suitable photograph and a
      descriptive prompt (\emph{``bright orange lily in a flower
        garden, small blue and white flowers, leaves''}) (left).
      Please zoom-in on the Dual Wang tiled texture (2nd) and regular
      Wang tiled texture (3rd) to fully appreciate the generated
      texture detail. Our tile generation method produces a more
      diverse set of tiles compared to graph-cut based tile synthesis
      (right).}
    \label{fig:img2tile}
\end{figure*}

\section{Results}
\label{sec:results}

We implemented our content-aware tiling in
PyTorch~\cite{Paszke:2019:Pytorch}, using
\emph{Stable-Diffusion-XL}~\cite{StableDiffusionXL} to generate the
exemplar image from a user-provided text-prompt, and
\emph{Stable-Diffusion-2-Inpainting}~\cite{StableDiffusionInpainting}
for tile generation.  Any sufficiently performant combination of
text2image and inpainting models could be easily used instead,
including future more advanced models, without the need for any
additional training or per-model customization.  All results in this
paper use a tile texture resolution of $256 \times 256$ and a tiling
of $3$ edge colors.  We use an Euler sampler with $40$ inference
steps, and lower the CFG scale to $7.5$.

\autoref{fig:results} shows four different generated
tiled textures using our Dual Wang tile scheme.  The tiled textures,
as shown, have an effective resolution of $7168 \times 2560$ or
$28 \times 10$ tiles (\ie, each tile appears on average $3.5$ times).
As can be seen, our tiled textures include fine details as well as a
large diversity in the generated tiles.  \emph{We refer to the
  supplemental material for additional generated tiled textures}.

\section{Practical Consideration}

\paragraph{ Latent Inpainting}
At a high level, our technique treats the inpainting model as a black
box, which accepts a pixel-space image and mask. However, special care
must be taken when the underlying diffusion model operates in a (VAE)
latent space. Because we treat the diffusion model as a black box,
such a latent diffusion model encodes the boundary conditions before
each inpaint step, and then decodes the result again to pixel-space
when inpainting is complete.  This causes two potential issues:
\begin{itemize}
\item Current VAEs are quite lossy, which can cause some color shift
  over the course of the tile generation process.  A potential
  solution would be to performing tile-generation completely in latent
  space (\ie, only encode and decode once for the whole tile set).  We
  did not implement this, as this generally requires customization per
  diffusion model, thereby reducing flexibility.
\item VAE encoding also requires care in specifying the boundary
  conditions, as the encoding/decoding of boundary pixels might be
  affected by pixels \emph{inside} the mask.  We therefore extend the
  template patch far enough past the actual boundary into the masked
  area to not influence the latents in the boundary region.  Values
  copied inside the masked area will be overwritten by the inpainting
  step.
\end{itemize}

\paragraph{Template Patch  Selection}
The exterior boundary conditions are determined by selecting patches
from the exemplar image.  For many exemplar images a simple random
selection often works surprisingly well (\eg, the ``pebbles in a
stream'' and ``woven basket'' in~\autoref{fig:results}).  Similar to
Cohen~\etal~\shortcite{Cohen:2003:WTI} we potentially can also
repeatedly select a random set of candidate template patches from
which we retain the template patches that produce the highest quality
tiles according to CLIP-IQA~\cite{Wang:2023:ECA}.  Alternatively, when
the image contains strongly aligned features (\eg, ``wooden floor'' in
the supplemental material), we can exploit that our method only
requires that opposing vertical features along the horizontal
boundaries are aligned and likewise for opposing horizontal features
along the vertical boundaries.  Hence, we constrain the selection to
first pick a random horizontal/vertical line along which all
vertical/horizontal template patches are randomly selected.  This
constrained selection method is unique to our method as prior graph
cut-based methods require that all four quadrants are mutually aligned
and thus are unlikely to align structures when the horizontal and
vertical (constrained selection) line is randomly selected.  Finally,
if artistic control is desired, then the user can also manually select
the template patches by simply marking the horizontal and vertical
boundary lines per patch.

\paragraph{Tile Rejection}
Diffusion-based inpainting can introduce unwanted features such as
noticeable seams and fake watermarks.  Since every tile is generated
independently (given a set of boundary conditions), we can simply
regenerate the tile that exhibits the undesirable image feature using
a different seed.  The user can either manually mark the affected
tiles or follow a similar selection strategy as for the automatic
template patch selection. For each tile, we generate multiple
candidates ($4$ in our implementation), each with a different seed,
and retain the candidate with the highest quality. In this case we use
SIFID~\cite{Shaham:2019:SLG}, as we found that CLIP-IQA is blind to
fake watermarks; the most common artefact produced by
\emph{Stable-Diffusion-2-Inpainting}.

\paragraph{Performance}
The computational cost of context-aware tile generation depends on the
tiling variant and on whether tile rejection is used.  We generally
require 40ms per U-Net evaluation on an NVidia A5000, times $40$
inference steps per inpainting operation, yielding a $1.7$s total
computation time (including VAE encoding/decoding) per tile. When
generating 3-color Wang tiles (\ie, 81 tiles), this corresponds to a
total of $140$s. When evaluating 4 candidate tiles, including the cost
of scoring, this cost increases to $12$ minutes. For 3-color Dual Wang
tiles, we perform $243$ inpainting operations, yielding $7$ minutes
without rejection, or $36$ minutes with $4$ candidate rejection. The
cost of rejection can be avoided by manually selecting which tiles to
regenerate (approximately 2-5 minutes overhead). Newer diffusion
models are significantly faster ($0.2$s instead of $1.7$s, or an
$8\times$ speedup), but we did not implement such
optimizations. Furthermore, our method can trivially be parallelized
over multiple GPUs. We believe that, with careful engineering, the
computational synthesis cost can be significantly lowered.

\section{Additional Applications}
\label{sec:applications}

\paragraph{Img2Tile}
All results shown so far have been generated starting from a
text-prompt.  However, we can also skip the initial step and extract
the template patches from a suitable photograph; we also also expect
the user to provide a text-prompt describing the image content.  Once
the template patches have been selected, the tile generation proceeds
as before. \autoref{fig:img2tile} shows an example of a $3$-color Dual
Wang tiling and a regular Wang tiling with textured tiles generated
from a photograph of flowers with manually selected template patches
from the input photograph.  To ensure that the flowers fit within a
tile, we downsample the input photograph by a factor $4$.  Because our
tile generation leverages inpainting, we never directly copy any
texels from the input photograph.  This results in a greater diversity
of salient objects; \ie, different orientations/poses and even
semantically correct recoloring (\eg, the exemplar only contains
orange Lilies, whereas the tiles also includes white Lilies).

\begin{figure}[t!]
  \setlength{\fboxrule}{2pt}
  \setlength{\fboxsep}{0pt}
  \fbox{\includegraphics[width=.97\columnwidth-4pt]{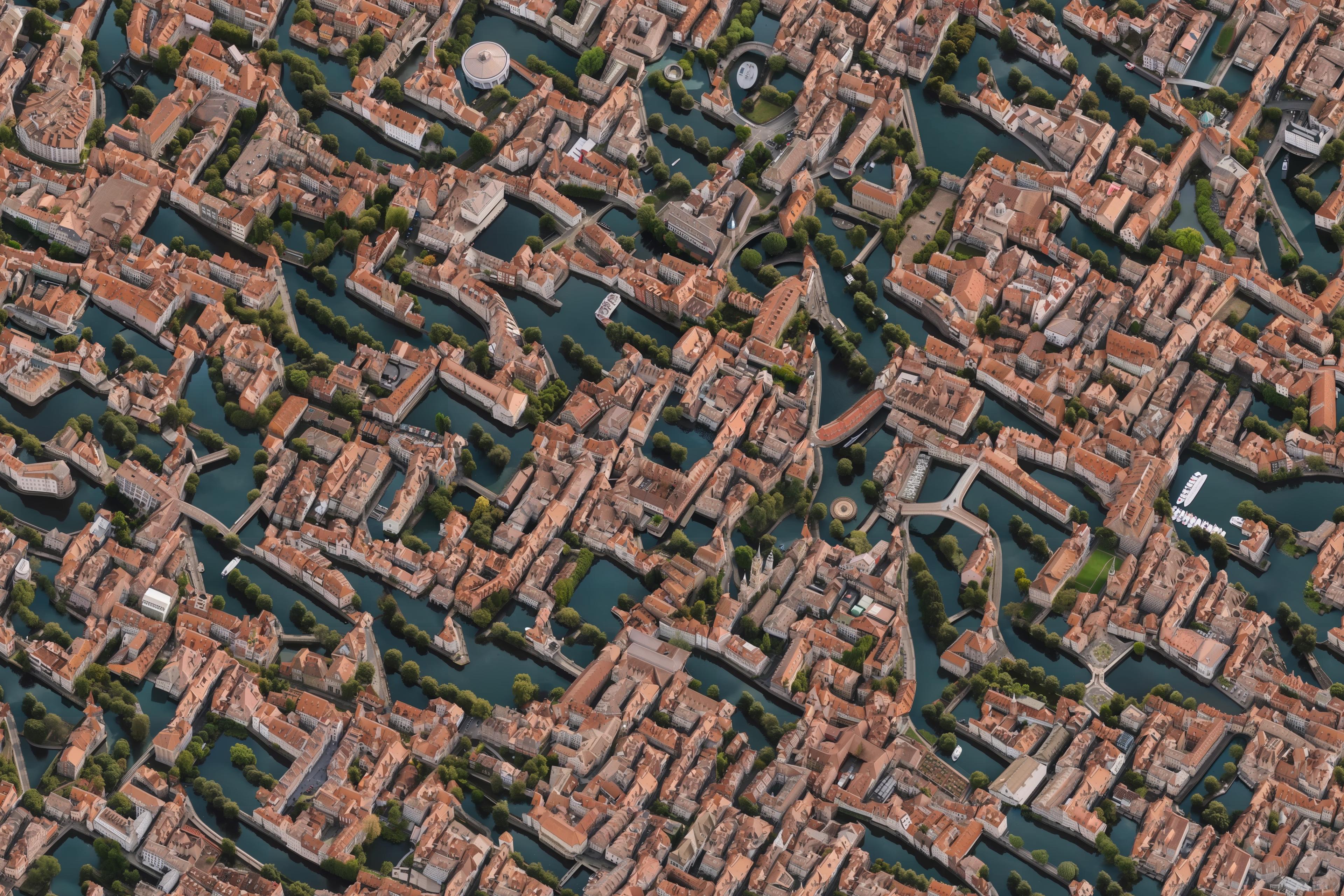}}
  \caption{\textbf{Infinite Stochastic Tiling:} An example of
    on-the-fly generation of Wang tile textures with a random seed to
    produce a unique texture per tile.}
  \label{fig:infinite}
\end{figure}

\paragraph{Infinite Stochastic Tiling}
Each tiling scheme discussed in~\autoref{sec:method} also supports
stochastic tiling; we simply generate additional tiles with the same
template patches using different inpainting seeds.  We can also
postpone the actual inpainting of the tiles to tiling-time.  During
preprocessing we compute the necessary boundary conditions (\ie, the
template patches). When evaluating the tiled texture, we generate the
corresponding tile on-the-fly with a random seed.  The effective
result would be equivalent to a tiling with an infinitely large
stochastic tile set.  \autoref{fig:infinite} shows an example of such
an infinite stochastic Wang tiled texture.  While more diverse, it
does come at a significant computational overhead; future advances in
inpainting might make this variant less costly and more practical.

\section{Evaluation}
\label{sec:evaluation}

\paragraph{Comparison to Graph-cut based Synthesis}
Cohen~\etal~\shortcite{Cohen:2003:WTI} propose to generate tile
textures using graph-cut quilting of diamond shaped templates cut from
an exemplar.  Because the same texels from the templates are copied to
multiple tile textures, the resulting tiles (\autoref{fig:img2tile},
right) are visually less diverse compared to our inpainting-based
method (\autoref{fig:img2tile}, 3rd column) that generates a related,
but unique, texture per tile.

\paragraph{Quantitative Comparison}
To quantitatively assess our content-aware tile generation method we
employ two commonly used quality metrics:
CLIP-IQA~\cite{Wang:2023:ECA} to measure the overall quality of the
tiles, and CLIPScore~\cite{Hessel:2021:CAR} to measure the semantic
similarity with the prompt.  Besides these classic image quality
measures, diversity between the different tiles also matters.  We
express this by computing the average pairwise correlation of the
inception features~\cite{Salimans:2016:IS} between tiles. The lower
the correlation, the more diverse the tile-set.  Other common metrics
such as SSIM and LPIPS assume aligned images, and are therefore not
suited for evaluating the quality of
tiles~\cite{Rodriguez-Pardo:2024:TAD}.  SIFID~\cite{Shaham:2019:SLG}
between the tiles and the exemplar is also not suited as it penalizes
diversity.  We also do not use TexTile~\cite{Rodriguez-Pardo:2024:TAD}
because it is intended for comparing self-tiling images.

We compute the average scores over all the examples included in this
paper and the supplemental material.  Because our exemplar images are
semantically more complex than commonly used stationary and stochastic
textures, we also report the scores on $12$ test textures (resolution
$> 256$) from SeamlessGAN~\cite{Rodriguez:2022:SSS} (\emph{see
  supplemental material for corresponding tilings}). We manually
create a text-prompt for each test texture in the SeamlessGAN set, and
use it as input to our content-aware tile generation, as well as for
evaluating CLIPScore. \autoref{tab:eval} lists the scores for each
test set for context-aware generated textures for different tiling
schemes. For completeness, we also include the scores for graph-cut
based Wang tile synthesis~\cite{Cohen:2003:WTI}. All tilings except
single self-tiling consist of $81$ tile textures.  The quantitative
comparisons show that our method produces higher quality Wang tile
textures for typical textures than the graph-cut based tile
synthesis. We found that context-aware tile generation generally
performs relatively better on semantic textures that exhibit less
regular structures than on the classic SeamlessGAN test set.  The
quantitative comparison also demonstrates that with a large tile set
($81$ in this case), stochastic self-tiling is a viable alternative
especially for inconspicuous boundaries.  Finally, Dual Wang tiling
performs overall best, and it produces the most diverse tilings for
\emph{both} test sets.

\begin{table}
  \caption{Quantitative comparison of context-aware tile generation
    for different tiling schemes using
    CLIPScore~\cite{Hessel:2021:CAR} (for semantic similarity; higher
    is better), CLIP-IQA~\cite{Wang:2023:ECA} (for quality; higher is
    better), and average correlation of the inception
    features~\cite{Salimans:2016:IS} (to measure diversity; lower is
    better) over the $12$ classic textures from
    SeamlessGAN~\cite{Rodriguez:2022:SSS}, as well as semantically
    more complex textures shown in this paper. Additionally, we also
    include the scores for Wang tiles synthesized using the graph-cut
    based method of Cohen~\etal~\shortcite{Cohen:2003:WTI} (3rd row
    marked in gray). All scores, except single self-tiling, are
    computed using $81$ tiles per scheme.}
  \label{tab:eval}
  {\small
    \newcommand{\BS}[1]{\textbf{#1}}
    \newcommand{\GR}[1]{\textcolor{gray}{#1}}
  \begin{tabular}{l||c|c|c||c|c|c}
                           & \multicolumn{3}{c||}{Classic Texture}   & \multicolumn{3}{c}{Semantic Textures} \\
                           \cline{2-7} 
                           & CLIP         & CLIP       & Incep.     & CLIP       & CLIP        & Incep. \\
                           & Score        & IQA        & Correl.    & Score      & IQA         & Correl. \\
    \hline 
    Single Self            & 24.587       & 0.765      & $\diagup$  & 23.370     & 0.746      & $\diagup$  \\
    Stoch. Self            & \BS{27.018}  & \UL{0.767} & 9.368      & 28.070     & 0.744      & \UL{5.144} \\
    \GR{Wang (cut)}        & \GR{24.467}   & \GR{\BS{0.785}} & \GR{10.089}     & \GR{26.548}     & \GR{0.739}      & \GR{5.773}      \\
    Wang Tile              & 26.002       & 0.762      & \UL{9.293} & \BS{28.57} & \UL{0.756} & 5.251      \\
    Dual Wang              & \UL{26.327}  & 0.765      & \BS{8.331} & \UL{28.44} & \BS{0.760} & \BS{4.875} \\
  \end{tabular}
  }
\end{table}

\begin{figure*}[t!]
    \setlength{\fboxrule}{2pt}
    \setlength{\fboxsep}{0pt}

    \includegraphics[width=.99\linewidth]{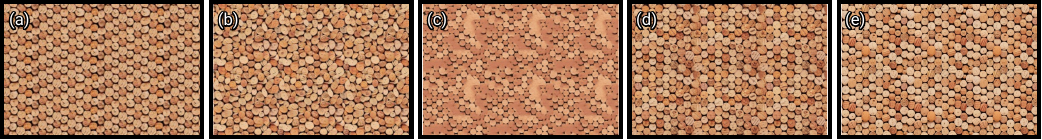}
  
    \caption{\textbf{Self-tiling Qualitative Comparison} of
      SeamlessGAN (a), Neural Texture Synthesis with TexTile loss (b),
      SinFusion with TexTile loss (c), Conditional Noise Rolling (d),
      and our context-aware self-tiling (e) on an exemplar from the
      SeamlessGAN test set.}
    \label{fig:selftile}
\end{figure*}

\paragraph{Self-tiling Comparison}
While the focus of our method is on generating tile sets of multiple
mutually tileable images, our method can also be used for generating
self-tiling images.  While there exist a number of prior
learning-based methods for generating self-tiling images, it should be
stressed than none can be extended to produce multi-tile sets or
stochastic self-tiling images.

For completeness,~\autoref{fig:selftile} and~\autoref{tab:eval2}
compare our method against recent self-tiling image generation
methods: the feedforward SeamlessGAN~\cite{Rodriguez:2022:SSS}, the
optimization based Neural Texture Synthesis~\cite{Heitz:2021:SWL} with
a TexTile loss~\cite{Rodriguez:2022:SSS} to promote tileability, the
SinFusion~\cite{Nikankin:2023:STD} single-image diffusion model in
which each denoising step is interleaved with an TexTile maximization
step, and Conditional Noise Rolling~\cite{Vecchio:2023:CAC} using the
same diffusion models as our method.  Because SeamlessGAN is not
trained for semantic textures, we only compare on the SeamlessGAN
dataset for fairness.  In addition to CLIPScore~\cite{Hessel:2021:CAR}
and CLIP-IQA~\cite{Wang:2023:ECA} used in the multi-tile comparison
(\autoref{tab:eval}), we also include SIFID~\cite{Shaham:2019:SLG} to
measure the similarity to the exemplar, and
TexTile~\cite{Rodriguez:2022:SSS} to quantify self-tileability.  For
both Conditional Noise Rolling and our method, we perform tile
selection with 4 candidate tiles using SIFID as the selection
criterion; for completeness we also include a variant where we replace
the selection criterion by TexTile to favor tileability rather than
quality.  From~\autoref{tab:eval2} we see that none of the methods
outperforms the others on all error metrics.  Conditional Noise
Rolling does not score well on tileability (TexTile) due to the very
small inpainting region ($1/16$th of the image size). Our method with
SIFID-based candidate selection performs similar to SeamlessGAN while
adhering better to the exemplar (SIFID).  When using TexTile for tile
selection, our method performs similar to the methods that explicitly
optimize tileability.  Qualitatively (\autoref{fig:selftile}) we see
that Conditional Noise Rolling (d) and our method with SIFID selection
(e) best preserve the characteristics of the exemplar without
deforming the corks (\eg, Neural Texture Synthesis (b)) or reducing
variation (\eg, SinFusion (c)). Moreover, in contrast to SeamlessGAN
(a), our results appear less regular and exhibit less tileability
artifacts (\eg, Conditional Noise Rolling).  We believe our
context-aware tile generation method (with SIFID selection) strikes a
good balance between exemplar similarity and tileability.

\begin{table}
  \caption{Quantitative comparison of different learning-based
    self-tiling texture generation methods
    (SeamlessGAN~\cite{Rodriguez:2022:SSS}, Neural Texture
    Synthesis~\cite{Heitz:2021:SWL} with a TexTile
    loss~\cite{Rodriguez:2022:SSS}, SinFusion~\cite{Nikankin:2023:STD}
    interleaved with a TexTile optimization step, and Conditional
    Noise Rolling~\cite{Vecchio:2023:CAC}) using
    CLIPScore~\cite{Hessel:2021:CAR} (for semantic similarity; higher
    is better), CLIP-IQA~\cite{Wang:2023:ECA} (for quality; higher is
    better), SIFID~\cite{Shaham:2019:SLG} (for similarity to the
    exemplar; lower is better), and TexTile score (for
    self-tileability; higher is better) over the $12$ classic textures
    from SeamlessGAN~\cite{Rodriguez:2022:SSS}.  For both Conditional
    Noise Rolling and our method, we employ tile selection (using
    SIFID or TexTile) from 4 candidate tiles.}
  \label{tab:eval2}
  {\small
    \newcommand{\BS}[1]{\textbf{#1}}
  \begin{tabular}{l||c|c|c|c|c}
                                 & CLIP        & CLIP          & SIFID      & TexTile     \\
                                 & Score       & IQA           &            &             \\
    \hline 
    SeamlessGAN                  & 22.528      & 0.7803        &  9.215     & 0.7149      \\
    Neural Tex. Synth. (TexTile) & 22.350      & \BS{0.8012}   &  8.599     & \BS{0.7783} \\
    SinFusion (TexTile)          & 20.829      & \UL{0.7918}   & 10.184     & 0.7407      \\
    Cond. Noise Roll. (SIFID)    & 22.668      & 0.7447        & \BS{6.607} & 0.4826      \\
    Ours Self Tiling (SIFID)     & \UL{22.857} & 0.7717        & 7.978      & 0.6950      \\
    Cond. Noise Roll. (TexTile)  & \BS{23.212} & 0.7445        & \UL{6.926} & 0.5270      \\
    Ours Self Tiling (TexTile)   & 22.228      & 0.7737        & 9.119      & \UL{0.7494} \\
  \end{tabular}
  }
\end{table}

\section{Discussion}
\label{sec:discussion}
\paragraph{Dual Wang Tile Storage Requirements}
Dual Wang tile sets contain double the number of tiles compared to a
regular Wang tile set.  However, it should be noted that the interior
and cross tile textures contain just $50\%$ of the texels compared to
regular Wang tile textures.  Hence, when packed in a single texture
map, the dual Wang tiles have exactly the same storage costs.

Wei~\shortcite{Wei:2004:TBT} packs a complete regular Wang tile set in
a single texture using every tile only once and where each edge is
matched.  Such a packing ensures minimal storage and fast fetching
and, importantly, correct texture filtering.  However, Wei's Wang tile
packing only guarantees that each tile occurs once (\ie, interior Dual
Wang tile), but it does not offer such guarantee for each cross tile.
Fortunately, a valid minimal Dual Wang tile packing is possible, and
we refer to \autoref{sec:packing} for novel generative algorithms for
Dual Wang tile packings with odd and even colors.

\paragraph{Wang Tile Equivalence}
It is possible to assemble the Dual Wang tiles into a regular Wang
tile set because the underlying tiling is the same.  Each texture in
each of underlying Wang tiles is determined by the colors of the four
edges, as well as the two colors of incident edges (not part of the
Wang tile) at each corner.  This implies that each texture is
determined by $12$ edges, and thus the equivalent number of unique
Wang tiles textures required equals $C^{12} = (C^3)^4$, or a Wang
tiling consisting of $C^3$ colors.  While the number of unique tiles
is huge compared to regular Wang tiles (\eg, for $C = 3$, regular Wang
tiles consists of $81$ textures and dual Wang tiles are equivalent to
$531,\!441$ regular Wang tile textures), the resulting Wang tiles are
a combination of a smaller set of basis textures.

\paragraph{Necessity of 3-step Dual Tile Generation}
Both regular Wang tile textures as well as Corner Wang tile textures
can be generated directly from the template patches.  This raises the
question whether it would be possible to generate Dual Wang Tiles in a
single pass with inpainting?  For argument's sake, lets assume that we
only need infinitely thin boundary conditions in order to generate the
interior of a tile.  These boundary conditions need to be free of
unwanted discontinuities along each edge. This is currently enforced
by the construction of template patches for regular Wang tiles by
picking a contiguous boundary in the source image. This is a
sufficient condition for regular Wang tiles, and hence direct
synthesis is possible.  However, for Dual Wang tiles we also require
that there are no unwanted discontinuities on the corners of the
generated tiles, even at a single point.  Hence, direct synthesis
would therefore require that the boundary conditions would also match
where two edges meet. This is difficult to achieve with template
patches selected from an exemplar texture.  Thus, we must first
synthesize a boundary condition with repetition where those boundaries
can meet without discontinuity. We resolve this by computing templates
which in turn are computed from Wang tiles; in each step we expand the
continuity along the edges and corners.  Hence the need for a
multi-stage generation process.

\paragraph{Limitations}
Not all exemplar images are suitable for tile generation. Images with
salient features larger than the tile size fail to produce
satisfactory results. We address this issue by simply downsampling the
image such that the salient features are smaller than the tile size.
Furthermore, images with perspective distortions (\eg, finite
vanishing points) or images with brightness gradients typically result
in noticeable repetitive patterns. However, prior graph-cut based
methods also struggle with such textures.  In general, our
context-aware tile generation effectively ‘paints’ around localized
challenges, but it cannot correct global deviations. When starting
from a prompt, diffusion is not aware of this assumption (unless
carefully specified in the prompt), hence we apply (unconditional)
noise-rolling to impose the assumptions during exemplar synthesis.
Note we did not apply noise-rolling as a preprocessing step on the
SeamlessGAN dataset used for evaluation in which many exemplar violate
the above assumptions, hence the SeamlessGAN results can be seen as a
worst case scenario; applying noise-rolling (or any other
stationarization method) as a preprocessing step could further improve
performance for suboptimal exemplars.

\section{Conclusion}
\label{sec:conclusion}

In this paper we presented an easy to implement, yet flexible, method
for generating tileable textures using inpainting.  Unlike prior work,
we do not reuse pixels or patches from the exemplar, but instead use
them to impose exterior boundary conditions on the tile textures.
Furthermore, our tile generation method can easily accommodate tiling
schemes beyond self-tiling.  By appropriate selection of the template
patches, we ensure that the boundaries match seamlessly with
corresponding boundaries on matching tiles.  We demonstrated our
method on a variety of tiling schemes, including a novel Dual Wang
tile scheme that provides greater tile diversity than prior Wang tile
schemes without incurring an additional storage cost.  For future
work, we would like to speed up and improve the automated methods for
selecting the template patches and defective tile rejection.
Furthermore, we would like to explore other tiling schemes.

\begin{acks}
This research was supported in part by NSF grant IIS-1909028.
\end{acks}

\bibliographystyle{ACM-Reference-Format}
\bibliography{references}


\begin{thebibliography}{71}


\ifx \showCODEN    \undefined \def \showCODEN     #1{\unskip}     \fi
\ifx \showDOI      \undefined \def \showDOI       #1{#1}\fi
\ifx \showISBNx    \undefined \def \showISBNx     #1{\unskip}     \fi
\ifx \showISBNxiii \undefined \def \showISBNxiii  #1{\unskip}     \fi
\ifx \showISSN     \undefined \def \showISSN      #1{\unskip}     \fi
\ifx \showLCCN     \undefined \def \showLCCN      #1{\unskip}     \fi
\ifx \shownote     \undefined \def \shownote      #1{#1}          \fi
\ifx \showarticletitle \undefined \def \showarticletitle #1{#1}   \fi
\ifx \showURL      \undefined \def \showURL       {\relax}        \fi
\providecommand\bibfield[2]{#2}
\providecommand\bibinfo[2]{#2}
\providecommand\natexlab[1]{#1}
\providecommand\showeprint[2][]{arXiv:#2}

\bibitem[Barnes et~al\mbox{.}(2009)]%
        {Barnes:2009:PMR}
\bibfield{author}{\bibinfo{person}{Connelly Barnes}, \bibinfo{person}{Eli
  Shechtman}, \bibinfo{person}{Adam Finkelstein}, {and} \bibinfo{person}{Dan~B
  Goldman}.} \bibinfo{year}{2009}\natexlab{}.
\newblock \showarticletitle{PatchMatch: A randomized correspondence algorithm
  for structural image editing}.
\newblock \bibinfo{journal}{\emph{ACM Trans. Graph.}} \bibinfo{volume}{28},
  \bibinfo{number}{3} (\bibinfo{year}{2009}), \bibinfo{pages}{24}.
\newblock


\bibitem[Bergmann et~al\mbox{.}(2017)]%
        {Bergmann:2017:LTM}
\bibfield{author}{\bibinfo{person}{Urs Bergmann}, \bibinfo{person}{Nikolay
  Jetchev}, {and} \bibinfo{person}{Roland Vollgraf}.}
  \bibinfo{year}{2017}\natexlab{}.
\newblock \showarticletitle{Learning texture manifolds with the Periodic
  Spatial GAN}. In \bibinfo{booktitle}{\emph{ICML}}. \bibinfo{pages}{469--477}.
\newblock


\bibitem[Chen et~al\mbox{.}(2023)]%
        {Chen:2023:FDG}
\bibfield{author}{\bibinfo{person}{Rui Chen}, \bibinfo{person}{Yongwei Chen},
  \bibinfo{person}{Ningxin Jiao}, {and} \bibinfo{person}{Kui Jia}.}
  \bibinfo{year}{2023}\natexlab{}.
\newblock \showarticletitle{Fantasia3D: Disentangling Geometry and Appearance
  for High-quality Text-to-3D Content Creation}. In
  \bibinfo{booktitle}{\emph{ICCV}}.
\newblock


\bibitem[Cohen et~al\mbox{.}(2003)]%
        {Cohen:2003:WTI}
\bibfield{author}{\bibinfo{person}{Michael~F Cohen}, \bibinfo{person}{Jonathan
  Shade}, \bibinfo{person}{Stefan Hiller}, {and} \bibinfo{person}{Oliver
  Deussen}.} \bibinfo{year}{2003}\natexlab{}.
\newblock \showarticletitle{Wang tiles for image and texture generation}.
\newblock \bibinfo{journal}{\emph{ACM Trans. Graph.}} \bibinfo{volume}{22},
  \bibinfo{number}{3} (\bibinfo{year}{2003}), \bibinfo{pages}{287--294}.
\newblock


\bibitem[Dhariwal and Nichol(2021)]%
        {Dhariwal:2021:DMB}
\bibfield{author}{\bibinfo{person}{Prafulla Dhariwal} {and}
  \bibinfo{person}{Alexander Nichol}.} \bibinfo{year}{2021}\natexlab{}.
\newblock \showarticletitle{Diffusion models beat gans on image synthesis}.
\newblock \bibinfo{journal}{\emph{NeurIPS}}  \bibinfo{volume}{34}
  (\bibinfo{year}{2021}), \bibinfo{pages}{8780--8794}.
\newblock


\bibitem[Efros(2001)]%
        {Efros:2001:IQT}
\bibfield{author}{\bibinfo{person}{Alexei~A Efros}.}
  \bibinfo{year}{2001}\natexlab{}.
\newblock \showarticletitle{Image Quilting for Texture Synthesis and Transfer}.
  In \bibinfo{booktitle}{\emph{Proc. Siggraph 2001}}.
  \bibinfo{pages}{341--346}.
\newblock


\bibitem[Efros and Leung(1999)]%
        {Efros:1999:TSN}
\bibfield{author}{\bibinfo{person}{Alexei~A Efros} {and}
  \bibinfo{person}{Thomas~K Leung}.} \bibinfo{year}{1999}\natexlab{}.
\newblock \showarticletitle{Texture synthesis by non-parametric sampling}. In
  \bibinfo{booktitle}{\emph{CVPR}}, Vol.~\bibinfo{volume}{2}.
  \bibinfo{pages}{1033--1038}.
\newblock


\bibitem[Fr{\"u}hst{\"u}ck et~al\mbox{.}(2019)]%
        {Fruhstuck:2019:TSL}
\bibfield{author}{\bibinfo{person}{Anna Fr{\"u}hst{\"u}ck},
  \bibinfo{person}{Ibraheem Alhashim}, {and} \bibinfo{person}{Peter Wonka}.}
  \bibinfo{year}{2019}\natexlab{}.
\newblock \showarticletitle{Tilegan: synthesis of large-scale non-homogeneous
  textures}.
\newblock \bibinfo{journal}{\emph{ACM Trans. Graph.}} \bibinfo{volume}{38},
  \bibinfo{number}{4} (\bibinfo{year}{2019}), \bibinfo{pages}{1--11}.
\newblock


\bibitem[Fu and Leung(2005)]%
        {Fu:2005:TTA}
\bibfield{author}{\bibinfo{person}{Chi-Wing Fu} {and} \bibinfo{person}{Man-Kang
  Leung}.} \bibinfo{year}{2005}\natexlab{}.
\newblock \showarticletitle{Texture Tiling on Arbitrary Topological Surfaces
  using Wang Tiles.}. In \bibinfo{booktitle}{\emph{Rendering Techniques}}.
  \bibinfo{pages}{99--104}.
\newblock


\bibitem[Gatys et~al\mbox{.}(2015)]%
        {Gatys:2015:TSC}
\bibfield{author}{\bibinfo{person}{Leon Gatys}, \bibinfo{person}{Alexander~S
  Ecker}, {and} \bibinfo{person}{Matthias Bethge}.}
  \bibinfo{year}{2015}\natexlab{}.
\newblock \showarticletitle{Texture synthesis using convolutional neural
  networks}.
\newblock \bibinfo{journal}{\emph{NeurIPS}}  \bibinfo{volume}{28}
  (\bibinfo{year}{2015}).
\newblock


\bibitem[Guo et~al\mbox{.}(2022)]%
        {Guo:2022:UAT}
\bibfield{author}{\bibinfo{person}{Shouchang Guo}, \bibinfo{person}{Valentin
  Deschaintre}, \bibinfo{person}{Douglas Noll}, {and} \bibinfo{person}{Arthur
  Roullier}.} \bibinfo{year}{2022}\natexlab{}.
\newblock \showarticletitle{U-attention to textures: hierarchical hourglass
  vision transformer for universal texture synthesis}. In
  \bibinfo{booktitle}{\emph{Proc. of the 19th ACM SIGGRAPH European Conference
  on Visual Media Production}}. \bibinfo{pages}{1--10}.
\newblock


\bibitem[Ham et~al\mbox{.}(2023)]%
        {Ham:2023:MPD}
\bibfield{author}{\bibinfo{person}{Cusuh Ham}, \bibinfo{person}{James Hays},
  \bibinfo{person}{Jingwan Lu}, \bibinfo{person}{Krishna~Kumar Singh},
  \bibinfo{person}{Zhifei Zhang}, {and} \bibinfo{person}{Tobias Hinz}.}
  \bibinfo{year}{2023}\natexlab{}.
\newblock \showarticletitle{Modulating Pretrained Diffusion Models for
  Multimodal Image Synthesis}. In \bibinfo{booktitle}{\emph{ACM SIGGRAPH 2023
  Conference Proceedings}}. Article \bibinfo{articleno}{35},
  \bibinfo{numpages}{11}~pages.
\newblock


\bibitem[Heeger and Bergen(1995)]%
        {Heeger:1995:PTA}
\bibfield{author}{\bibinfo{person}{David~J Heeger} {and}
  \bibinfo{person}{James~R Bergen}.} \bibinfo{year}{1995}\natexlab{}.
\newblock \showarticletitle{Pyramid-based texture analysis/synthesis}. In
  \bibinfo{booktitle}{\emph{Proceedings of the 22nd annual conference on
  Computer graphics and interactive techniques}}. \bibinfo{pages}{229--238}.
\newblock


\bibitem[Heitz et~al\mbox{.}(2021)]%
        {Heitz:2021:SWL}
\bibfield{author}{\bibinfo{person}{Eric Heitz}, \bibinfo{person}{Kenneth
  Vanhoey}, \bibinfo{person}{Thomas Chambon}, {and} \bibinfo{person}{Laurent
  Belcour}.} \bibinfo{year}{2021}\natexlab{}.
\newblock \showarticletitle{A sliced wasserstein loss for neural texture
  synthesis}. In \bibinfo{booktitle}{\emph{CVPR}}. \bibinfo{pages}{9412--9420}.
\newblock


\bibitem[Henzler et~al\mbox{.}(2021)]%
        {Henzler:2021:GMB}
\bibfield{author}{\bibinfo{person}{Philipp Henzler}, \bibinfo{person}{Valentin
  Deschaintre}, \bibinfo{person}{Niloy~J Mitra}, {and} \bibinfo{person}{Tobias
  Ritschel}.} \bibinfo{year}{2021}\natexlab{}.
\newblock \showarticletitle{Generative modelling of BRDF textures from flash
  images}.
\newblock \bibinfo{journal}{\emph{ACM Trans. Graph.}} \bibinfo{volume}{40},
  \bibinfo{number}{6} (\bibinfo{year}{2021}), \bibinfo{pages}{1--13}.
\newblock


\bibitem[Henzler et~al\mbox{.}(2020)]%
        {Henzler:2020:LN3}
\bibfield{author}{\bibinfo{person}{Philipp Henzler}, \bibinfo{person}{Niloy~J
  Mitra}, {and} \bibinfo{person}{Tobias Ritschel}.}
  \bibinfo{year}{2020}\natexlab{}.
\newblock \showarticletitle{Learning a neural 3d texture space from 2d
  exemplars}. In \bibinfo{booktitle}{\emph{CVPR}}. \bibinfo{pages}{8356--8364}.
\newblock


\bibitem[Hessel et~al\mbox{.}(2021)]%
        {Hessel:2021:CAR}
\bibfield{author}{\bibinfo{person}{Jack Hessel}, \bibinfo{person}{Ari
  Holtzman}, \bibinfo{person}{Maxwell Forbes}, \bibinfo{person}{Ronan~Le Bras},
  {and} \bibinfo{person}{Yejin Choi}.} \bibinfo{year}{2021}\natexlab{}.
\newblock \showarticletitle{{CLIPScore:} A Reference-free Evaluation Metric for
  Image Captioning}. In \bibinfo{booktitle}{\emph{EMNLP}}.
\newblock


\bibitem[Karras et~al\mbox{.}(2022)]%
        {Karras:2022:EDS}
\bibfield{author}{\bibinfo{person}{Tero Karras}, \bibinfo{person}{Miika
  Aittala}, \bibinfo{person}{Timo Aila}, {and} \bibinfo{person}{Samuli Laine}.}
  \bibinfo{year}{2022}\natexlab{}.
\newblock \showarticletitle{Elucidating the Design Space of Diffusion-Based
  Generative Models}. In \bibinfo{booktitle}{\emph{NeurIPS}}.
\newblock


\bibitem[Kawar et~al\mbox{.}(2023)]%
        {Kawar:2023:MTR}
\bibfield{author}{\bibinfo{person}{Bahjat Kawar}, \bibinfo{person}{Shiran
  Zada}, \bibinfo{person}{Oran Lang}, \bibinfo{person}{Omer Tov},
  \bibinfo{person}{Huiwen Chang}, \bibinfo{person}{Tali Dekel},
  \bibinfo{person}{Inbar Mosseri}, {and} \bibinfo{person}{Michal Irani}.}
  \bibinfo{year}{2023}\natexlab{}.
\newblock \showarticletitle{Imagic: Text-based real image editing with
  diffusion models}. In \bibinfo{booktitle}{\emph{CVPR}}.
  \bibinfo{pages}{6007--6017}.
\newblock


\bibitem[Kim et~al\mbox{.}(2022)]%
        {Kim:2022:DTG}
\bibfield{author}{\bibinfo{person}{Gwanghyun Kim}, \bibinfo{person}{Taesung
  Kwon}, {and} \bibinfo{person}{Jong~Chul Ye}.}
  \bibinfo{year}{2022}\natexlab{}.
\newblock \showarticletitle{Diffusionclip: Text-guided diffusion models for
  robust image manipulation}. In \bibinfo{booktitle}{\emph{CVPR}}.
  \bibinfo{pages}{2426--2435}.
\newblock


\bibitem[Kol{\'a}{\v{r}} et~al\mbox{.}(2016)]%
        {Kolavr:2016:RTS}
\bibfield{author}{\bibinfo{person}{Martin Kol{\'a}{\v{r}}},
  \bibinfo{person}{Alan Chalmers}, {and} \bibinfo{person}{Kurt Debattista}.}
  \bibinfo{year}{2016}\natexlab{}.
\newblock \showarticletitle{Repeatable texture sampling with interchangeable
  patches}.
\newblock \bibinfo{journal}{\emph{The Visual Computer}}  \bibinfo{volume}{32}
  (\bibinfo{year}{2016}), \bibinfo{pages}{1263--1272}.
\newblock


\bibitem[Kopf et~al\mbox{.}(2006)]%
        {Kopf:2006:RWT}
\bibfield{author}{\bibinfo{person}{Johannes Kopf}, \bibinfo{person}{Daniel
  Cohen-Or}, \bibinfo{person}{Oliver Deussen}, {and} \bibinfo{person}{Dani
  Lischinski}.} \bibinfo{year}{2006}\natexlab{}.
\newblock \showarticletitle{Recursive Wang tiles for real-time blue noise}.
\newblock Vol.~\bibinfo{volume}{25}. \bibinfo{pages}{509--518}.
\newblock


\bibitem[Kwatra et~al\mbox{.}(2005)]%
        {Kwatra:2005:TOE}
\bibfield{author}{\bibinfo{person}{Vivek Kwatra}, \bibinfo{person}{Irfan Essa},
  \bibinfo{person}{Aaron Bobick}, {and} \bibinfo{person}{Nipun Kwatra}.}
  \bibinfo{year}{2005}\natexlab{}.
\newblock \showarticletitle{Texture optimization for example-based synthesis}.
\newblock In \bibinfo{booktitle}{\emph{ACM SIGGRAPH 2005 Papers}}.
  \bibinfo{pages}{795--802}.
\newblock


\bibitem[Kwatra et~al\mbox{.}(2003)]%
        {Kwatra:2003:GTI}
\bibfield{author}{\bibinfo{person}{Vivek Kwatra}, \bibinfo{person}{Arno
  Sch{\"o}dl}, \bibinfo{person}{Irfan Essa}, \bibinfo{person}{Greg Turk}, {and}
  \bibinfo{person}{Aaron Bobick}.} \bibinfo{year}{2003}\natexlab{}.
\newblock \showarticletitle{Graphcut textures: Image and video synthesis using
  graph cuts}.
\newblock \bibinfo{journal}{\emph{ACM Trans. Graph.}} \bibinfo{volume}{22},
  \bibinfo{number}{3} (\bibinfo{year}{2003}), \bibinfo{pages}{277--286}.
\newblock


\bibitem[Lagae and Dutr{\'e}(2005)]%
        {Lagae:2005:POD}
\bibfield{author}{\bibinfo{person}{Ares Lagae} {and} \bibinfo{person}{Philip
  Dutr{\'e}}.} \bibinfo{year}{2005}\natexlab{}.
\newblock \showarticletitle{A procedural object distribution function}.
\newblock \bibinfo{journal}{\emph{ACM Trans. Graph.}} \bibinfo{volume}{24},
  \bibinfo{number}{4} (\bibinfo{year}{2005}), \bibinfo{pages}{1442--1461}.
\newblock


\bibitem[Li and Wand(2016)]%
        {Li:2016:PRT}
\bibfield{author}{\bibinfo{person}{Chuan Li} {and} \bibinfo{person}{Michael
  Wand}.} \bibinfo{year}{2016}\natexlab{}.
\newblock \showarticletitle{Precomputed real-time texture synthesis with
  markovian generative adversarial networks}. In
  \bibinfo{booktitle}{\emph{ECCV}}. \bibinfo{pages}{702--716}.
\newblock


\bibitem[Li et~al\mbox{.}(2017)]%
        {Li:2017:DTS}
\bibfield{author}{\bibinfo{person}{Yijun Li}, \bibinfo{person}{Chen Fang},
  \bibinfo{person}{Jimei Yang}, \bibinfo{person}{Zhaowen Wang},
  \bibinfo{person}{Xin Lu}, {and} \bibinfo{person}{Ming-Hsuan Yang}.}
  \bibinfo{year}{2017}\natexlab{}.
\newblock \showarticletitle{Diversified texture synthesis with feed-forward
  networks}. In \bibinfo{booktitle}{\emph{CVPR}}. \bibinfo{pages}{3920--3928}.
\newblock


\bibitem[Liu et~al\mbox{.}(2023)]%
        {Liu:2023:Z1T}
\bibfield{author}{\bibinfo{person}{Ruoshi Liu}, \bibinfo{person}{Rundi Wu},
  \bibinfo{person}{Basile Van~Hoorick}, \bibinfo{person}{Pavel Tokmakov},
  \bibinfo{person}{Sergey Zakharov}, {and} \bibinfo{person}{Carl Vondrick}.}
  \bibinfo{year}{2023}\natexlab{}.
\newblock \showarticletitle{Zero-1-to-3: Zero-shot one image to 3d object}. In
  \bibinfo{booktitle}{\emph{ICCV}}. \bibinfo{pages}{9298--9309}.
\newblock


\bibitem[Liu et~al\mbox{.}(2022)]%
        {Liu:2022:FMS}
\bibfield{author}{\bibinfo{person}{Xiaokang Liu}, \bibinfo{person}{Chenran Li},
  \bibinfo{person}{Lin Lu}, \bibinfo{person}{Oliver Deussen}, {and}
  \bibinfo{person}{Changhe Tu}.} \bibinfo{year}{2022}\natexlab{}.
\newblock \showarticletitle{Fabricable Multi-Scale Wang Tiles}. In
  \bibinfo{booktitle}{\emph{Comp. Graph. Forum}}, Vol.~\bibinfo{volume}{41}.
  \bibinfo{pages}{149--159}.
\newblock


\bibitem[Liu et~al\mbox{.}(2020)]%
        {Liu:2020:OEO}
\bibfield{author}{\bibinfo{person}{Xihui Liu}, \bibinfo{person}{Zhe Lin},
  \bibinfo{person}{Jianming Zhang}, \bibinfo{person}{Handong Zhao},
  \bibinfo{person}{Quan Tran}, \bibinfo{person}{Xiaogang Wang}, {and}
  \bibinfo{person}{Hongsheng Li}.} \bibinfo{year}{2020}\natexlab{}.
\newblock \showarticletitle{Open-edit: Open-domain image manipulation with
  open-vocabulary instructions}. In \bibinfo{booktitle}{\emph{ECCV}}.
  \bibinfo{pages}{89--106}.
\newblock


\bibitem[Mardani et~al\mbox{.}(2020)]%
        {Nardani:2020:NFU}
\bibfield{author}{\bibinfo{person}{Morteza Mardani}, \bibinfo{person}{Guilin
  Liu}, \bibinfo{person}{Aysegul Dundar}, \bibinfo{person}{Shiqiu Liu},
  \bibinfo{person}{Andrew Tao}, {and} \bibinfo{person}{Bryan Catanzaro}.}
  \bibinfo{year}{2020}\natexlab{}.
\newblock \showarticletitle{Neural ffts for universal texture image synthesis}.
\newblock \bibinfo{journal}{\emph{NeurIPS}}  \bibinfo{volume}{33}
  (\bibinfo{year}{2020}), \bibinfo{pages}{14081--14092}.
\newblock


\bibitem[Mokady et~al\mbox{.}(2023)]%
        {Mokady:2023:NTI}
\bibfield{author}{\bibinfo{person}{Ron Mokady}, \bibinfo{person}{Amir Hertz},
  \bibinfo{person}{Kfir Aberman}, \bibinfo{person}{Yael Pritch}, {and}
  \bibinfo{person}{Daniel Cohen-Or}.} \bibinfo{year}{2023}\natexlab{}.
\newblock \showarticletitle{Null-text inversion for editing real images using
  guided diffusion models}. In \bibinfo{booktitle}{\emph{CVPR}}.
  \bibinfo{pages}{6038--6047}.
\newblock


\bibitem[Moritz et~al\mbox{.}(2017)]%
        {Moritz:2017:TST}
\bibfield{author}{\bibinfo{person}{Joep Moritz}, \bibinfo{person}{Stuart
  James}, \bibinfo{person}{Tom S.~F. Haines}, \bibinfo{person}{Tobias
  Ritschel}, {and} \bibinfo{person}{Tim Weyrich}.}
  \bibinfo{year}{2017}\natexlab{}.
\newblock \showarticletitle{Texture Stationarization: Turning Photos Into
  Tileable Textures}.
\newblock \bibinfo{journal}{\emph{Comp. Graph. Forum}} \bibinfo{volume}{36},
  \bibinfo{number}{2} (\bibinfo{year}{2017}), \bibinfo{pages}{177--188}.
\newblock


\bibitem[Ng et~al\mbox{.}(2005)]%
        {Ng:2005:GWT}
\bibfield{author}{\bibinfo{person}{T-Y Ng}, \bibinfo{person}{T-S Tan}, {and}
  \bibinfo{person}{Xinyu Zhang}.} \bibinfo{year}{2005}\natexlab{}.
\newblock \showarticletitle{Generating $\omega$-tile set for texture
  synthesis}. In \bibinfo{booktitle}{\emph{IEEE Int. Comp. Graph.}}
  \bibinfo{pages}{177--184}.
\newblock


\bibitem[Nichol et~al\mbox{.}(2022)]%
        {Nichol:2022:GTP}
\bibfield{author}{\bibinfo{person}{Alexander~Quinn Nichol},
  \bibinfo{person}{Prafulla Dhariwal}, \bibinfo{person}{Aditya Ramesh},
  \bibinfo{person}{Pranav Shyam}, \bibinfo{person}{Pamela Mishkin},
  \bibinfo{person}{Bob Mcgrew}, \bibinfo{person}{Ilya Sutskever}, {and}
  \bibinfo{person}{Mark Chen}.} \bibinfo{year}{2022}\natexlab{}.
\newblock \showarticletitle{GLIDE: Towards Photorealistic Image Generation and
  Editing with Text-Guided Diffusion Models}. In
  \bibinfo{booktitle}{\emph{ICML}}. \bibinfo{pages}{16784--16804}.
\newblock


\bibitem[Nikankin et~al\mbox{.}(2023)]%
        {Nikankin:2023:STD}
\bibfield{author}{\bibinfo{person}{Yaniv Nikankin}, \bibinfo{person}{Niv Haim},
  {and} \bibinfo{person}{Michal Irani}.} \bibinfo{year}{2023}\natexlab{}.
\newblock \showarticletitle{SinFusion: training diffusion models on a single
  image or video}. In \bibinfo{booktitle}{\emph{ICML}}.
  \bibinfo{pages}{26199--26214}.
\newblock


\bibitem[Paszke et~al\mbox{.}(2019)]%
        {Paszke:2019:Pytorch}
\bibfield{author}{\bibinfo{person}{Adam Paszke}, \bibinfo{person}{Sam Gross},
  \bibinfo{person}{Francisco Massa}, \bibinfo{person}{Adam Lerer},
  \bibinfo{person}{James Bradbury}, \bibinfo{person}{Gregory Chanan},
  \bibinfo{person}{Trevor Killeen}, \bibinfo{person}{Zeming Lin},
  \bibinfo{person}{Natalia Gimelshein}, \bibinfo{person}{Luca Antiga},
  {et~al\mbox{.}}} \bibinfo{year}{2019}\natexlab{}.
\newblock \showarticletitle{Pytorch: An imperative style, high-performance deep
  learning library}.
\newblock \bibinfo{journal}{\emph{NeurIPS}}  \bibinfo{volume}{32}
  (\bibinfo{year}{2019}).
\newblock


\bibitem[Perlin(1985)]%
        {Perlin:1985:AIS}
\bibfield{author}{\bibinfo{person}{Ken Perlin}.}
  \bibinfo{year}{1985}\natexlab{}.
\newblock \showarticletitle{An image synthesizer}.
\newblock \bibinfo{journal}{\emph{ACM Siggraph Computer Graphics}}
  \bibinfo{volume}{19}, \bibinfo{number}{3} (\bibinfo{year}{1985}),
  \bibinfo{pages}{287--296}.
\newblock


\bibitem[Portenier et~al\mbox{.}(2020)]%
        {Portenier:2020:GD3}
\bibfield{author}{\bibinfo{person}{Tiziano Portenier}, \bibinfo{person}{Siavash
  Arjomand~Bigdeli}, {and} \bibinfo{person}{Orcun Goksel}.}
  \bibinfo{year}{2020}\natexlab{}.
\newblock \showarticletitle{Gram{GAN}: Deep 3d texture synthesis from 2d
  exemplars}.
\newblock \bibinfo{journal}{\emph{NeurIPS}}  \bibinfo{volume}{33}
  (\bibinfo{year}{2020}), \bibinfo{pages}{6994--7004}.
\newblock


\bibitem[Ramesh et~al\mbox{.}(2022)]%
        {Ramesh:2022:HTC}
\bibfield{author}{\bibinfo{person}{Aditya Ramesh}, \bibinfo{person}{Prafulla
  Dhariwal}, \bibinfo{person}{Alex Nichol}, \bibinfo{person}{Casey Chu}, {and}
  \bibinfo{person}{Mark Chen}.} \bibinfo{year}{2022}\natexlab{}.
\newblock \showarticletitle{Hierarchical Text-Conditional Image Generation with
  CLIP Latents}.
\newblock \bibinfo{journal}{\emph{arXiv preprint arXiv:2204.06125}}
  (\bibinfo{year}{2022}).
\newblock


\bibitem[Richardson et~al\mbox{.}(2023)]%
        {Richardson:2023:TTG}
\bibfield{author}{\bibinfo{person}{Elad Richardson}, \bibinfo{person}{Gal
  Metzer}, \bibinfo{person}{Yuval Alaluf}, \bibinfo{person}{Raja Giryes}, {and}
  \bibinfo{person}{Daniel Cohen-Or}.} \bibinfo{year}{2023}\natexlab{}.
\newblock \showarticletitle{Texture: Text-guided texturing of 3d shapes}.
\newblock \bibinfo{journal}{\emph{arXiv preprint arXiv:2302.01721}}
  (\bibinfo{year}{2023}).
\newblock


\bibitem[Rodriguez-Pardo et~al\mbox{.}(2024)]%
        {Rodriguez-Pardo:2024:TAD}
\bibfield{author}{\bibinfo{person}{Carlos Rodriguez-Pardo},
  \bibinfo{person}{Dan Casas}, \bibinfo{person}{Elena Garces}, {and}
  \bibinfo{person}{Jorge Lopez-Moreno}.} \bibinfo{year}{2024}\natexlab{}.
\newblock \showarticletitle{TexTile: A Differentiable Metric for Texture
  Tileability}. In \bibinfo{booktitle}{\emph{CVPR}}.
\newblock


\bibitem[Rodriguez-Pardo and Garces(2022)]%
        {Rodriguez:2022:SSS}
\bibfield{author}{\bibinfo{person}{Carlos Rodriguez-Pardo} {and}
  \bibinfo{person}{Elena Garces}.} \bibinfo{year}{2022}\natexlab{}.
\newblock \showarticletitle{Seamlessgan: Self-supervised synthesis of tileable
  texture maps}.
\newblock \bibinfo{journal}{\emph{IEEE Trans. Vis. and Comp. Graph.}}
  (\bibinfo{year}{2022}).
\newblock


\bibitem[Rodriguez-Pardo et~al\mbox{.}(2019)]%
        {Rodriguez:2019:AES}
\bibfield{author}{\bibinfo{person}{Carlos Rodriguez-Pardo},
  \bibinfo{person}{Sergio Suja}, \bibinfo{person}{David Pascual},
  \bibinfo{person}{Jorge Lopez-Moreno}, {and} \bibinfo{person}{Elena Garces}.}
  \bibinfo{year}{2019}\natexlab{}.
\newblock \showarticletitle{Automatic extraction and synthesis of regular
  repeatable patterns}.
\newblock \bibinfo{journal}{\emph{Computers \& Graphics}}  \bibinfo{volume}{83}
  (\bibinfo{year}{2019}), \bibinfo{pages}{33--41}.
\newblock


\bibitem[Rombach et~al\mbox{.}(2022)]%
        {Rombach:2022:HRI}
\bibfield{author}{\bibinfo{person}{Robin Rombach}, \bibinfo{person}{Andreas
  Blattmann}, \bibinfo{person}{Dominik Lorenz}, \bibinfo{person}{Patrick
  Esser}, {and} \bibinfo{person}{Bj\"orn Ommer}.}
  \bibinfo{year}{2022}\natexlab{}.
\newblock \showarticletitle{High-Resolution Image Synthesis With Latent
  Diffusion Models}. In \bibinfo{booktitle}{\emph{CVPR}}.
  \bibinfo{pages}{10684--10695}.
\newblock


\bibitem[Saharia et~al\mbox{.}(2022)]%
        {Saharia:2022:PTI}
\bibfield{author}{\bibinfo{person}{Chitwan Saharia}, \bibinfo{person}{William
  Chan}, \bibinfo{person}{Saurabh Saxena}, \bibinfo{person}{Lala Li},
  \bibinfo{person}{Jay Whang}, \bibinfo{person}{Emily~L Denton},
  \bibinfo{person}{Kamyar Ghasemipour}, \bibinfo{person}{Raphael
  Gontijo~Lopes}, \bibinfo{person}{Burcu Karagol~Ayan}, \bibinfo{person}{Tim
  Salimans}, {et~al\mbox{.}}} \bibinfo{year}{2022}\natexlab{}.
\newblock \showarticletitle{Photorealistic text-to-image diffusion models with
  deep language understanding}.
\newblock \bibinfo{journal}{\emph{NeurIPS}}  \bibinfo{volume}{35}
  (\bibinfo{year}{2022}), \bibinfo{pages}{36479--36494}.
\newblock


\bibitem[Salimans et~al\mbox{.}(2016)]%
        {Salimans:2016:IS}
\bibfield{author}{\bibinfo{person}{Tim Salimans}, \bibinfo{person}{Ian
  Goodfellow}, \bibinfo{person}{Wojciech Zaremba}, \bibinfo{person}{Vicki
  Cheung}, \bibinfo{person}{Alec Radford}, {and} \bibinfo{person}{Xi Chen}.}
  \bibinfo{year}{2016}\natexlab{}.
\newblock \showarticletitle{Improved techniques for training gans}.
\newblock \bibinfo{journal}{\emph{NeurIPS}}  \bibinfo{volume}{29}
  (\bibinfo{year}{2016}).
\newblock


\bibitem[Sartor and Peers(2023)]%
        {Sartor:2023:MGD}
\bibfield{author}{\bibinfo{person}{Sam Sartor} {and} \bibinfo{person}{Pieter
  Peers}.} \bibinfo{year}{2023}\natexlab{}.
\newblock \showarticletitle{MatFusion: A Generative Diffusion Model for SVBRDF
  Capture}. In \bibinfo{booktitle}{\emph{SIGGRAPH Asia 2023 Conference
  Papers}}. \bibinfo{pages}{1--10}.
\newblock


\bibitem[Shaham et~al\mbox{.}(2019)]%
        {Shaham:2019:SLG}
\bibfield{author}{\bibinfo{person}{Tamar~Rott Shaham}, \bibinfo{person}{Tali
  Dekel}, {and} \bibinfo{person}{Tomer Michaeli}.}
  \bibinfo{year}{2019}\natexlab{}.
\newblock \showarticletitle{Singan: Learning a generative model from a single
  natural image}. In \bibinfo{booktitle}{\emph{CVPR}}.
  \bibinfo{pages}{4570--4580}.
\newblock


\bibitem[Song et~al\mbox{.}(2021)]%
        {Song:2021:SBG}
\bibfield{author}{\bibinfo{person}{Yang Song}, \bibinfo{person}{Jascha
  Sohl-Dickstein}, \bibinfo{person}{Diederik~P Kingma},
  \bibinfo{person}{Abhishek Kumar}, \bibinfo{person}{Stefano Ermon}, {and}
  \bibinfo{person}{Ben Poole}.} \bibinfo{year}{2021}\natexlab{}.
\newblock \showarticletitle{Score-Based Generative Modeling through Stochastic
  Differential Equations}. In \bibinfo{booktitle}{\emph{ICLR}}.
\newblock


\bibitem[{Stability AI}(2022a)]%
        {StableDiffusionInpainting}
\bibfield{author}{\bibinfo{person}{{Stability AI}}.}
  \bibinfo{year}{2022}\natexlab{a}.
\newblock \bibinfo{title}{Stable Diffusion V2 - Inpainting}.
\newblock
\newblock
\newblock
\shownote{\url{https://huggingface.co/stabilityai/stable-diffusion-2-inpainting}}.


\bibitem[{Stability AI}(2022b)]%
        {StableDiffusionXL}
\bibfield{author}{\bibinfo{person}{{Stability AI}}.}
  \bibinfo{year}{2022}\natexlab{b}.
\newblock \bibinfo{title}{Stable Diffusion XL}.
\newblock
\newblock
\newblock
\shownote{\url{https://huggingface.co/docs/diffusers/using-diffusers/sdxl}}.


\bibitem[Tumanyan et~al\mbox{.}(2023)]%
        {Tumanyan:2023:PPD}
\bibfield{author}{\bibinfo{person}{Narek Tumanyan}, \bibinfo{person}{Michal
  Geyer}, \bibinfo{person}{Shai Bagon}, {and} \bibinfo{person}{Tali Dekel}.}
  \bibinfo{year}{2023}\natexlab{}.
\newblock \showarticletitle{Plug-and-Play Diffusion Features for Text-Driven
  Image-to-Image Translation}. In \bibinfo{booktitle}{\emph{CVPR}}.
  \bibinfo{pages}{1921--1930}.
\newblock


\bibitem[Ulyanov et~al\mbox{.}(2016)]%
        {Ulyanov:2016:TNF}
\bibfield{author}{\bibinfo{person}{Dmitry Ulyanov}, \bibinfo{person}{Vadim
  Lebedev}, \bibinfo{person}{Andrea Vedaldi}, {and} \bibinfo{person}{Victor
  Lempitsky}.} \bibinfo{year}{2016}\natexlab{}.
\newblock \showarticletitle{Texture networks: feed-forward synthesis of
  textures and stylized images}. In \bibinfo{booktitle}{\emph{ICML}}.
  \bibinfo{pages}{1349--1357}.
\newblock


\bibitem[Ulyanov et~al\mbox{.}(2017)]%
        {Ulyanov:2017:ITN}
\bibfield{author}{\bibinfo{person}{Dmitry Ulyanov}, \bibinfo{person}{Andrea
  Vedaldi}, {and} \bibinfo{person}{Victor Lempitsky}.}
  \bibinfo{year}{2017}\natexlab{}.
\newblock \showarticletitle{Improved texture networks: Maximizing quality and
  diversity in feed-forward stylization and texture synthesis}. In
  \bibinfo{booktitle}{\emph{CVRP}}. \bibinfo{pages}{6924--6932}.
\newblock


\bibitem[Vanhoey et~al\mbox{.}(2013)]%
        {Vanhoey:2013:OMS}
\bibfield{author}{\bibinfo{person}{Kenneth Vanhoey}, \bibinfo{person}{Basile
  Sauvage}, \bibinfo{person}{Fr{\'e}d{\'e}ric Larue}, {and}
  \bibinfo{person}{Jean-Michel Dischler}.} \bibinfo{year}{2013}\natexlab{}.
\newblock \showarticletitle{On-the-fly multi-scale infinite texturing from
  example}.
\newblock \bibinfo{journal}{\emph{ACM Trans. Graph.}} \bibinfo{volume}{32},
  \bibinfo{number}{6} (\bibinfo{year}{2013}).
\newblock


\bibitem[Vecchio et~al\mbox{.}(2023)]%
        {Vecchio:2023:CAC}
\bibfield{author}{\bibinfo{person}{Giuseppe Vecchio}, \bibinfo{person}{Rosalie
  Martin}, \bibinfo{person}{Arthur Roullier}, \bibinfo{person}{Adrien Kaiser},
  \bibinfo{person}{Romain Rouffet}, \bibinfo{person}{Valentin Deschaintre},
  {and} \bibinfo{person}{Tamy Boubekeur}.} \bibinfo{year}{2023}\natexlab{}.
\newblock \showarticletitle{ControlMat: A Controlled Generative Approach to
  Material Capture}.
\newblock \bibinfo{journal}{\emph{arXiv preprint arXiv:2309.01700}}
  (\bibinfo{year}{2023}).
\newblock


\bibitem[Voynov et~al\mbox{.}(2023)]%
        {Voynov:2023:SGT}
\bibfield{author}{\bibinfo{person}{Andrey Voynov}, \bibinfo{person}{Kfir
  Aberman}, {and} \bibinfo{person}{Daniel Cohen-Or}.}
  \bibinfo{year}{2023}\natexlab{}.
\newblock \showarticletitle{Sketch-Guided Text-to-Image Diffusion Models}. In
  \bibinfo{booktitle}{\emph{ACM SIGGRAPH 2023 Conference Proceedings}}. Article
  \bibinfo{articleno}{55}, \bibinfo{numpages}{11}~pages.
\newblock


\bibitem[\v{S}ubrtov\'{a} et~al\mbox{.}(2023)]%
        {Subrtova:2023:DIA}
\bibfield{author}{\bibinfo{person}{Ad\'{e}la \v{S}ubrtov\'{a}},
  \bibinfo{person}{Michal Luk\'{a}\v{c}}, \bibinfo{person}{Jan \v{C}ech},
  \bibinfo{person}{David Futschik}, \bibinfo{person}{Eli Shechtman}, {and}
  \bibinfo{person}{Daniel S\'{y}kora}.} \bibinfo{year}{2023}\natexlab{}.
\newblock \showarticletitle{Diffusion Image Analogies}. In
  \bibinfo{booktitle}{\emph{ACM SIGGRAPH 2023 Conference Proceedings}}. Article
  \bibinfo{articleno}{79}, \bibinfo{numpages}{10}~pages.
\newblock


\bibitem[Wang(1961)]%
        {Wang:1961:PTP}
\bibfield{author}{\bibinfo{person}{Hao Wang}.} \bibinfo{year}{1961}\natexlab{}.
\newblock \showarticletitle{Proving theorems by pattern recognition—II}.
\newblock \bibinfo{journal}{\emph{Bell system technical journal}}
  \bibinfo{volume}{40}, \bibinfo{number}{1} (\bibinfo{year}{1961}),
  \bibinfo{pages}{1--41}.
\newblock


\bibitem[Wang et~al\mbox{.}(2023)]%
        {Wang:2023:ECA}
\bibfield{author}{\bibinfo{person}{Jianyi Wang}, \bibinfo{person}{Kelvin~C.K.
  Chan}, {and} \bibinfo{person}{Chen~Change Loy}.}
  \bibinfo{year}{2023}\natexlab{}.
\newblock \showarticletitle{Exploring CLIP for assessing the look and feel of
  images}. In \bibinfo{booktitle}{\emph{AAAI}}. Article
  \bibinfo{articleno}{284}, \bibinfo{numpages}{9}~pages.
\newblock


\bibitem[Wei(2004)]%
        {Wei:2004:TBT}
\bibfield{author}{\bibinfo{person}{Li-Yi Wei}.}
  \bibinfo{year}{2004}\natexlab{}.
\newblock \showarticletitle{Tile-based texture mapping on graphics hardware}.
  In \bibinfo{booktitle}{\emph{Proceedings of the ACM SIGGRAPH/EUROGRAPHICS
  conference on Graphics hardware}}. \bibinfo{pages}{55--63}.
\newblock


\bibitem[Xiang et~al\mbox{.}(2023)]%
        {Xiang:2023:3IG}
\bibfield{author}{\bibinfo{person}{Jianfeng Xiang}, \bibinfo{person}{Jiaolong
  Yang}, \bibinfo{person}{Binbin Huang}, {and} \bibinfo{person}{Xin Tong}.}
  \bibinfo{year}{2023}\natexlab{}.
\newblock \showarticletitle{3D-aware Image Generation using 2D Diffusion
  Models}.
\newblock \bibinfo{journal}{\emph{arXiv preprint arXiv:2303.17905}}
  (\bibinfo{year}{2023}).
\newblock


\bibitem[Xu et~al\mbox{.}(2023)]%
        {Xu:2023:MMT}
\bibfield{author}{\bibinfo{person}{Xudong Xu}, \bibinfo{person}{Zhaoyang Lyu},
  \bibinfo{person}{Xingang Pan}, {and} \bibinfo{person}{Bo Dai}.}
  \bibinfo{year}{2023}\natexlab{}.
\newblock \showarticletitle{Matlaber: Material-aware text-to-3d via latent brdf
  auto-encoder}.
\newblock \bibinfo{journal}{\emph{arXiv preprint arXiv:2308.09278}}
  (\bibinfo{year}{2023}).
\newblock


\bibitem[Ye et~al\mbox{.}(2023)]%
        {Ye:2023:IAT}
\bibfield{author}{\bibinfo{person}{Hu Ye}, \bibinfo{person}{Jun Zhang},
  \bibinfo{person}{Sibo Liu}, \bibinfo{person}{Xiao Han}, {and}
  \bibinfo{person}{Wei Yang}.} \bibinfo{year}{2023}\natexlab{}.
\newblock \showarticletitle{IP-Adapter: Text Compatible Image Prompt Adapter
  for Text-to-Image Diffusion Models}.
\newblock \bibinfo{journal}{\emph{arXiv preprint arXiv:2308.06721}}
  (\bibinfo{year}{2023}).
\newblock


\bibitem[Yu et~al\mbox{.}(2019)]%
        {Yu:2019:TMN}
\bibfield{author}{\bibinfo{person}{Ning Yu}, \bibinfo{person}{Connelly Barnes},
  \bibinfo{person}{Eli Shechtman}, \bibinfo{person}{Sohrab Amirghodsi}, {and}
  \bibinfo{person}{Michal Lukac}.} \bibinfo{year}{2019}\natexlab{}.
\newblock \showarticletitle{Texture mixer: A network for controllable synthesis
  and interpolation of texture}. In \bibinfo{booktitle}{\emph{CVPR}}.
  \bibinfo{pages}{12164--12173}.
\newblock


\bibitem[Zeng et~al\mbox{.}(2023)]%
        {Zeng:2023:PPA}
\bibfield{author}{\bibinfo{person}{Xianfang Zeng}, \bibinfo{person}{Xin Chen},
  \bibinfo{person}{Zhongqi Qi}, \bibinfo{person}{Wen Liu},
  \bibinfo{person}{Zibo Zhao}, \bibinfo{person}{Zhibin Wang},
  \bibinfo{person}{Bin Fu}, \bibinfo{person}{Yong Liu}, {and}
  \bibinfo{person}{Gang Yu}.} \bibinfo{year}{2023}\natexlab{}.
\newblock \showarticletitle{Paint3D: Paint Anything 3D with Lighting-Less
  Texture Diffusion Models}.
\newblock \bibinfo{journal}{\emph{arXiv preprint arXiv:2312.13913}}
  (\bibinfo{year}{2023}).
\newblock


\bibitem[Zhang et~al\mbox{.}(2023)]%
        {Zhang:2023:ACC}
\bibfield{author}{\bibinfo{person}{Lvmin Zhang}, \bibinfo{person}{Anyi Rao},
  {and} \bibinfo{person}{Maneesh Agrawala}.} \bibinfo{year}{2023}\natexlab{}.
\newblock \showarticletitle{Adding conditional control to text-to-image
  diffusion models}. In \bibinfo{booktitle}{\emph{CVPR}}.
  \bibinfo{pages}{3836--3847}.
\newblock


\bibitem[Zhou et~al\mbox{.}(2022)]%
        {Zhou:2022:TCM}
\bibfield{author}{\bibinfo{person}{Xilong Zhou}, \bibinfo{person}{Milos Hasan},
  \bibinfo{person}{Valentin Deschaintre}, \bibinfo{person}{Paul Guerrero},
  \bibinfo{person}{Kalyan Sunkavalli}, {and} \bibinfo{person}{Nima~Khademi
  Kalantari}.} \bibinfo{year}{2022}\natexlab{}.
\newblock \showarticletitle{TileGen: Tileable, Controllable Material Generation
  and Capture}. In \bibinfo{booktitle}{\emph{SIGGRAPH Asia 2022 Conference
  Papers}}. Article \bibinfo{articleno}{34}, \bibinfo{numpages}{9}~pages.
\newblock


\bibitem[Zhou et~al\mbox{.}(2023)]%
        {Zhou:2023:NTS}
\bibfield{author}{\bibinfo{person}{Yang Zhou}, \bibinfo{person}{Kaijian Chen},
  \bibinfo{person}{Rongjun Xiao}, {and} \bibinfo{person}{Hui Huang}.}
  \bibinfo{year}{2023}\natexlab{}.
\newblock \showarticletitle{Neural Texture Synthesis With Guided
  Correspondence}. In \bibinfo{booktitle}{\emph{CVPR}}.
  \bibinfo{pages}{18095--18104}.
\newblock


\bibitem[Zhou et~al\mbox{.}(2018)]%
        {Zhou:2018:NST}
\bibfield{author}{\bibinfo{person}{Yang Zhou}, \bibinfo{person}{Zhen Zhu},
  \bibinfo{person}{Xiang Bai}, \bibinfo{person}{Dani Lischinski},
  \bibinfo{person}{Daniel Cohen-Or}, {and} \bibinfo{person}{Hui Huang}.}
  \bibinfo{year}{2018}\natexlab{}.
\newblock \showarticletitle{Non-stationary texture synthesis by adversarial
  expansion}.
\newblock \bibinfo{journal}{\emph{ACM Trans. Graph.}} \bibinfo{volume}{37},
  \bibinfo{number}{4} (\bibinfo{year}{2018}), \bibinfo{pages}{1--13}.
\newblock


\end{thebibliography}

\appendix
\section{Dual Wang Tile Packing}
\label{sec:packing}
\begin{figure}
  \begin{tabular}{cc}
    \includegraphics[width=.45\columnwidth]{2color_withtiles.pdf} &
    \includegraphics[width=.45\columnwidth]{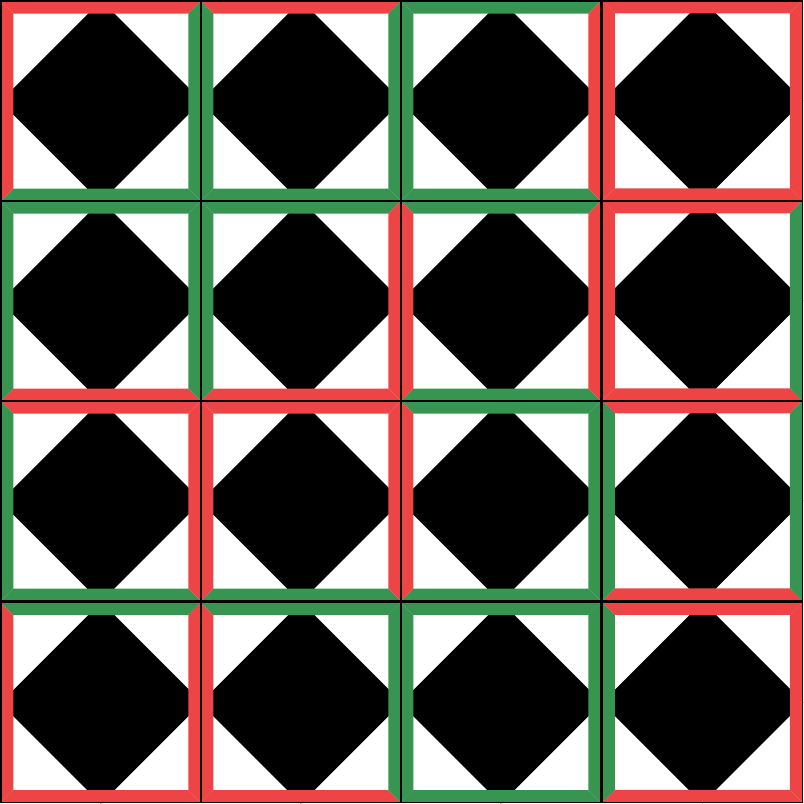}
  \end{tabular}                                                                                    
  \caption{Left: an example of a $2$-color Dual Wang tile packing that
    includes all possible interior tiles (white) and cross tiles
    (black). Note, the cross tiles at the edges continue on the
    opposite side (i.e., wrap around edges). Right: the corresponding
    complement of the $2$-color Dual Wang tile packing where each edge
    is replaced by its perpendicular edge.  Note that the complement
    is shifted by half a tile, and that the role of interior and cross
    tiles have swapped. For example, the colors of the edges incident
    on the first complete black diamond in the Dual Wang tile packing
    (left) are \emph{red, green, green, red} (N-E-S-W), which
    corresponds to the first tile in the complement (right).}
  \label{fig:packing_example}
\end{figure}

A Dual Wang tile packing is a $\C^2 \times \C^2$ Wang tiling such that
each interior and cross Dual Wang tile occurs exactly once, which
ensures minimum storage requirements as well as no visible seams.  A
Dual Wang tile packing is a specialization of a regular Wang tile
packing as illustrated in~\autoref{fig:packing_example} (left), where
the white diamonds are the \emph{interior} Dual Wang tiles that occupy
the center of each regular Wang tile, and the black diamonds are the
\emph{cross} Dual Wang tiles that straddle the boundaries of four
regular Wang tiles.  Note that the cross tiles at the boundaries of
the packing continue on the opposite side (\ie \emph{wrap-around}
texture), and thus the packing itself is also tileable.  Each interior
Dual Wang tile is identified by the 4 colors of the encompassing
regular Wang tile, and each cross Dual Wang tile is identified by the
colors of the four edges incident at the cross tile's center.

Brute force computation of a (Dual) Wang tile packing is NP-complete,
and only practical for tilings with few colors.  For $\C = 2$, we
found that there exist $9,\!408$ possible packings.  For tilings with
more colors, even more Dual Wang tile packings exists.  However, we
are not interested in enumerating each possible packing, but only
desire a fast algorithm for generating a single Dual Wang tile packing
for any given number of colors.  Wei~\shortcite{Wei:2004:TBT} showed
that such algorithm exists for regular Wang tile sets, but while Wei's
packing algorithm includes all interior Dual Wang tiles, does not
guarantee a complete packing for cross Dual Wang tiles.

To derive a Dual Wang tile packing algorithm, we will first introduce
some necessary definitions and tools (\autoref{sec:definitions}) to
aid in defining a generative algorithm.  Next, we introduce a packing
algorithm for Dual Wang tiles with an odd number of colors
(\autoref{sec:odd}), followed by a algorithm for Dual Wang tiles with
an even number of colors (\autoref{sec:even}).

\begin{figure*}[!h]
  \def\tilespace{2.4mm}
  \begin{tabular}{cl}
    \multirow{2}{*}{{\rotatebox[origin=c]{90}{2 color \hspace{-1cm}}}} &
    \includegraphics[height=1cm]{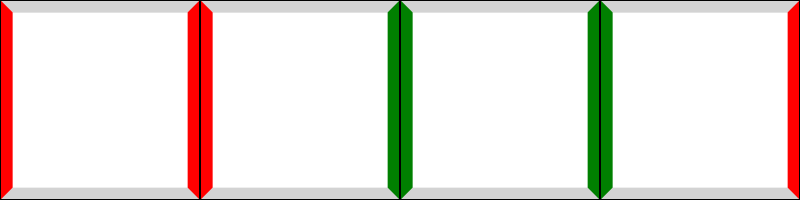} \\
    &
      \hspace{\tilespace}\{0,0\}, \hspace{\tilespace}\{0, 1\}, \hspace{\tilespace}\{1, 1\}, \hspace{\tilespace}\{1, 0\} \\

    \multirow{2}{*}{{\rotatebox[origin=c]{90}{3 color \hspace{-1cm}}}} &
    \includegraphics[height=1cm]{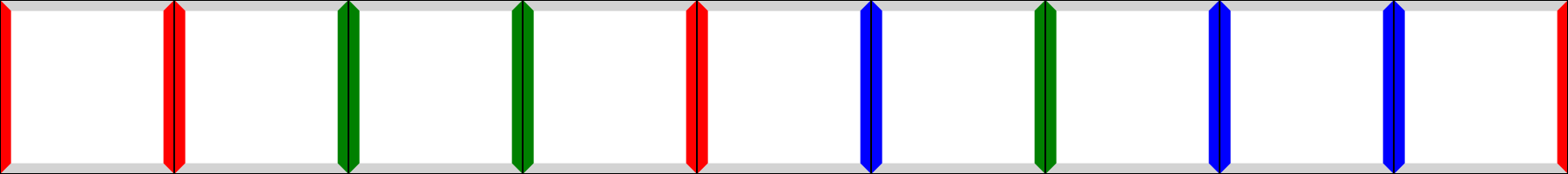} \\
                                                                       &
      \hspace{\tilespace}\{0,0\}, \hspace{\tilespace}\{0, 1\}, \hspace{\tilespace}\{1, 1\}, \hspace{\tilespace}\{1, 0\}, \hspace{\tilespace}\{0, 2\}, \hspace{\tilespace}\{2, 1\}, \hspace{\tilespace}\{1, 2\}, \hspace{\tilespace}\{2, 2\}, \hspace{\tilespace}\{2,0\} \\
    
    \multirow{2}{*}{{\rotatebox[origin=c]{90}{4 color \hspace{-1cm}}}} &
    \includegraphics[height=1cm]{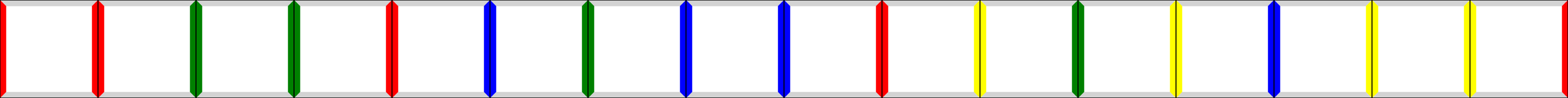} \\
    &
      \hspace{\tilespace}\{0,0\}, \hspace{\tilespace}\{0, 1\}, \hspace{\tilespace}\{1, 1\}, \hspace{\tilespace}\{1, 0\}, \hspace{\tilespace}\{0, 2\}, \hspace{\tilespace}\{2, 1\}, \hspace{\tilespace}\{1, 2\}, \hspace{\tilespace}\{2, 2\}, \hspace{\tilespace}\{2,0\}. \hspace{\tilespace}\{0, 3\}, \hspace{\tilespace}\{3, 1\}, \hspace{\tilespace}\{1, 3\}, \hspace{\tilespace}\{3, 2\}, \hspace{\tilespace}\{2, 3\}, \hspace{\tilespace}\{3, 3\}, \hspace{\tilespace}\{3, 0\}
  \end{tabular}
  \caption{Domino strings generated using~\autoref{alg:domino} for
    $2, 3$ and $4$ colors. For clarity, we also list the (numerical
    representation of) color-pairs for each domino tile. Note that
    each sequence extends the previous sequence.}
  \label{fig:domino}
\end{figure*}

\subsection{Definitions}
\label{sec:definitions}

We will denote a 2D tiling as $\T(\x, \y)$ that for each 2D coordinate
$(\x, \y)$ returns four colors ${\c_\N, \c_\E, \c_\S, \c_\W}$ for the
north, east, south, and west edge respectively.  Furthermore, we
assume the tiling wraps around at the edges.  For a 2D tiling with
$\C$ colors, we therefore have:
\begin{equation}
  \T(\x, \y) = \T(\x \mod \C^2, \y \mod \C^2).
\end{equation}
A tiling $\T$ is complete if it contains all $\C^4$ tiles that can be
formed with $\C$ colored edges (i.e., each tile occurs exactly once),
and that are tiled such that the edge colors of neighboring tiles
match.

We define the complement $\bar{\T}$ of a 2D Wang tiling as the 2D Wang
tiling where each edge is replaced by its perpendicular edge going
through the edge's center. The complement is related to a regular
tiling by:
\begin{equation}
  \bar{\T}(\x, \y) = \{ \T(\x,\y)_\E, \T(\x+1, \y)_\S, \T(\x, \y+1)_\E, \T(\x, \y)_S \}.
\end{equation}
\autoref{fig:packing_example} (right) shows the complement of a
$2$-color Dual Wang tile packing. Note that the role of interior and
cross tiles switches between the regular and complement.
Consequently, a $\C^2 \times \C^2$ tiling $\T$ is a $\C$-color Dual
Wang tile packing if both $\T$ and $\bar{\T}$ are complete (i.e., it
contains exactly all $\C^4$ interior and $\C^4$ cross tiles once).

To aid in defining the generative algorithms we will also consider a
1D domino string (i.e., a tiling with only 2 edges) denoted by
$\D(\x) = \{\c_\W(\x), \c_\E(\x)\}$. We will also assume that these 1D
domino strings wrap around at the outer edges.  We will differentiate
between horizontal and vertical domino strings when tiled in a 2D
domain as $\D(\x)$ and $\D(\y)$ respectively.  A $\C^2$-length domino
string with $\C$ colors is complete if it contains all $\C^2$ possible
domino tiles (i.e., each tile occurs exactly once).

Finally, each $\C^2 \times \C^2$ tiling can be written as a
combination of $\C^2$ different horizontal and vertical $\C^2$-length
domino strings:
\begin{equation}
  \T(\x, \y) = \{ \D_\x(\y)_N, \D_\y(\x)_\E, \D_\x(\y)_\S, \D_\y(\x)_W \},
\end{equation}
where $\D_\x$ and $\D_\y$ indicate the
$\x$-th vertical domino string and
$\y$-th horizontal domino string respectively.  Given the tiling
$\T$, we can define the complementary domino strings
$\bar{\D}_\x$ and
$\bar{\D}_\y$ as the corresponding horizontal and vertical domino
strings that constitute the complement tiling $\bar{\T}$.

Our goal for developing a generative algorithm is to derive a fast
algorithm for computing a valid sequence $\D_\x$ and $\D_\y$ such that
the resulting tiling is a Dual Wang tile packing.

\subsection{Odd Color Dual Wang Tile Packing}
\label{sec:odd}

We will first start by deriving a generative algorithm for a Dual Wang
tile packing with an odd number of colors $\C$.

\begin{figure}
  \includegraphics[width=.45\columnwidth]{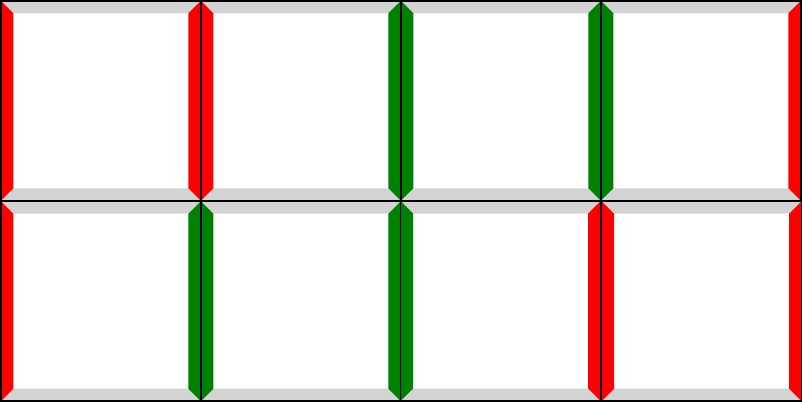}
  \caption{Shifting the top row domino string one position to the left
    in the subsequent row, produces a complete sequence of vertical
    edge combinations. Note, the \emph{red-red} combination occurs at
    the edges, followed by (left to right), \emph{red-green},
    \emph{green-green}, and \emph{green-red}, which exactly matches
    the domino-string in the top-row.}
  \label{fig:shift}
\end{figure}

We start with the following key observation. If $\D(\x)$ is a complete
$\C^2$-length domino string and given a tiling $\T(\x,\y)$, where
$\D_\y(\x) = \D(\x \pm \y)$, then each $\bar{\D}_\x(\y)$ is also a
complete domino string, and:
\begin{equation}
  \bar{\D}_\x(\y) = \bar{\D}(\y \pm \x) = \D'(\x \pm \y),
\label{eq:shift_dual}
\end{equation}
where $\D'$ is identical to $\D$ when shifting in the positive
direction, and mirrored when shifting in the negative direction.
\autoref{eq:shift_dual} follows trivially from the fact that
$\D(\x)_\E = \D(\x + 1)_\W$, and $\D(x - 1)_\E = \D(\x)_\W$, and thus
the domino formed by the colors from the corresponding edges in
neighboring domino strings (shifted by 1) corresponds to the domino at
the corresponding position in the original domino string.  Since the
domino string is complete, it follows that the string formed by
corresponding edges must therefore also be complete
(\autoref{fig:shift}).

However, complete horizontal and vertical domino strings are not a
sufficient condition for the resulting tiling to be complete.  In
order for the resulting (shifted) tiling to be complete, each element
from the horizontal string $\D(\y)$ needs to be combined with each
element in the vertical string $\D(\x)$. This can be achieved by
shifting the horizonal and vertical domino strings in opposite
directions:
\begin{equation}
  \T(\x,\y) = \{\D(\y - \x)_\N, \D(\x + \y)_\E, \D(\y - \x)_\S, \D(\x + \y)_W \}.
\end{equation}
Note if both the horizontal and vertical domino string are shifted in
the same direction, the north/east and south/west edge would always be
the same. 

Combining both observations, yields that the resulting tiling $\T$ and
its complement $\bar{\T}$ are complete, and thus $\T$ forms a valid
Dual Wang tile packing.

There exist many possible complete domino strings for $\C$ colors.
Algorithm~\ref{alg:domino} provides a generative algorithm for
generating a complete $\C$-colored domino string.  This algorithm
exploits the idea that interleaving all colors one by one in between a
fixed color, yields all combinations for that given fixed color on
either the north/east and south/west edge. By ensuring that each
sequence of tiles (for a given fixed color) starts with the same
color, we can concatenate the sequences without loss of a tile
combination. The algorithm for generating domino strings is not
limited to odd colored domino strings only; \autoref{fig:domino} shows
complete domino strings generated using~\autoref{alg:domino} for
$2, 3$ and $4$ colors.

\begin{algorithm}[t]
  \caption{Produce a complete $\C^2$-length domino string that contains all $\C$-color domino tiles.}
  \label{alg:domino}
  \SetAlgoLined
  \KwData{colors \C}
  \KwResult{$\C^2$-length domino string $\D(\y)$}
  $\y \leftarrow 0$\\
  \For{$c \gets0$ to $(\C-1$)}{
    \For{$i \gets0$ to $(c-1)$}{
      $\D(\y+0) \leftarrow \{i, c\}$\\
      $\D(\y+1) \leftarrow \{c, i+1\}$\\
      $\y \leftarrow \y + 2$
    }
    $\D(\y) \leftarrow \{c, 0\}$\\
    $\y \leftarrow \y + 1$
  }
\end{algorithm}

\begin{figure*}[t]
  \def\tilespace{3.2mm}
  \def\tilestart{2.0mm}
  \begin{tabular}{cl}
    \multirow{3}{*}{{\rotatebox[origin=c]{90}{2 color \hspace{-2.2cm}}}} &
    \includegraphics[height=2cm]{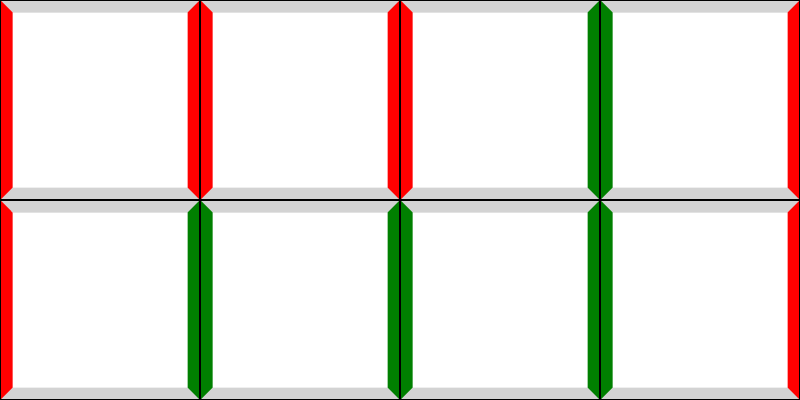} \\
    & \hspace{\tilestart}\{0,0\}, \hspace{\tilespace}\{0,0\}, \hspace{\tilespace}\{0,1\}, \hspace{\tilespace}\{1,0\}\\
    & \hspace{\tilestart}\{0,1\}, \hspace{\tilespace}\{1,1\}, \hspace{\tilespace}\{1,1\}, \hspace{\tilespace}\{1,0\}\\
    
    \multirow{3}{*}{{\rotatebox[origin=c]{90}{4 color \hspace{-2.2cm}}}} &
    \includegraphics[height=2cm]{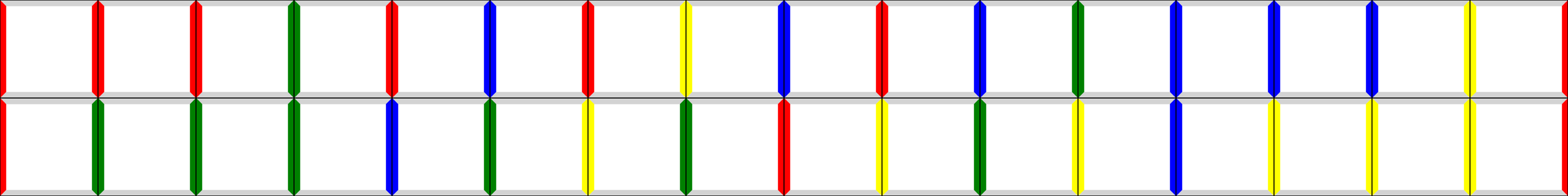} \\
    & \hspace{\tilestart}\{0,0\}, \hspace{\tilespace}\{0,0\}, \hspace{\tilespace}\{0,1\}, \hspace{\tilespace}\{1,0\}, \hspace{\tilespace}\{0,2\}, \hspace{\tilespace}\{2,0\}, \hspace{\tilespace}\{0,3\}, \hspace{\tilespace}\{3,2\}, \hspace{\tilespace}\{2,0\}, \hspace{\tilespace}\{0,2\}, \hspace{\tilespace}\{2,1\}, \hspace{\tilespace}\{1,2\}, \hspace{\tilespace}\{2,2\}, \hspace{\tilespace}\{2,2\}, \hspace{\tilespace}\{2,3\}, \hspace{\tilespace}\{3,0\}\\
    & \hspace{\tilestart}\{0,1\}, \hspace{\tilespace}\{1,1\}, \hspace{\tilespace}\{1,1\}, \hspace{\tilespace}\{1,2\}, \hspace{\tilespace}\{2,1\}, \hspace{\tilespace}\{1,3\}, \hspace{\tilespace}\{3,1\}, \hspace{\tilespace}\{1,0\}, \hspace{\tilespace}\{0,3\}, \hspace{\tilespace}\{3,1\}, \hspace{\tilespace}\{1,3\}, \hspace{\tilespace}\{3,2\}, \hspace{\tilespace}\{2,3\}, \hspace{\tilespace}\{3,3\}, \hspace{\tilespace}\{3,3\}, \hspace{\tilespace}\{3,0\}
      
  \end{tabular}
  \caption{Double domino strings generated using~\autoref{alg:zigzag}
    for $2$ and $4$ colors. For clarity, we also list the (numerical
    representation of) color-pairs for each domino tile.  Note that
    each sequence extends the previous sequence (albeit with the last
    domino's outer edge's color altered).}
  \label{fig:even_domino}
\end{figure*}

\subsection{Even Color Dual Wang Tile Packing}
\label{sec:even}

\begin{figure}[t!]
  \includegraphics[width=.45\columnwidth]{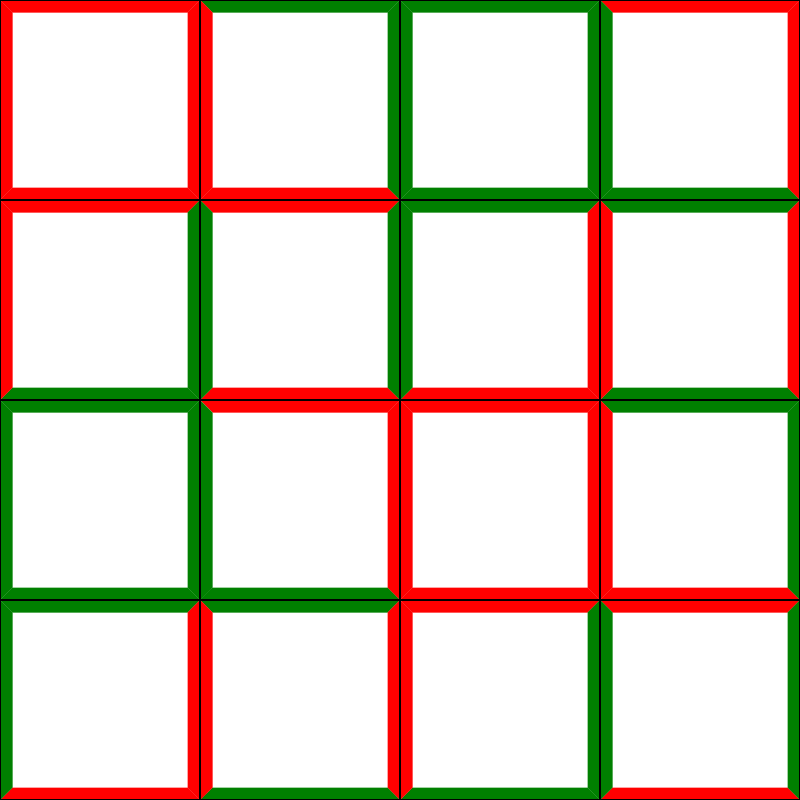}
  \caption{Invalid packing for an even ($2$) color tiling generated by
    shifting a template domino string (\autoref{alg:domino}) in
    opposite horizontal and vertical directions.  For
    example, the top-left tile (all red) also occurs at position
    (2,2).}
  \label{fig:evenfail}
\end{figure}

Applying the generative algorithm for odd color Dual Wang tile
packings to Dual Wang tiles with an even number of colors does not
yield a valid packing.  As shown in~\autoref{fig:evenfail}, only half
of the tiles occur twice and the other half is excluded.  When
tracking a single domino (e.g., the first domino tile in the
horizontal domino string), we observe that for each column, the
relative distance traveled in the vertical domino string equals two.
Hence, after $C^2 / 2$ steps (of visiting odd numbered dominoes in the
string), we wrap around. Due to the even number of colors, $C^2$ is
also even, and thus, we arrive (after wrap around) at the first domino
tile again.  In the odd colored case, $C^2$ is also odd, and thus
after wrap around, we arrive at the second domino tile, and
subsequently visit the odd dominoes, thus visiting all possible
combinations.

For even color Dual Wang tiles, a possible solution would be to use
a different domino string template for the vertical domino strings
that would be shifted by 2, i.e., $\D_\x(\y) = \D(\y \pm
2\x)$. However, this is not possible for complete domino strings.
This can easily be proven for the case where $\C = 2$. Assign the
first color to the first edge. As the string needs to be complete, we
will need to include the first color also on the second edge (in order
to include the domino with both edges the same first color).  However,
if we also want to fulfill the shift-by-two condition, then the third
edge also needs to be assigned the first color (for including the same
combination with both edges the same color), yielding a sequence of
three consecutive first colors and only leaving one more edge to be
assigned. Consequently, the resulting sequence will never be able to
include the domino with both edges assigned the second color.  We have
also computationally verified that no such sequence exists for
$\C = 4$.

The previous failed experiment also indicates that in order to obtain
a packing that we cannot rely on complete domino strings if using the
shifted domino template for the other dimension.  Hence, instead of
utilizing a single domino string, we opt for using a template of two
neighboring domino strings that together are (twice) complete.
Consider the case where the colors of the vertical edges are
determined by the double domino string template without shifting, and
the horizontal edge colors by a shifted single string as before:
\begin{equation}
  \T(\x, \y) = \{\D(\x + \y)_\N, \D_{(\x \mod 2)}(\y)_\E, \D(\x + \y)_S, \D_{(\x \mod 2)}(\y)_\W\},
\end{equation}
where $\D_0$ and $\D_1$ (i.e., the result of $(\x \mod 2)$ is either
$0$ or $1$) are the first and second domino string respectively, and
$\D_0$ is copied in even number rows and $\D_1$ in odd numbered rows.

\begin{algorithm}[t]
  \caption{Produce two $\C^2$-length domino strings containing all
    $\C$-color dominoes twice, and meeting the conditions
    in~\autoref{sec:even}.}
  \label{alg:zigzag}
  \SetAlgoLined
  \KwData{colors \C}
  \KwResult{$\C^2$-length domino strings $\D_0(\y)$ and $\D_1(\y)$.}
  $\y \leftarrow 0$\\
  \For{$c \gets0$ to $(\C-1$) by 2}{
    \For{$i \gets0$ to $(\C-1)$}{
      \tcp{First domino sequence}\tcp{Fixed color 'c' at even edges}
      $\D_0(\y-1) \leftarrow \{(i+\C-1) \mod \C, c\}$\\
      $\D_0(\y+0) \leftarrow \{c, i\}$\\
      \tcp{Second domino sequence}\tcp{Fixed color 'c+1' at odd edges}
      $\D_1(\y+0) \leftarrow \{i, c+1\}$\\
      $\D_1(\y+1) \leftarrow \{c+1, (i+1) \mod \C\}$\\
      $\y \leftarrow \y + 2$
    }
  }
\end{algorithm}

To derive the necessary conditions for $\D_0$ and $\D_1$, observe that
the first domino $\D(0)$ in the vertical domino string is combined
(after shifting) with the even number dominoes in $\D_0$ and the odd
dominoes in $\D_1$.  Consequently, the union of both:
$\{\D_0(2i), \D_1(2i + 1)\}_i$ must form a complete sequence.
Similarly, the second domino $\D(1)$ visits the complementary sequence
$\{\D_1(2i), \D_0(2i+1)\}_i$ which must also be complete.

The above conditions ensure that the resulting tiling is complete,
however, it does not ensure that the resulting complementary tiling is
also complete.  The necessary condition for a complete complementary
tiling can be derived in a similar manner.  First observe that by
copying the template without shift, that each complementary row will
be a repetition of the same two dominoes, namely
$\{\D_0(\y)_\N, \D_1(\y)_\N\}$ and its mirror
$\{\D_1(\y)_\N, \D_0(\y)_\N\}$.  Similarly following the the first
element from the complementary vertical domino string, yields that it
only visits the even number rows in $\D_0$ and odd rows in $\D_1$.
Consequently, by requiring that the sequence
$\{ {\D_{(i \mod 2)}(i)_\N, \D_{((i+1)\mod 2)}(i)_\N} \}_i$ is
complete ensures a complete complementary packing.

Similar as for the complete single domino string, many possible
sequences exist that meet both above conditions for the double domino
strings.  Algorithm~\ref{alg:zigzag} provides a generative algorithm
to produce a duo of domino strings that meets the above conditions.
Intuitively, we place every domino with one edge fixed to any of the
even colors in $\D_0$ and the combinations with one edge fixed with an
odd color in $\D_1$ as well as switching which edge will be fixed for
each (odd vs even edges).  The latter ensures that the complement will
be complete, whereas the former ensures that the regular (zig-zag)
sequences are complete.  \autoref{fig:even_domino} shows examples of
$2$ and $4$ color template strings.

\begin{figure*}[h]
  \centering
  \scalebox{.7}{
  \begin{tabular}{cc}
    \parbox[c]{.33\linewidth}{
      \vspace{-12cm}
    \hspace{1.3cm}\includegraphics[width=.5\linewidth]{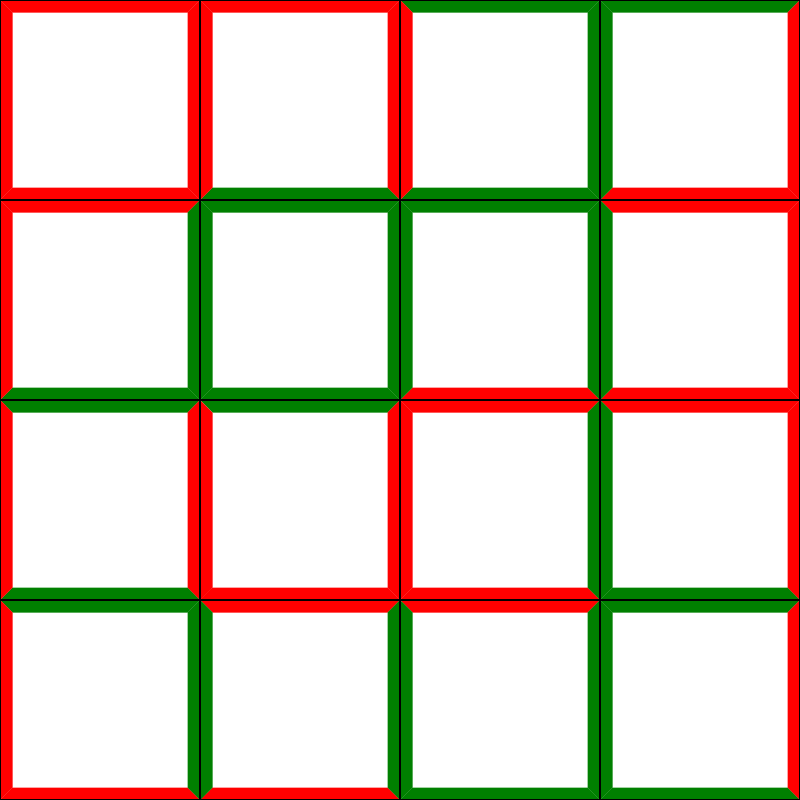}\\
    ~\\
    \includegraphics[width=\linewidth]{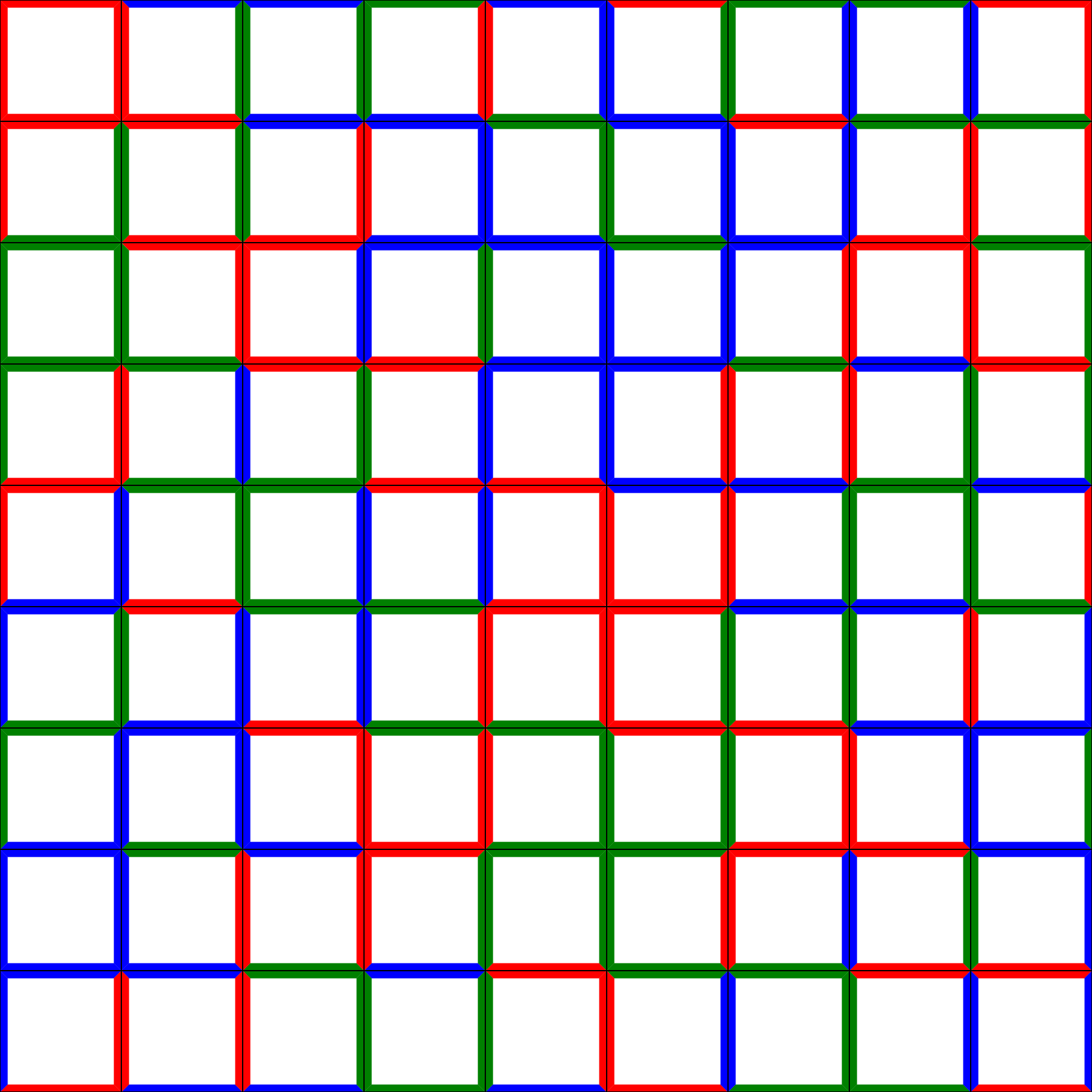}
    } &
    \includegraphics[width=.65\linewidth]{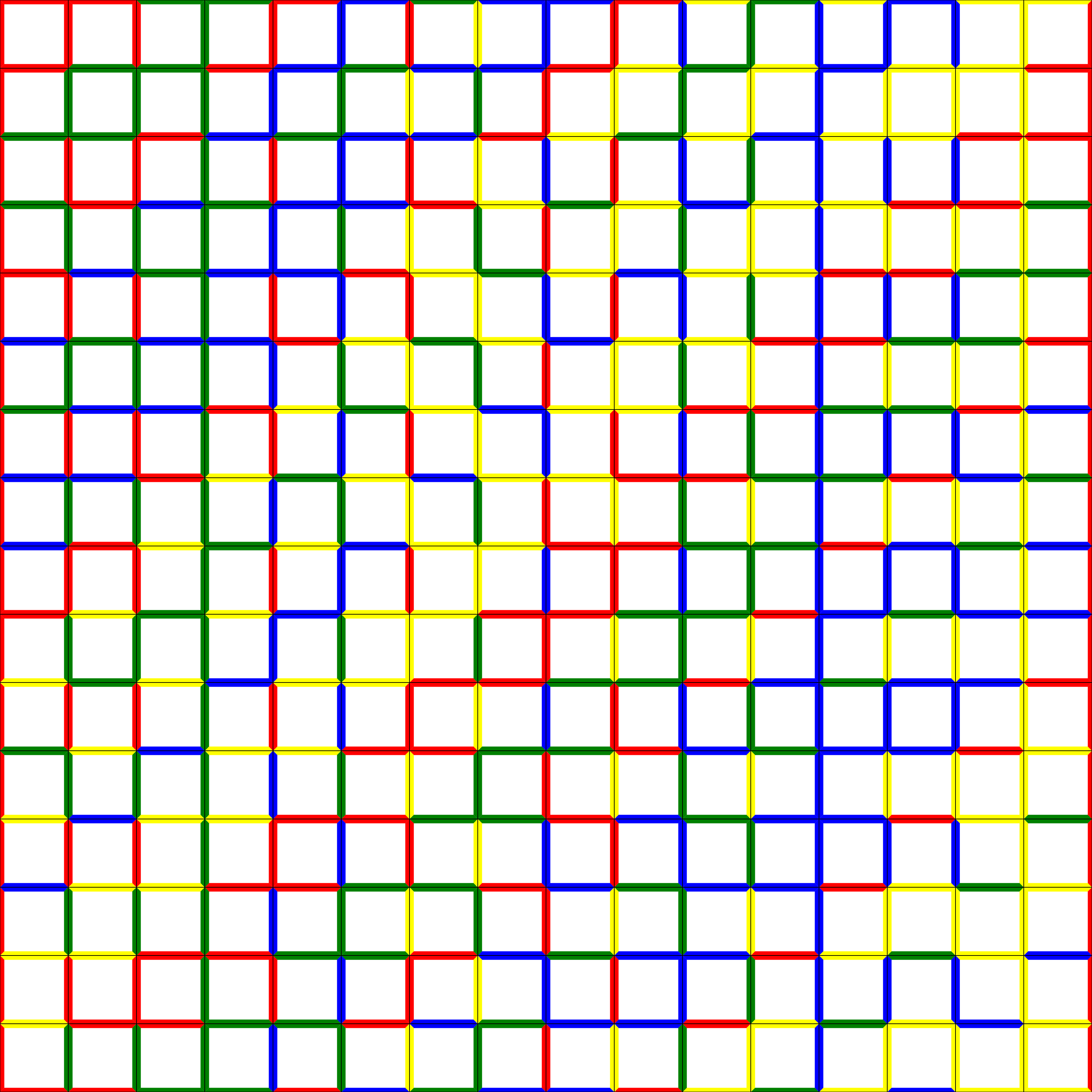} \\
    ~\\
    \multicolumn{2}{c}{\includegraphics[width=\linewidth]{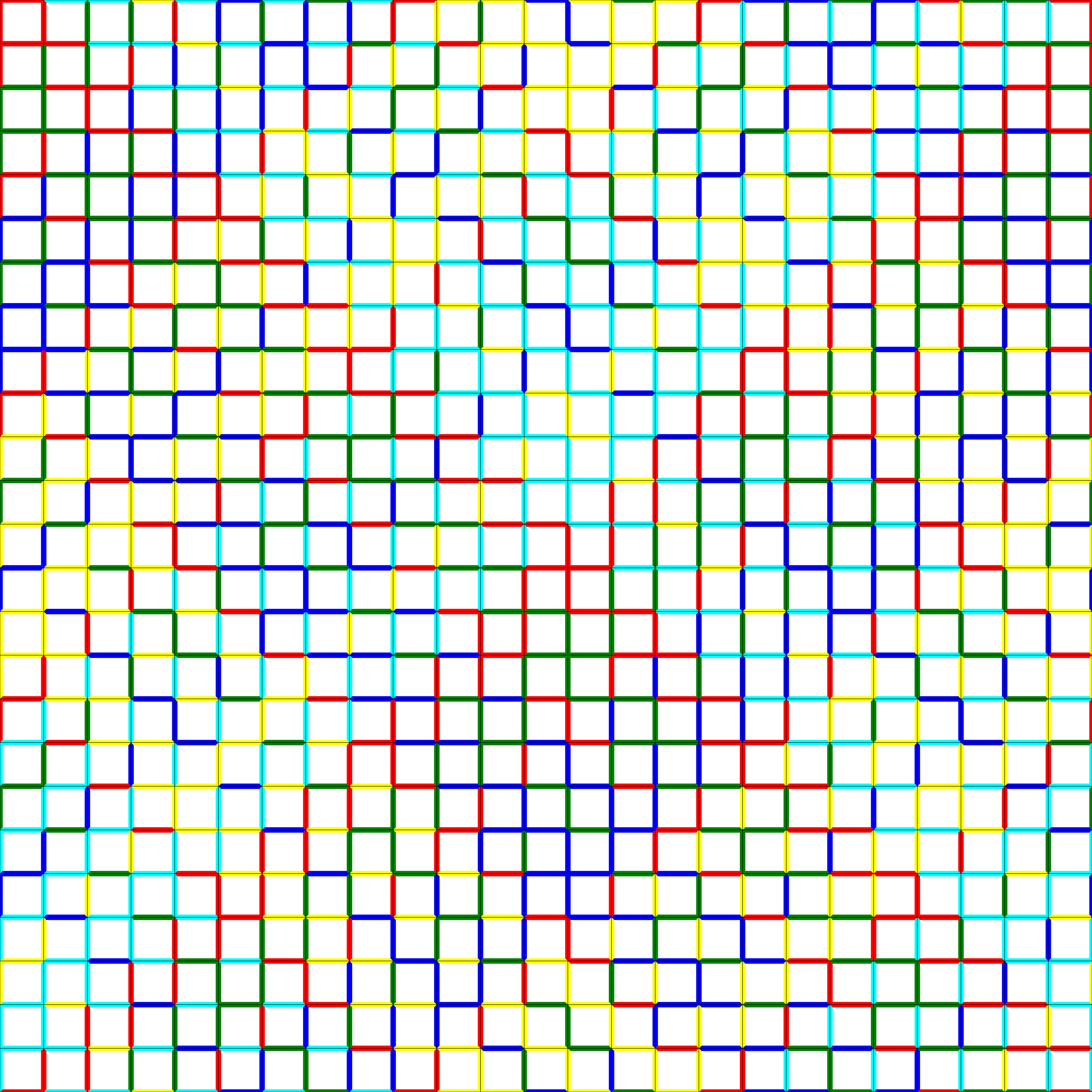}}
  \end{tabular}
}
\caption{Dual Wang tile packing for $2, 3, 4$, and $5$ colors.}
\label{fig:packing}
\end{figure*}

\autoref{fig:packing} shows examples of generated Dual Wang tile
packings for $2, 3, 4$, and $5$ colors.  We refer to
\url{https://github.com/samsartor/content_aware_tiles} for a reference
implementation of both packing algorithms.

\end{document}